\documentclass{article}

\usepackage{arxiv_preprint,times}

\usepackage{microtype}
\usepackage{graphicx}
\usepackage{subcaption}
\usepackage{booktabs}
\usepackage{array}
\usepackage{longtable}
\usepackage{arydshln}
\usepackage[hidelinks]{hyperref}
\usepackage{xurl}
\usepackage[acronym,nomain]{glossaries}
\setacronymstyle{long-short}
\glsdisablehyper
\newacronym{rdl}{RDL}{relational deep learning}
\newacronym[longplural={Graph Neural Nsetworks}]{gnn}{GNN}{graph neural network}
\newacronym[longplural={Graph Transformers},shortplural={GTs}]{gt}{GT}{Graph Transformer}
\newacronym{mae}{MAE}{mean absolute error}
\newacronym{rocauc}{ROC--AUC}{area under the receiver operating characteristic curve}

\newacronym{ce}{CE}{cross-entropy}
\newacronym{fk}{FK}{foreign-key}
\newacronym{id}{ID}{identifier}
\newacronym{rq}{RQ}{research question}
\newacronym{dnf}{DNF}{did not finish}
\newacronym{ctr}{CTR}{click-through rate}
\newacronym{ts}{TS}{Temporal Subgraph Contrast}
\newacronym{thgfm}{\textsc{THGFM}}{Dual-Branch Temporal Heterogeneous Graph Fusion Model}
\newacronym{timemix}{\textsc{TimeMix}}{multi-scale time encoding}
\newacronym{timerope}{\textsc{TimeRoPE}}{rotary time encoding}
\newacronym{relhist}{\textsc{Rel-Hist}}{Historical Relation Recovery}
\newacronym{relfuture}{\textsc{Rel-Future}}{Horizon-aware Relation Activity Prediction}
\newacronym{tempsub}{\textsc{Temp-Sub}}{Temporal Subgraph Contrast}

\usepackage{amsmath}
\usepackage{amssymb}
\usepackage{mathtools}
\usepackage{amsthm}
\usepackage[capitalize,noabbrev]{cleveref}

\theoremstyle{plain}

\theoremstyle{definition}

\theoremstyle{remark}

\newcommand{\method}{\gls{thgfm}}
\newcommand{\heterognn}{\textsc{HeteroGNN}}
\newcommand{\mix}{\gls{timemix}}
\newcommand{\rotm}{\gls{timerope}}
\newcommand{\ehist}{\gls{relhist}}
\newcommand{\efuture}{\gls{relfuture}}
\newcommand{\nattr}{\textsc{AttrMasking}}
\newcommand{\nschema}{\textsc{ContextPred}}
\newcommand{\gsubgraph}{\gls{tempsub}}

\title{Temporal Heterogeneous Graph Pretraining for Relational Deep Learning}

\author{%
Yixin Peng$^{1\star}$, Er Jin$^{1}$, Diego Collarana$^{2}$, and Stefan Decker$^{1,2}$ \\
$^{1}$ RWTH Aachen University, Aachen, Germany \\
\texttt{\{peng,jin,decker\}@dbis.rwth-aachen.de} \\
$^{2}$ Fraunhofer FIT, Sankt Augustin, Germany \\
\href{mailto:diego.collarana.vargas@fit.fraunhofer.de}{\texttt{diego.collarana.vargas@fit.fraunhofer.de}}}

\hypersetup{
  pdftitle={Temporal Heterogeneous Graph Pretraining for Relational Deep Learning},
  pdfauthor={Yixin Peng, Er Jin, Diego Collarana, Stefan Decker}
}

\begin{document}

\maketitle
\begingroup
\renewcommand{\thefootnote}{\ensuremath{\star}}
\footnotetext[0]{Corresponding author: Yixin Peng (\href{mailto:peng@dbis.rwth-aachen.de}{\texttt{peng@dbis.rwth-aachen.de}}).}
\endgroup

\begin{abstract}

% ————Version 2
Relational deep learning represents database rows and foreign-key links as a heterogeneous
graph, enabling prediction from record attributes and relational context.
Two forms of temporal signals play distinct roles in these graphs: record age changes as the prediction cutoff advances, whereas the interval between two observed records remains fixed. Prior work has explored temporal modeling and temporal pretraining for heterogeneous graphs, 
but typically treats time as a single source of information or focuses on either representation or supervision in isolation.
We investigate how explicitly representing
both temporal signals affects the benefits of temporal pretraining on downstream tasks.
Our framework pairs two complementary temporal encodings: Multi-scale Time Encoding, which captures record age through learnable time scales and type-specific projections, and Rotary Time Encoding, which encodes signed time differences between linked records through rotary transformations during graph propagation.
Rather than treating these encodings as architectural additions alone, 
we train them through three self-supervised objectives: 
recovering historical relations, forecasting horizon-dependent future relation activity, and contrasting temporally valid historical subgraphs.
All model inputs are restricted to information available at their observation cutoffs. 
We further structure pretraining into two stages: learning neighborhood representations via subgraph contrastive learning, followed by refining these representations through either relation recovery or future activity prediction.
We evaluate across five RelBench datasets and 11 classification and regression
tasks using representative heterogeneous \gls{gnn} and \gls{gt} backbones.
With both temporal encodings, the best evaluated staged schedules for the two backbones outperform direct supervised training using the same encodings by $3.02\%$ and $1.06\%$, respectively, and controls without pretraining or encodings by $3.24\%$ and $2.37\%$, respectively.

\end{abstract}

\section{Introduction}
\label{sec:introduction}

Relational databases store records in linked tables, with predictive signals
distributed across both record attributes and relationships.
For example, predicting whether a customer will buy again may require combining
customer details, product information, and purchase times from several tables.
Conventional pipelines assemble these signals through task-specific joins and
aggregations \citep{kanter2015dfs}. 
\Gls{rdl} instead represents rows as nodes and
foreign-key links as typed edges, allowing graph models to learn directly
from attributes and relational neighborhoods \citep{fey2024rdl}.

Learning from relational graphs requires modeling both heterogeneity and time.
Customer and product records describe distinct entity types, while purchase
records capture interactions between them. Together, they form a heterogeneous
graph with type-specific node attributes and multiple relation types.
Time provides two complementary signals about these records and their relationships.
\emph{Record age} measures the time elapsed between a record's observation and the
prediction cutoff. A signed \emph{inter-record interval} measures the difference between
the observation times of two linked records, capturing their temporal order and
distance. For example, as the prediction cutoff advances, a purchase record
becomes older, while the interval between the customer's registration and
that purchase remains fixed.

Temporal encodings make record age and inter-record intervals explicit in model
representations \citep{xu2020tgat,hu2020hgt,yu2023dygformer,cong2023graphmixer,li2023than,robinson2024relbench,dwivedi2026relgt}.
Pretraining guides models to integrate these temporal signals with record
attributes and graph structure before adaptation to downstream tasks.
Existing pretraining approaches use heterogeneous reconstruction and relation-
or schema-based contrastive learning \citep{hu2020gpt,jiang2021pre,sun2025phe},
while temporal objectives further target interaction timing or statistics of future
records \citep{ma2026tgpm,truong2025tve}.
Temporal encodings and pretraining objectives thus serve complementary roles:
the former specify how time is represented, and the latter guide what and how the model learns from these representations. This motivates us to study how temporal heterogeneity-aware pretraining
can be combined with explicit encodings of both temporal signals.
\Cref{fig:pretraining-comparison} summarizes these design choices across prior
methods and our framework.

We address this challenge with a framework that integrates temporal encodings with staged pretraining.
Our temporal encodings capture both record age and relative timing:
\Gls{timemix} incorporates record-age information into node inputs through learnable time scales and type-specific projections, while
\Gls{timerope} adapts prior temporal rotary attention \citep{peng2026thgfm} to use bounded, signed log-intervals and extends it to GNN message construction.

Building on these encodings, we introduce three complementary pretraining objectives.
\Gls{tempsub} aligns corrupted views of historical neighborhoods through type-aware pooling and relation-preserving augmentations, encouraging representations to remain stable under missing attributes or links.
\Gls{relhist} recovers hidden typed links using only the context available when those links were observed, whereas \Gls{relfuture} predicts whether an entity will receive links of a specified relation and how many events will occur within a future window.
We organize these objectives into a staged pretraining scheme:
\gsubgraph{} first learns neighborhood representations, which are then refined using either \ehist{} or \efuture{}.
All model inputs respect their observation cutoffs throughout pretraining.
The resulting pretrained model is subsequently fine-tuned for individual tasks within the same database.

We evaluate our framework on five RelBench databases \citep{robinson2024relbench} across
11 tasks spanning binary classification, multiclass classification, and
regression using representative heterogeneous
GNN and \gls{gt} backbones. Controlled ablations assess the complementary
benefits of the temporal encodings and the effectiveness of staged pretraining (\Cref{fig:temporal-encoding-ablation}).

Our contributions are:
\begin{itemize}
    \item \textbf{Complementary temporal representations.}
    We establish that jointly encoding record age and inter-record
    intervals improves aggregate performance over either alone on both
    backbones, with record age contributing more in component ablations.

    \item \textbf{Effective staged pretraining.}
    Our two staged schedules yield positive aggregate transfer in all
    four backbone--encoding settings. With both encodings, the best
    schedule on each backbone outperforms either constituent objective
    alone, demonstrating the value of staging complementary supervision.

    \item \textbf{Gains beyond temporal encoding alone.}
    Matched comparisons establish additional mean type gains of $3.02\%$
    on the GNN and $1.06\%$ on the \gls{gt} over supervised training
    with both encodings. Our best schedules also outperform the evaluated
    pretraining baselines, including their staged variants.
\end{itemize}

\section{Background and Related Work}
\label{sec:related}

\subsection{Relational Deep Learning}
\label{sec:setup}

RDL provides an end-to-end framework for learning from linked database
records \citep{fey2024rdl,dwivedi2025rdlsurvey}. It combines multimodal
attributes, schema-defined relationships, and temporal observations in a
graph representation. RelBench supplies predictive tasks with chronological
splits \citep{robinson2024relbench}.
% , and WikiDBs expands the available corpus of relational databases \citep{vogel2024wikidbs}.

\begin{figure}[t]
    \centering
    \includegraphics[width=0.7\textwidth]{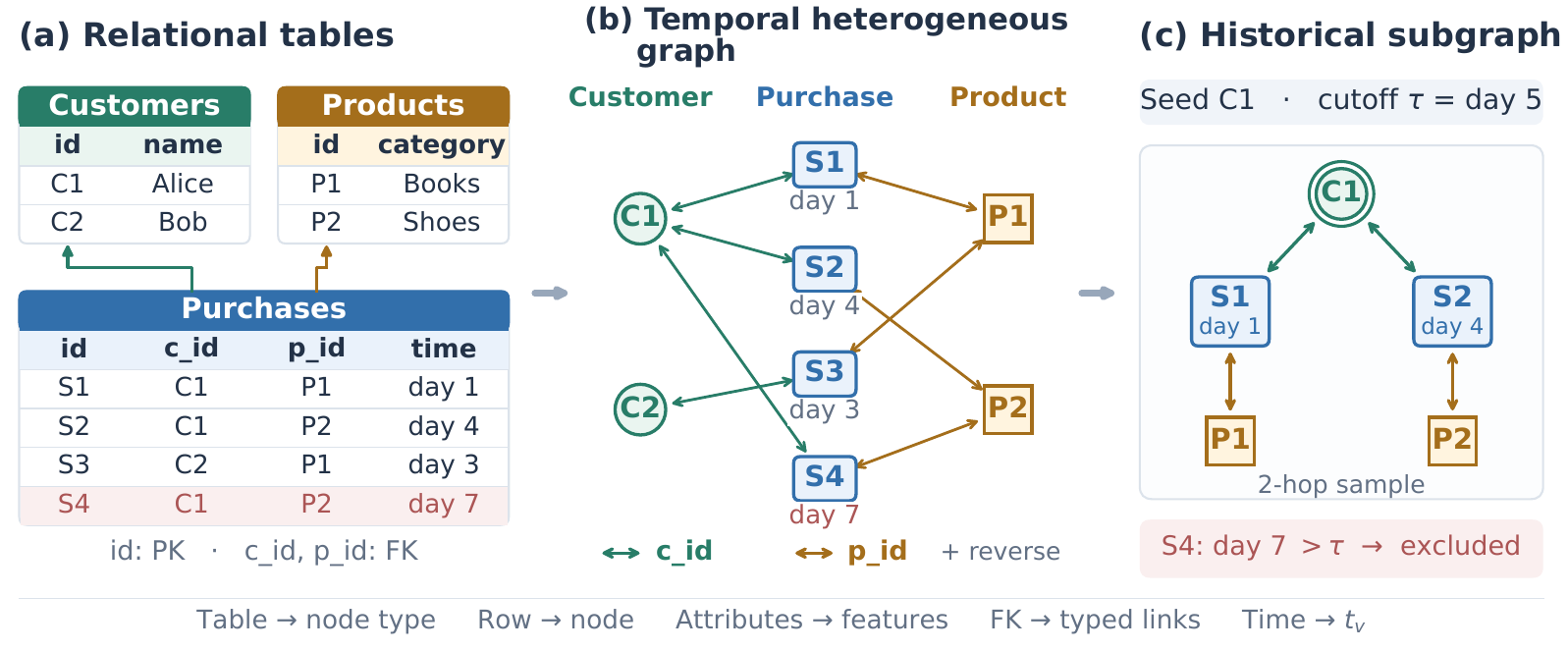}
    \caption{From relational tables to historical subgraphs for RDL.
    (a) Customer, purchase, and product tables:
    \texttt{id} denotes primary keys, \texttt{c\_id} and \texttt{p\_id}
    denote foreign keys, and \texttt{time} records purchase observation times.
    (b) Rows become nodes whose types follow their source tables;
    foreign-key references become typed edges with corresponding reverse
    relations for bidirectional message passing.
    Purchase nodes carry timestamps $t_v$.
    (c) An illustrative two-hop sample around seed C1,
    using only information available by cutoff $\tau=\text{day }5$.
    S4 is excluded because its timestamp exceeds $\tau$;
    S3 and C2 lie outside the two-hop neighborhood.}
    \label{fig:rdl-pipeline}
\end{figure}

\paragraph{Definitions.}
Let $\mathcal D=(\mathcal T,\mathcal R)$ be a database with tables
$\mathcal T=\{T_1,\ldots,T_n\}$ and foreign-key relations $\mathcal R$.
Each row is identified by a primary key and contains attributes and, where
applicable, foreign keys referencing other rows. A timestamp, when present,
records its observation time.
Following the relational entity graph formulation
\citep{fey2024rdl}, we construct
$G=(V,E,\phi,\psi,t)$, where $V$ contains the rows,
$E$ contains their foreign-key links,
$\phi:V\rightarrow\mathcal T$ assigns node types by source table, and
$\psi:E\rightarrow\mathcal R$ assigns relation types.
The timestamp function $t$ is defined on timestamped nodes, with
$t_v=t(v)$. We include reverse edges and corresponding relation types in
$E$ and $\mathcal R$ for message passing.
\Cref{fig:rdl-pipeline}(a,b) illustrates this mapping: customer, purchase,
and product tables become distinct node types, and the two foreign keys in
each purchase define its links to a customer and a product.

% 我这里改的版本
% Following the relational entity graph formulation \citep{fey2024rdl}, we
% construct $G=(V,E,\phi,\psi,t)$, where $V=\bigcup_{i} T_i$ collects the rows as
% nodes, $E$ their foreign-key references as directed edges,
% $\phi:V\rightarrow\mathcal T$ the table a node came from,
% $\psi:E\rightarrow\mathcal R^{\pm}$ the relation an edge came from with
% $\mathcal R^{\pm}=\mathcal R\cup\{r^{-1}:r\in\mathcal R\}$, and
% $t:V_T\rightarrow\mathbb R$ the timestamp $t_v$ of a row, defined on the
% timestamped subset $V_T\subseteq V$. \Cref{fig:rdl-pipeline}(b) shows each
% purchase node linking to one customer and one product. Foreign keys are
% directed, so for each one we add a reverse edge to $E$ with the corresponding
% type $r^{-1}$, allowing messages to pass in both directions.

% \begin{figure}[t]
%     \centering
%     \includegraphics[width=0.7\textwidth]{figures/rdl_pipeline.pdf}
%     \caption{Key concepts of RDL.
%     (a) Tables store attributes, foreign keys, and timestamps.
%     (b) Rows become typed nodes, and primary--foreign-key relationships become typed links.
%     (c) The two-hop neighborhood of C1 at cutoff $\tau=\text{day }5$.}
%     \label{fig:rdl-pipeline}
% \end{figure}

%%
%From relational tables to the input the model actually sees.
%  (a) Three tables; \texttt{id} is a primary key, \texttt{c\_id} and
%    \texttt{p\_id} are foreign keys, and \texttt{time} records when a purchase
%   occurred.
%    (b) The relational entity graph: each row becomes a node coloured by its
%    source table, each foreign key becomes a typed edge, and every edge is
%    duplicated in reverse so that messages travel both ways. Purchase nodes
%   carry timestamps $t_v$.
%    (c) For seed C1 (double circle) at prediction cutoff $\tau=\text{day }5$,
%   a two-hop sample restricted to records observed by $\tau$. Two mechanisms
%    remove nodes: S4 is dropped because $\text{day }7>\tau$, while C2 and S3
%    are dropped because they lie more than two hops from C1.
 %   Bottom: the table-to-graph mapping in one line.
%%

\subsection{Graph Models and Temporal Encoding for RDL}

We consider two encoder families for RDL: GNNs based on relational
message aggregation, including \heterognn{} and RelGNN
\citep{robinson2024relbench,chen2025relgnn}, and GTs based on
heterogeneous attention, including HGT, RelGT, and \method{}
\citep{hu2020hgt,dwivedi2026relgt,peng2026thgfm}.

Viewed in the RDL setting, existing methods incorporate record
age through temporal features added to node inputs or included
in attention and sequence representations.
These features use positional encodings with trainable
projections in \heterognn{}, learnable trigonometric functions
in TGAT, THAN, and DyGFormer, and fixed-frequency cosine
functions in GraphMixer
\citep{robinson2024relbench,xu2020tgat,li2023than,
yu2023dygformer,cong2023graphmixer}.
RelGT incorporates seed-relative age features during record
tokenization \citep{dwivedi2026relgt}.
For inter-record intervals, HGT adds projected sinusoidal
encodings of source--target time differences to source
representations, while \method{} applies
interval-dependent query--key rotations
\citep{hu2020hgt,peng2026thgfm}.
Additionally, record-age distributions and time scales vary across the
evaluated RelBench databases (Appendix~\ref{app:temporal-representations}).
Motivated by these observations and the complementary roles of
these two forms of temporal information, we develop two encodings and integrate both into the GNN and GT backbones.
% :\mix{} and \rotm{}.
% record age through learnable time scales
% and type-specific projections, while \rotm{} encodes signed
% intervals between linked records through rotary transformations and applies them to GNN message
% construction and GT query--key attention computation.

\subsection{Graph Pretraining: Heterogeneity, Time, and Staging}

\begin{figure}[t]
    \centering
    \includegraphics[width=0.7\textwidth]{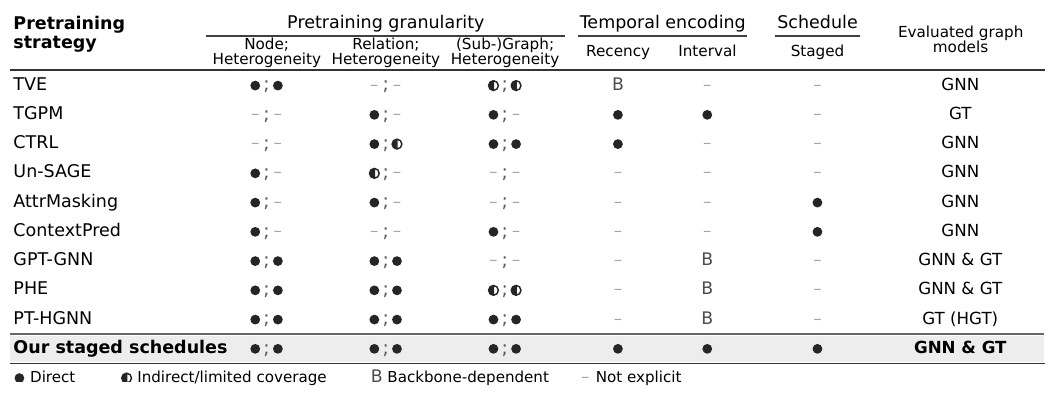}
    \caption{Comparison of representative graph pretraining strategies.
    In each of the first three columns, paired symbols indicate coverage
    of the supervision granularity and explicit use of node or relation
    types at that granularity, respectively.
    Filled and half-filled circles denote direct and indirect or limited
    coverage; B denotes backbone-dependent encoding, and a dash denotes
    no explicit coverage.}
    \label{fig:pretraining-comparison}
\end{figure}

% \begin{figure}[t]
%     \centering
%     \includegraphics[width=0.7\textwidth]{figures/pretraining_comparison.pdf}
%     \caption{Comparison of representative pretraining strategies across granularity, heterogeneity,
%     temporal encoding, staging, and evaluated graph architectures.}
%     \label{fig:pretraining-comparison}
% \end{figure}

Graph pretraining provides supervision over nodes, relations, and
subgraphs, but coverage of these granularities does not by itself imply explicit use of heterogeneous types or schema (\Cref{fig:pretraining-comparison}). AttrMasking and GraphMAE learn through
attribute reconstruction \citep{hu2020pretrain,hou2022graphmae,hou2023graphmae2};
Un-SAGE and ContextPred learn from proximity and neighborhood context
\citep{hamilton2017inductive,hu2020pretrain}, while GCC and GraphCL contrast
subgraph or graph views \citep{qiu2020gcc,you2020graphcl}.
Heterogeneous pretraining makes schema structure explicit in the
supervision: GPT-GNN uses type-specific attribute and link generation,
PT-HGNN combines relation-wise sampling with schema-instance agreement,
and PHE incorporates type-aware schema representations and semantic
neighbors \citep{hu2020gpt,jiang2021pre,sun2025phe}. HGMAE and MUG alternatively
use metapaths in reconstruction or alignment
\citep{tian2022hgmae,shan2026mug}. For RDL, this distinction matters because table and relation types determine which entities and links a pretraining
target describe. These methods establish ways to learn from typed structure; temporal encoding and temporally defined targets introduce additional design choices.

Time can condition the representation of observed history or define what the model must predict. TGPM combines masked patch
reconstruction with next-time prediction \citep{ma2026tgpm}, while CTRL models temporal influence through heterogeneous event-subgraph and
edge prediction
\citep{li2026ctrl}. In relational databases, Griffin
uses completion pretraining and supervised training, while RelPrism
constructs attribute-based pseudo-tasks
\citep{wang2025griffin,yang2026relprism}. TVE directly targets future
relational information: it predicts statistics, including counts, of
records reached through schema traversals \citep{truong2025tve}.
These targets differ in what they predict: interaction timing, event structure, or aggregated relational statistics.

How these objectives are organized is a further question.
\citet{hu2020pretrain} stage node-level and graph-level pretraining,
whereas TGPM jointly trains temporal objectives and TVE supports
weighted combinations of losses \citep{ma2026tgpm,truong2025tve}.
Our schedules study a different transition: first learn stable historical representations through contrastive
learning, then refine them through relation recovery or future activity
prediction, evaluating these schedules with explicit temporal encodings
across heterogeneous GNN and GT backbones.
Together, seed-node and historical-subgraph contrast, typed relation
recovery, and activity prediction cover node, relation, and subgraph
supervision, with heterogeneous types explicitly used in pooling,
sampling, decoding, and augmentation.

\section{Method}
\label{sec:method}

For seed node $s$ and prediction cutoff $\tau$, let $G_{s,\leq\tau}$ denote
a sampled neighborhood containing only information available by $\tau$.
We refer to this input as a \emph{historical neighborhood} or \emph{subgraph}.
We use $s$ to identify its seed and $u,v$ to denote general nodes or edge
endpoints. Temporal encoding and graph propagation operate on nodes and
edges throughout the sampled neighborhood.
In \cref{fig:rdl-pipeline}(c), S4 is excluded by time, while S3 lies outside
C1's two-hop neighborhood. Static tables contribute attributes available at
the cutoff even when they have no timestamp column. The model produces
the seed representation $h_s=f_\theta(G_{s,\leq\tau})\in\mathbb R^d$,
where $d$ is the hidden dimension.

\subsection{Temporal Encoding}
\label{sec:temporal-encoding}

\paragraph{Record age: \mix{}.}
\mix{} represents how old a record is at the prediction cutoff, allowing
the model to learn how recency matters for each table type.
For each timestamped node $v$ in the sampled neighborhood, we compute
its age relative to the same cutoff $\tau$:
\begin{equation}
    a_v=\max(0,\tau-t_v),\qquad
    \xi_{v,k}=\log\!\left(1+\frac{a_v}{s_k}\right),
\end{equation}
where $a_v$ is the record age and $\xi_{v,k}$ is its feature at learned
positive time scale $s_k$, for $k=1,\ldots,K_{\mathrm{time}}$.
We parameterize $s_k=\operatorname{clip}(\exp(\eta_k),s_{\min},s_{\max})$,
where $\eta_k\in\mathbb R$ is trainable and
$0<s_{\min}<s_{\max}$ are fixed bounds.
Here $K_{\mathrm{time}}$ is the number of scales, and ages and scales use
the same time unit. The logarithm preserves variation among recent records
while compressing large ages. Future records are excluded by sampling
before this encoding is applied.

Let $x_v\in\mathbb R^d$ be the table encoder's representation of record
attributes. For node type $a=\phi(v)$, we add a mixture of projected age features:
\begin{equation}
    e_v^{\mathrm{time}}=\sum_{k=1}^{K_{\mathrm{time}}}\pi_k g_{a,k}(\xi_{v,k}),\qquad
    \widetilde{x}_v=x_v+e_v^{\mathrm{time}}.
\end{equation}
Here $g_{a,k}:\mathbb R\rightarrow\mathbb R^d$ is a learned linear
projection for type $a$ and scale $k$, and $\pi_k$ is a learned nonnegative
mixture weight, with $\sum_k\pi_k=1$. The time embedding
$e_v^{\mathrm{time}}$ augments the attribute representation, giving
$\widetilde{x}_v$ as the input to graph propagation. Scales and weights are
shared across node types; type-specific projections allow the same age to
have different effects in different tables. For records without timestamps,
we use $\widetilde{x}_v=x_v$. Appendix~\ref{app:time-mix-details} gives
more details.

\paragraph{Time differences: \rotm{}.}
\rotm{} captures the order and spacing of linked records.
For a directed edge $u\rightarrow v$ in the sampled neighborhood, $u$ is
the source and $v$ is the destination.
When both endpoints have timestamps, define
$\Delta t_{uv}=t_v-t_u$ as the signed time difference. We convert it to a
rotation phase:
\begin{equation}
    \rho_{uv}=\operatorname{clip}\!\left(
    \operatorname{sign}(\Delta t_{uv})
    \log\!\left(1+\frac{|\Delta t_{uv}|}{s_{\mathrm R}}\right),
    -\rho_{\max},\rho_{\max}\right),
\end{equation}
where \(s_{\mathrm R}\) is a fixed positive time scale and \(\rho_{\max}>0\) is a fixed hyperparameter that bounds the magnitude of the rotation phase. The sign distinguishes temporal order;
the logarithm and clipping limit the effect of long gaps. Edges with an
untimestamped endpoint use the backbone's message or attention operation
without rotation.

For each node $v$, let $y_v\in\mathbb R^d$ denote its input to the current
graph layer; in the first layer, $y_v=\widetilde{x}_v$.
Let $R(\rho)$ rotate the $j$-th consecutive pair of feature dimensions
by angle $\omega_j\rho$, where $\omega_j$ is the corresponding RoPE
frequency~\citep{su2021roformer}. Our GNN backbone constructs messages
using opposite half-phase rotations and aggregates them by relation:
\begin{equation}
    \begin{aligned}
    m_{u\rightarrow v}&=\tfrac12\left[
    R(-\rho_{uv}/2)y_u+R(+\rho_{uv}/2)y_v\right],\\
    h_v^{(r)}&=W_{\mathrm{neigh}}^{(r)}
    \left(\frac{1}{|\mathcal N_r(v)|}
    \sum_{u\in\mathcal N_r(v)}m_{u\rightarrow v}\right)
    +W_{\mathrm{self}}^{(r)}y_v.
    \end{aligned}
\end{equation}
Here $\mathcal N_r(v)$ contains incoming neighbors under relation $r$,
and $h_v^{(r)}$ is the relation's output. The learned matrices
$W_{\mathrm{neigh}}^{(r)}$ and $W_{\mathrm{self}}^{(r)}$ transform averaged
messages and node inputs, respectively.

For our \gls{gt} backbone, we adapt the query--key rotation
mechanism of THGFM's Rotary Temporal Attention
\citep{peng2026thgfm}, using the bounded, signed log-interval
$\rho_{uv}$ defined above as the temporal phase.
% Within each attention head, queries and keys receive opposite
% half-phase rotations, while values remain unrotated and are
% aggregated using the resulting attention weights.
Thus, \rotm{} incorporates inter-record timing into GNN message
construction and GT attention weighting.
Appendix~\ref{app:encoding-details} provides the rotation
convention and backbone implementation details.

\subsection{Temporal Pretraining Objectives}

\paragraph{Historical link recovery: \ehist{}.}
\label{sec:ehist}
\ehist{} asks whether two records were linked, given the context available
when the link was observed. We sample an observed link $u\rightarrow v$
of relation $r$ and use its assigned observation time as cutoff $\tau$.
We then sample $K_{\mathrm{neg}}$ negative destinations of the same type as $v$ that are
available at $\tau$ and are not known destinations of $u$ under $r$ in
the pretraining data.

We encode historical neighborhoods centered on the source and on each candidate destination at the same cutoff \(\tau\), 
obtaining their node representations. We remove the target edge and its reverse from every input so that the encoder cannot directly observe the answer.
For endpoint representations $h_u$ and $h_v$, a relation-specific
bilinear decoder assigns the score
$s_{\mathrm{hist}}(u,v,r)=h_u^\top W_r h_v$, where
$W_r\in\mathbb R^{d\times d}$ is a learned matrix for relation $r$.
The mean softplus ranking loss $\mathcal L_{\mathrm{Rel\text{-}Hist}}$
encourages the observed destination to score above the sampled negatives.
Appendix~\ref{app:hist-details} gives the timestamp assignment and full loss.

\paragraph{Future relation activity: \efuture{}.}
\label{sec:efuture}
\efuture{} predicts whether a seed node $s$ will receive links of
relation $r$ in a future window and how many. Each forward foreign-key
link $e=(u\rightarrow s)$ represents an event, with time $t_e=t_u$
given by the source record's timestamp. Let $\mathcal E_r(s)$ be
the set of these events. For cutoff $\tau$ and horizon $\delta$,
we define the event count and prediction targets as
\begin{equation}
    \label{eq:future-targets}
    \begin{gathered}
    c_r(s;\tau,\delta)=\sum_{e\in\mathcal E_r(s)}
    \mathbf 1\{\tau<t_e\leq\tau+\delta\},\\
    y_{\mathrm{act}}=\mathbf 1\{c_r(s;\tau,\delta)>0\},\qquad
    y_{\mathrm{count}}=\log\!\bigl(1+c_r(s;\tau,\delta)\bigr).
    \end{gathered}
\end{equation}

To construct a positive example, we uniformly sample an eligible
relation and then an eligible event of that relation.
Let $s^+$ be its destination and $t_{\mathrm{evt}}$ its time.
We choose a candidate horizon $\delta$ and set
$\tau=t_{\mathrm{evt}}-\delta$, requiring
$t_{\mathrm{first}}(s^+)\leq\tau$, where $t_{\mathrm{first}}$
denotes the entity's first observation time.
This ensures that the seed is available at the cutoff and the
window contains at least the sampled event.
For the same $(r,\tau,\delta)$, we sample negative seeds $s^-$
of the same type as $s^+$ that are available at $\tau$ and have
no events in the window.

For each seed $s$, the encoder uses only $G_{s,\leq\tau}$ to
compute $h_s$. Two decoders, conditioned on $r$ and $\delta$,
map $h_s$ to an occurrence logit and a nonnegative log-count
prediction. We optimize
\begin{equation}
    \mathcal L_{\mathrm{Rel\text{-}Future}}
    =\mathcal L_{\mathrm{act}}
    +\lambda_{\mathrm{count}}\mathcal L_{\mathrm{count}},
\end{equation}
where $\mathcal L_{\mathrm{act}}$ is the binary occurrence loss,
$\mathcal L_{\mathrm{count}}$ is the Smooth L1 loss on log-count
targets, and $\lambda_{\mathrm{count}}$ weights the count loss.
Each loss gives equal weight to the positive example and the
mean over negatives. Appendix~\ref{app:future-details} provides
the timestamp assignment, sampling rules, and full losses.

\paragraph{Temporal subgraph contrast: \gsubgraph{}.}
\label{sec:gsubgraph}
\gsubgraph{} learns representations that remain consistent when some
features or links are missing from a historical neighborhood.
Let $\tau_{\mathrm{pre}}$ be the global pretraining cutoff, set to
the dataset's validation timestamp.
For timestamped seeds, we require $t_s\leq\tau_{\mathrm{pre}}$
and set $\tau=t_s$; for seeds without timestamps, we set
$\tau=\tau_{\mathrm{pre}}$.
For each input $G_{s,\leq\tau}$, we create two independently corrupted
views sharing the same seed node $s$ and cutoff $\tau$
by randomly masking input
embedding entries and dropping edges. We either keep or drop each edge together with its corresponding reverse edge, and each view retains at least one edge per
represented forward relation. 

To summarize each view, we first pool nodes within each type, then combine
the type summaries. Let $H$ be an augmented subgraph, $\mathcal T(H)$
its set of node types, and $V_a(H)$ its nodes of type $a$. We compute
\begin{equation}
    \mu_{H,a}=\frac{1}{|V_a(H)|}\sum_{u\in V_a(H)}h_u,\qquad
    c_{H,a}=\tanh\!\left(P_a\mu_{H,a}+e_a\right),
\end{equation}
where $h_u$ is computed within $H$, $\mu_{H,a}$ is the mean representation
of type $a$, and $c_{H,a}$ is its type summary. The learned projection
$P_a\in\mathbb R^{d\times d}$ and type embedding $e_a\in\mathbb R^d$
adapt the summary to its table type. A shared learned linear scoring
function $g_{\mathrm{pool}}:\mathbb R^d\rightarrow\mathbb R$ then assigns
attention weights $\beta_{H,a}$ to combine the summaries into $z_H$:
\begin{equation}
    \beta_{H,a}=\frac{\exp(g_{\mathrm{pool}}(c_{H,a}))}
    {\sum_{b\in\mathcal T(H)}\exp(g_{\mathrm{pool}}(c_{H,b}))},\qquad
    z_H=\sum_{a\in\mathcal T(H)}\beta_{H,a}c_{H,a}.
\end{equation}
Averaging within types prevents larger tables from dominating solely
through their node counts, while attention learns how much each type
contributes to the neighborhood representation.

We apply contrastive learning at two levels: the whole neighborhood
and its seed node $s$. For neighborhood-level contrast, we map the pooled representations
of both views through the same projection head. Each neighborhood
in the first view is trained to match its counterpart in the second
view, with other neighborhoods in the batch serving as negatives.
We compute an InfoNCE loss~\citep{oord2018cpc} using cosine similarity
divided by temperature $T$. We then repeat the matching from the
second view to the first and average the two losses to obtain
$\mathcal L_{\mathrm{graph}}$. For seed-level contrast, we take the representation of $s$
from each view and apply the same matching procedure through a
separate projection head, obtaining $\mathcal L_{\mathrm{seed}}$.
% Thus, $\mathcal L_{\mathrm{graph}}$ encourages consistent summaries
% of the same historical neighborhood, while $\mathcal L_{\mathrm{seed}}$
% encourages consistent representations of the central entity used
% for downstream prediction. 
To complement contrastive learning in the projection space, we
additionally regularize the seed representations with a VICReg
penalty~\citep{bardes2022vicreg} that discourages collapse by maintaining
variance across seeds and reduces redundancy across feature dimensions.
This penalty acts directly on the encoder outputs before the seed
projection head and is averaged over the two views,
giving $\mathcal R_{\mathrm{reg}}$. The full objective is
\begin{equation}
    \mathcal L_{\mathrm{Temp\text{-}Sub}}
    = \mathcal L_{\mathrm{graph}}
    + \mathcal L_{\mathrm{seed}}
    + \mathcal R_{\mathrm{reg}}.
\end{equation}
Appendix~\ref{app:objective-settings} and Appendix~\ref{app:subgraph-details} provide more details.

\section{Experiments and Results}
\label{sec:experiments}

\subsection{Experimental Setup}

\paragraph{Datasets and evaluation.}
We evaluate on 11 tasks from five RelBench databases, Arxiv, Avito, F1,
H\&M, and Event~\citep{robinson2024relbench}, using the original
chronological training, validation, and test splits. We report accuracy
for multiclass classification, ROC--AUC for binary classification, and MAE
for regression. All experiments are repeated with four random seeds,
$\{42,43,44,45\}$; reported task metrics are the means and standard
deviations across these runs, with each run evaluated at its best validation
checkpoint. Appendix~\ref{app:datasets} provides dataset and task details.
% Our primary summary metric, mean type gain, averages
% relative gains within each task type and then weights the three types
% equally. 

\paragraph{Backbones and baselines.}
We benchmark five backbones: two GNNs, \heterognn{} and RelGNN,
and three \glspl{gt}, HGT, RelGT, and \method{}.
% \citep{robinson2024relbench,chen2025relgnn,hu2020hgt,dwivedi2026relgt,peng2026thgfm}
The controlled encoding and pretraining study focuses on \heterognn{}
and \method{}, selected as the best-performing representatives of their
respective families by mean validation rank across the 11 tasks
(Appendix~\ref{app:backbone-selection}). Selecting one backbone per family is a trade-off between
architectural coverage and computational cost.
All backbones use three layers with hidden dimension 128, and \glspl{gt} use eight attention heads. 
% Temporal neighborhood sampling spans
% three hops with fanouts $(128,64,32)$, respectively.
We compare against nine pretraining baselines:
\nattr{}, \nschema{},
Un-SAGE, GPT-GNN, PT-HGNN, PHE, CTRL, TGPM, and TVE. Single-stage pretraining and downstream fine-tuning each use 10 epochs with 10,000 steps
per epoch. Two-stage pretraining uses 10 epochs with 5,000 steps per epoch
in each stage, matching the total single-stage pretraining budget of
100,000 steps. Models trained without pretraining use the same downstream
budget. Appendix~\ref{app:experimental-config} provides
the full architecture and training configuration.

% \nattr{}, \nschema{}~\citep{hu2020pretrain},
% Un-SAGE~\citep{hamilton2017inductive}, GPT-GNN~\citep{hu2020gpt},
% PT-HGNN~\citep{jiang2021pre}, and PHE~\citep{sun2025phe}.

% To isolate the effect of pretraining, we use controls trained directly on
% downstream tasks with the same backbone, temporal encodings, and downstream
% protocol. We also apply \gsubgraph{} before three baseline objectives to
% assess whether staging benefits extend beyond \ehist{} and \efuture{};
% Appendix~\ref{app:pretraining-transfer} lists all configurations.

\subsection{Main Results}
\label{sec:main-results}

\begin{figure}[t]
\centering
\includegraphics[width=\textwidth]{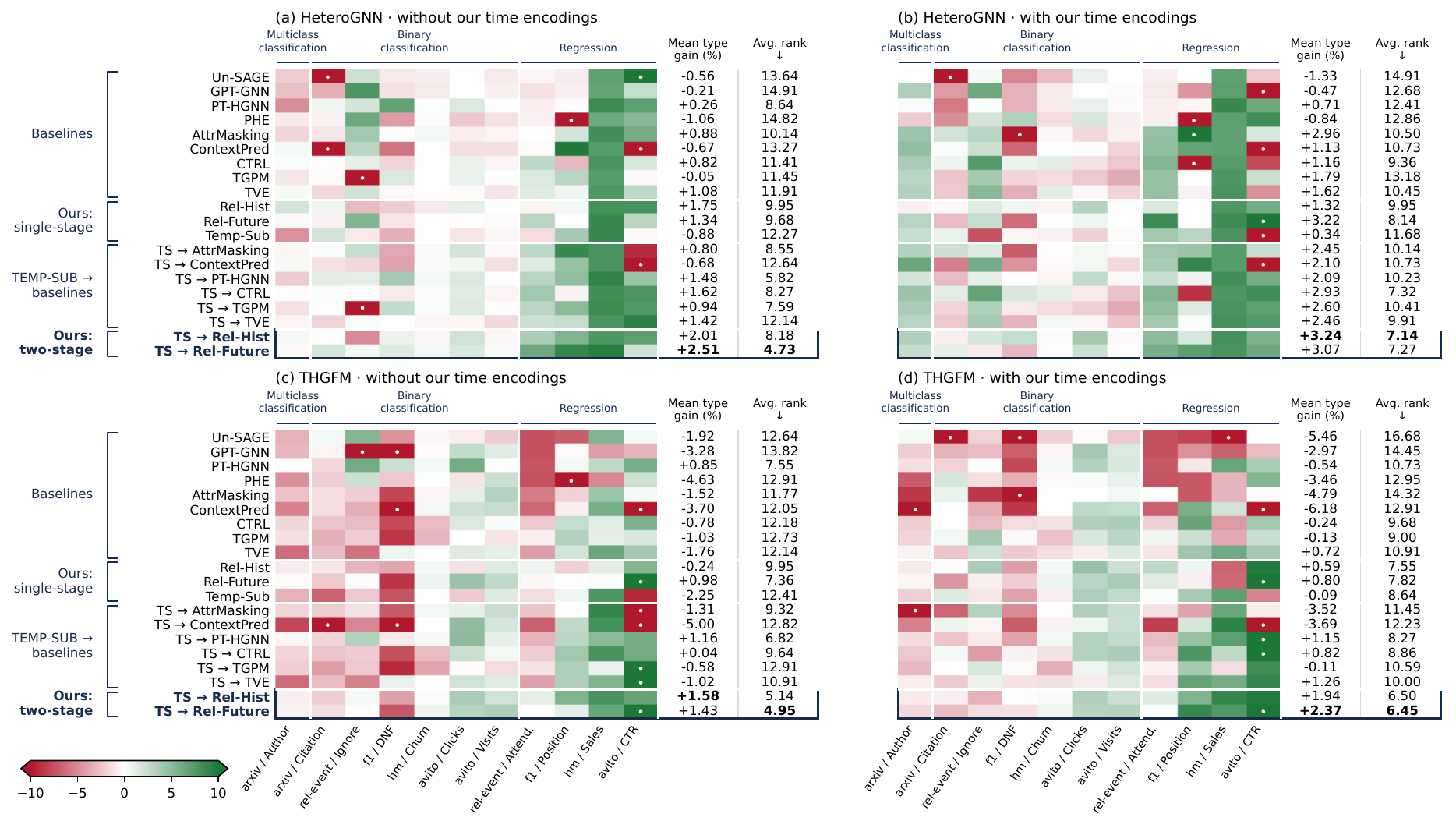}
\caption{Relative test-performance gains (\%) across 11 tasks,
computed from task metrics averaged over four random seeds.
All panels use the same-backbone supervised control without our time encodings (only with DayPE) or pretraining as reference (A0).
Mean type gain averages gains within each task type and then equally
across types; average rank weights all tasks equally (lower is better).
Colors saturate at $\pm10\%$; white dots mark values beyond this range.
TS denotes \gsubgraph{}; outlined rows identify our two-stage schedules,
and bold summary values mark the best result in each panel.}
\label{fig:pretraining-gains}
\end{figure}

% \begin{figure}[t]
% \centering
% \includegraphics[width=\textwidth]{figures/pretraining_gain_heatmap.pdf}
% \caption{Relative task gains (\%) over the corresponding backbone trained
% without pretraining or either proposed temporal encoding (A0).
% Right-hand columns show mean type gain and average rank (lower is better).
% White dots mark gains beyond the $\pm10\%$ color range.
% TS denotes \gsubgraph{}; outlined rows are our staged schedules.}
% \label{fig:pretraining-gains}
% \end{figure}

\paragraph{Aggregate performance across backbones.}
We denote the fixed day-based single-scale record-age encoding used by
\heterognn{} \citep{robinson2024relbench} as \textsc{DayPE} and use it as
the baseline for multi-scale \mix{} on both backbones
(Appendix~\ref{app:temporal-representations}).
\Cref{fig:pretraining-gains} uses the same-backbone supervised control
with only \textsc{DayPE} and without \rotm{} or pretraining as its reference
(A0) throughout, measuring the effects of pretraining and, where enabled,
our temporal encodings.
In each of the four backbone–encoding settings, one of our two proposed staged schedules achieves the highest mean type gain among the evaluated methods.
With both encodings, TS$\rightarrow$\ehist{} leads on the GNN at
$+3.24\%$, and TS$\rightarrow$\efuture{} leads on the \gls{gt} at
$+2.37\%$. The strongest baselines, including TS-initialized variants,
reach $+2.96\%$ on the GNN (\nattr{}) and $+1.15\%$ on the
\gls{gt} (TS$\rightarrow$PT-HGNN).
Without our encodings, TS$\rightarrow$\efuture{} leads on the GNN
($+2.51\%$), while TS$\rightarrow$\ehist{} leads on the \gls{gt}
($+1.58\%$).

\paragraph{Performance under task-balanced aggregation.}
The benchmark contains one multiclass, six binary, and four regression
tasks, so equal weighting of task types gives individual tasks different
weights. We therefore also report average ranks with all 11 tasks
weighted equally. Our staged schedules rank best in all four
panels, with average ranks of $4.73$, $7.14$, $4.95$, and $6.45$,
respectively. TS$\rightarrow$\ehist{} leads on the GNN with both
encodings, and TS$\rightarrow$\efuture{} leads in the other three
panels; with both encodings, our two staged schedules occupy the
top two positions on each backbone.
% Averaging relative gains equally across all 11 tasks, the same GNN and GT schedules achieve gains of +2.83\% and +3.10\% over A0, respectively.
Thus, the aggregate advantage is supported by both gain and rank
summaries.

\paragraph{Task coverage and sources of improvement.}
With both encodings, the best staged configuration for each backbone improves over A0 on 8 of the 11 tasks for the GNN and 7 for the GT, including all four regression tasks on both backbones.
Regression provides the largest task-type gains:
$+5.12\%$ on the GNN and $+8.55\%$ on the \gls{gt}.
On the latter, gains span driver position ($+8.74\%$), item sales
($+7.54\%$), and ad CTR ($+17.76\%$).
Classification gains are less uniform: the GNN schedule improves
author-category accuracy by $3.40\%$ and binary classification by
$1.21\%$ on average; the \gls{gt} schedule gains $0.26\%$
on binary classification but loses $1.70\%$ on author-category accuracy.
The positive aggregates therefore reflect benefits across several
regression targets alongside more variable classification transfer.
Appendix~\ref{app:pretraining-transfer} reports detailed test results and standard
deviations.

\subsection{Ablation Studies}
\label{sec:ablations}

\begin{figure}[t]
\centering
\includegraphics[width=0.8\textwidth]{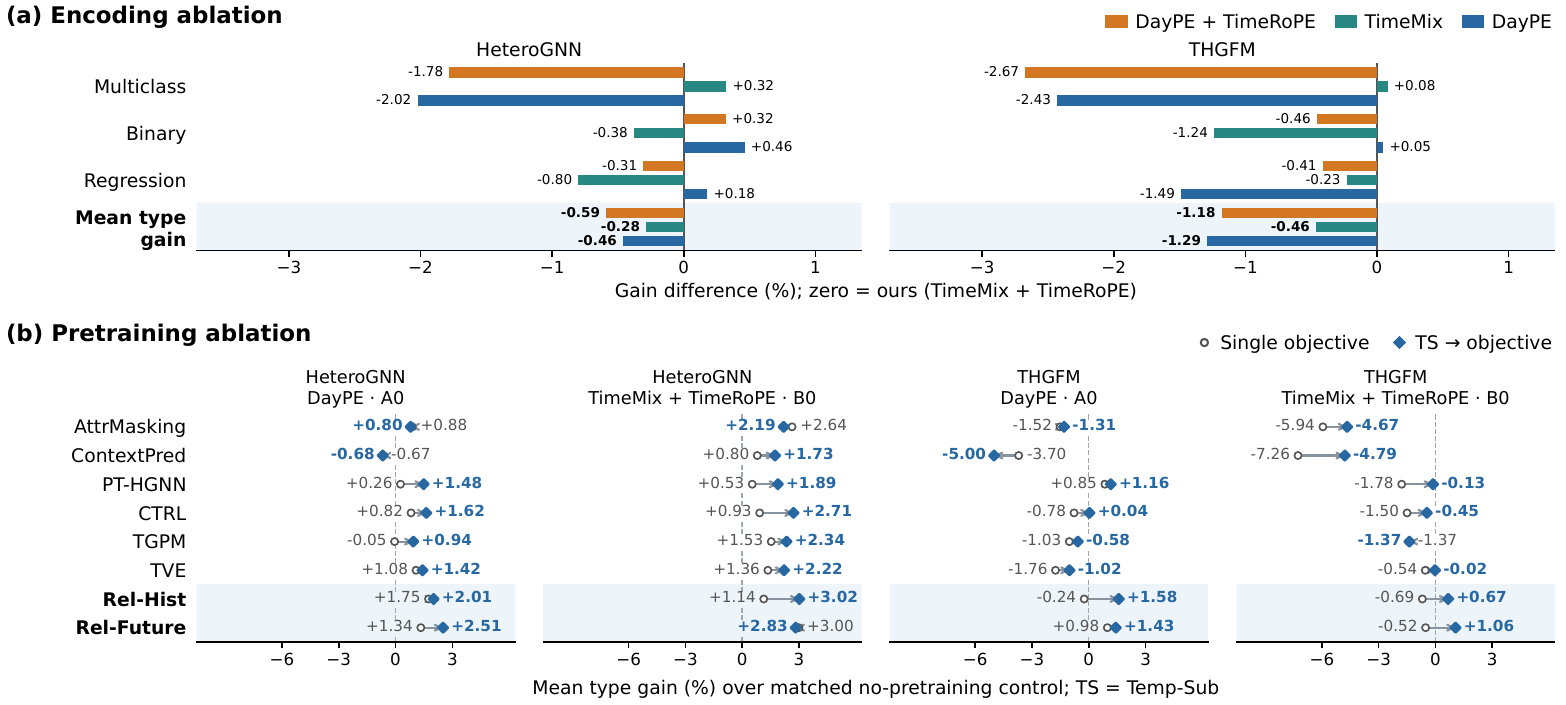}
\caption{Contributions of temporal encoding and staged pretraining.
(a) Gain differences from ours
(\mix{}+\rotm{}), without pretraining.
\textsc{DayPE} is the single-scale baseline for \mix{}.
(b) Pretraining gains over matched supervised controls
(A0: \textsc{DayPE}; B0: \mix{}+\rotm{}).
Arrows connect single- and two-stage results with equal total pretraining steps.
}
\label{fig:encoding-pretraining-ablations}
\label{fig:temporal-encoding-ablation}
\label{fig:pretraining-staging-ablation}
\end{figure}

% \begin{figure}[t]
% \centering
% \includegraphics[width=0.8\textwidth]{figures/encoding_and_pretraining_ablations.pdf}
% \caption{Ablation studies. (a) Changes in gain after removing temporal
% encodings, without pretraining. (b) Pretraining gains over supervised
% controls with the same backbone and encodings. Overall gains weight task
% types equally; TS denotes \gsubgraph{}.}
% \label{fig:encoding-pretraining-ablations}
% \label{fig:temporal-encoding-ablation}
% \label{fig:pretraining-staging-ablation}
% \end{figure}

\paragraph{Contributions of the temporal encodings.}
% Removing \mix{} restores \textsc{DayPE}, the single-scale baseline
% introduced in \cref{sec:main-results}, regardless of whether \rotm{} is enabled;
% Appendix~\ref{app:temporal-representations} details the encoding
% configurations and task-level results.
Removing \mix{} restores \textsc{DayPE}, the single-scale baseline
introduced in \cref{sec:main-results}, regardless of whether \rotm{} is enabled.
In \cref{fig:temporal-encoding-ablation}a, removing \mix{}
reduces mean type gain by $0.59$ percentage points on the GNN
and $1.18$ on the \gls{gt}, compared with $0.28$ and $0.46$
points from disabling \rotm{}.
Removing both lowers gains by $0.46$ and $1.29$ points,
respectively.
The combined configuration therefore achieves the highest
mean type gain on both backbones, with removing \mix{}
causing the larger decrease.
Effects vary across task types: disabling \rotm{} slightly
improves multiclass performance but reduces binary and
regression gains on both backbones.

\paragraph{Effect of staged pretraining.}
\Cref{fig:pretraining-staging-ablation}b measures pretraining gains
over supervised controls with the same backbone and temporal encodings.
Both proposed schedules yield positive mean type gains in all four
settings. With both encodings, TS$\rightarrow$\ehist{} achieves
$+3.02\%$ on the GNN, while TS$\rightarrow$\efuture{} achieves
$+1.06\%$ on the \gls{gt}, demonstrating benefits beyond temporal
encoding alone.
Under the same total pretraining budget, adding an initial
\gsubgraph{} stage improves over the corresponding single objective
in seven of the eight comparisons for \ehist{} and \efuture{}.
Notably, on the \gls{gt} with both encodings, staging turns negative
transfer into positive gains for both objectives, whereas all six
TS-initialized baselines shown remain below the supervised control.
The benefit nevertheless depends on the configuration:
\efuture{} alone slightly outperforms its staged counterpart on
the GNN with both encodings ($+3.00\%$ versus $+2.83\%$).
Among the baselines, staging improves PT-HGNN, CTRL, TGPM and TVE in all four settings,
while its effects on AttrMasking and ContextPred are mixed. 
One possible explanation is that the type-aware representations
learned by \gsubgraph{} better support relation- and
schema-based supervision (\Cref{fig:pretraining-comparison}).
% , whereas their benefits for attribute
% reconstruction and neighborhood-context prediction depend more
% on the configuration 

\subsection{Discussion and Limitations}
\paragraph{Temporal scales and horizon design.}
The statistics in Appendix~\ref{app:temporal-representations}
reveal substantial variation in record-age distributions across
databases. Avito includes records observed within one day before
the prediction cutoff, with age variation at second, minute, and
hour scales. Record ages in Event and H\&M are concentrated at
scales of days to months, whereas Arxiv and F1 include records
dating back years. These differences motivate the multi-scale
record-age representation in \mix{}.
% The transformations $\log(1+a_v/s_k)$ preserve differences among recent
% records while compressing large ages, and the learned scales adapt to
% the observed age distribution. Type-specific projections also allow
% the same record age to have different effects across tables.

The supported prediction horizons also vary across databases.
Avito supports only 1- and 7-day horizons, while Event and H\&M also
support windows spanning weeks to months. In Arxiv and F1, $67.7\%$
and $87.1\%$ of eligible events, respectively, support a 365-day horizon.
These differences motivate using multiple horizons for \efuture{}:
short windows provide training examples where history is limited,
while longer windows provide supervision on relation activity over
months or a year where sufficient history is available.
We therefore use candidate horizons from one day to one year.

\paragraph{Limitations.}
Our framework focuses on within-database training; transfer to unseen
databases remains to be explored. Single- and two-stage pretraining use
the same total number of training steps, although computational cost per
step can vary across objectives. Comparisons under matched computational
costs and with alternative stage orders would further clarify the
contributions of objective composition and stage order.

\section{Conclusion}
We present an RDL framework that encodes record age and inter-record
intervals and stages subgraph contrast before historical relation
recovery or future activity prediction.
Across five RelBench databases and 11 tasks, the two encodings together
outperform either alone in aggregate on both backbones, and staged pretraining yields
additional gains over supervised controls with the same encodings
on both GNN and \gls{gt} backbones.
These results support combining explicit temporal representations
with staged self-supervision for within-database prediction.
Generalization to unseen databases and comparisons of stage orders
under matched computational budgets remain directions for future work.

\subsection*{AI use statement}
Generative AI tools assisted with language editing and polishing. The authors are responsible for the final
content, claims, and artifacts in this work.

\subsection*{Reproducibility statement}
To facilitate reproducibility, \cref{sec:setup} specifies the graph construction
and observation rule, and \cref{sec:method} describes the encodings, objectives,
and training schedule. Appendix~\ref{app:datasets} documents database statistics,
task definitions, and temporal splits; Appendix~\ref{app:detailed-results}
reports task-level metrics and their variation;
Appendix~\ref{app:experimental-config} specifies the architecture, optimization,
and four-seed evaluation settings; Appendix~\ref{app:encoding-details} provides
temporal implementation details; Appendix~\ref{app:objective-details} specifies
the decoders, losses, and sampling details for the pretraining objectives;
and Appendix~\ref{app:training-dynamics}
presents the optimization curves. The appendix will be provided
as a separate PDF in the Supplementary Materials. All code needed to reproduce
the reported results will be provided as supplementary code in the
Supplementary Materials, including instructions for downloading and processing
the publicly available RelBench datasets~\citep{robinson2024relbench} used in our experiments.

\bibliography{references}
\bibliographystyle{arxiv_preprint}

\clearpage
\appendix

% Standalone appendix contents page
\begingroup
\makeatletter

\section*{Contents of the Appendix}

\setcounter{tocdepth}{2}

% Main appendix sections
\newcommand{\appsectionentry}[2]{%
  \addvspace{0.8em}%
  \@dottedtocline{1}{0em}{2.5em}%
    {\numberline{\ref{#1}}#2}{\pageref{#1}}%
}

% Appendix subsections
\newcommand{\appsubsectionentry}[2]{%
  \@dottedtocline{2}{1.5em}{3em}%
    {\numberline{\ref{#1}}#2}{\pageref{#1}}%
}

\appsectionentry{app:datasets}
  {Datasets and Prediction Tasks}
\appsubsectionentry{app:database-statistics}
  {Database Statistics}
\appsubsectionentry{app:task-statistics}
  {Task Statistics and Temporal Splits}
\appsubsectionentry{app:task-descriptions}
  {Database Details and Prediction Targets}

\appsectionentry{app:detailed-results}
  {Detailed Results}
\appsubsectionentry{app:backbone-selection}
  {Backbone Selection: GNNs and GTs}
\appsubsectionentry{app:temporal-representations}
  {Temporal Encoding and Temporal Information Distributions}
\appsubsectionentry{app:pretraining-transfer}
  {Pretraining Detailed Results}

\appsectionentry{app:experimental-config}
  {Architecture and Training Configuration}

\appsectionentry{app:encoding-details}
  {Temporal Encoding Details}

\appsectionentry{app:objective-details}
  {Pretraining Objective Details}

\appsectionentry{app:training-dynamics}
  {Pretraining and Fine-Tuning Curves}

\makeatother
\endgroup
\clearpage

\section{Datasets and Prediction Tasks}
\label{app:datasets}

The five RelBench databases~\citep{robinson2024relbench} provide one
multiclass classification task, six binary classification tasks, and four
regression tasks. The tables below give their sizes; the task definitions
specify prediction windows and eligibility rules.

\subsection{Database Statistics}
\label{app:database-statistics}

\Cref{tab:database-statistics} counts records before temporal filtering,
excluding downstream task tables and reverse edges. Each declared foreign-key
column counts as one directed schema relation, including separate columns
that reference the same table.

\begin{table}[htbp]
\centering
\small
\caption{Stored database statistics. FK denotes foreign-key columns; Tasks counts the tasks evaluated here.}
\label{tab:database-statistics}
\begin{tabular}{llrrrr}
\toprule
Database & Domain & Tables & Rows & \acrshort{fk} relations & Tasks \\
\midrule
Arxiv & Scholarly publications & 6 & 2,733,846 & 6 & 2 \\
Avito & Online advertising & 8 & 24,653,915 & 11 & 3 \\
F1 & Motor racing & 9 & 97,606 & 13 & 2 \\
H\&M & Fashion retail & 3 & 16,931,173 & 2 & 2 \\
Event & Social events & 5 & 44,822,992 & 7 & 2 \\
\bottomrule
\end{tabular}
\end{table}

\subsection{Task Statistics and Temporal Splits}
\label{app:task-statistics}

In \cref{tab:task-statistics}, a sample is an (entity, prediction time)
pair; an entity can therefore appear more than once. ``Entities'' counts
distinct target identifiers across all three splits.

\begin{table}[htbp]
\centering
\small
\setlength{\tabcolsep}{3pt}
\caption{Task sizes in the stored splits. Entities are counted once across training, validation, and test.}
\label{tab:task-statistics}
\begin{tabular}{lllrrrr}
\toprule
Database & Task & Type & Train & Validation & Test & Entities \\
\midrule
Arxiv & author-category & Multiclass classification & 210,769 & 39,015 & 39,655 & 126,219 \\
 & paper-citation & Binary classification & 534,233 & 155,845 & 193,696 & 193,696 \\
\midrule
Avito & user-clicks & Binary classification & 59,454 & 21,183 & 47,996 & 66,449 \\
 & user-visits & Binary classification & 86,619 & 29,979 & 36,129 & 63,405 \\
 & ad-ctr & Regression & 5,100 & 1,766 & 1,816 & 4,997 \\
\midrule
F1 & driver-dnf & Binary classification & 11,411 & 566 & 702 & 821 \\
 & driver-position & Regression & 7,453 & 499 & 760 & 826 \\
\midrule
H\&M & user-churn & Binary classification & 3,832,692 & 76,556 & 74,575 & 999,345 \\
 & item-sales & Regression & 5,488,184 & 105,542 & 105,542 & 105,542 \\
\midrule
Event & user-ignore & Binary classification & 19,239 & 2,013 & 1,958 & 9,694 \\
 & user-attendance & Regression & 19,239 & 2,013 & 1,958 & 9,694 \\
\bottomrule
\end{tabular}
\end{table}

We use the cached chronological splits. Validation and test cutoffs are
2022-01-01 and 2023-01-01 for Arxiv; 2015-05-08 and 2015-05-14 for
Avito; 2020-09-07 and 2020-09-14 for H\&M; and 2012-11-21 and 2012-11-29
for Event. F1 uses several prediction times. For driver-dnf, validation spans
2005-03-02 to 2008-03-16 and test spans 2010-03-02 to 2013-03-16.
For driver-position, the corresponding spans are 2005-03-02 to 2009-10-07
and 2010-03-02 to 2016-05-29. Targets cover
$(t,t+\Delta]$ after cutoff $t$, with the task-specific horizons below.

\subsection{Database Details and Prediction Targets}
\label{app:task-descriptions}

The task definitions follow the task-construction code used in this study.

Arxiv (\texttt{rel-arxiv}) contains 222,769 papers, 143,691 authors,
53 categories, 616,585 authorship rows, 155,061 paper--category rows, and
1,595,687 citation rows. Submission dates timestamp papers and associated
relation records. \texttt{author-category} predicts an author's most frequent
primary research category over the next 182 days, among 53 classes, and
includes only authors who publish in that window. \texttt{paper-citation}
predicts whether a paper submitted by the cutoff receives a citation in
the next 182 days.

Avito (\texttt{rel-avito}) links 5,960,558 advertisements and 98,250 users
to 2,579,289 searches, 9,254,702 search-stream records, 6,454,562 visits,
and 302,974 phone requests, with location and category tables as context.
All three tasks use a four-day prediction window. \texttt{user-clicks}
predicts whether a user records more than one ad click, counting repeated
clicks separately. \texttt{user-visits} predicts visits to more than one
distinct advertisement. \texttt{ad-ctr} predicts clicks divided by
search-stream impressions for advertisements with at least one click
in the window.

F1 (\texttt{rel-f1}) records drivers, constructors, circuits, races,
qualifying sessions, results, and standings. Its nine tables include
857 drivers, 211 constructors, 77 circuits, 1,101 races, and 26,080 race
results; race dates timestamp competition records. \texttt{driver-dnf}
predicts whether a driver has a result with \texttt{statusId} $\ne 1$
in the next 30 days, following the benchmark rule that any status other
than \texttt{Finished} receives a \gls{dnf} label.
\texttt{driver-position} predicts mean finishing order
(\texttt{positionOrder}) over the next 60 days.

H\&M (\texttt{rel-hm}) contains 1,371,980 customers, 105,542 articles,
and 15,453,651 transactions. Dated purchases link customers to articles
and record prices; customer attributes and product descriptions provide
additional features. \texttt{user-churn} includes customers who purchased
in the previous seven days and predicts no purchase in the next seven days.
\texttt{item-sales} predicts the sum of an article's transaction prices
over the next seven days, using the database's price scale and assigning
zero when there are no sales.

Event (\texttt{rel-event}) contains 38,209 users, 3,137,972 events,
30,386,403 friendship rows, 11,245,010 attendance rows, and 15,398 interest
rows. Event start times, user join times, and interest timestamps provide
temporal context. For events starting in the next seven days,
\texttt{user-ignore} predicts whether more than two of a user's attendance
records remain \texttt{invited}, and \texttt{user-attendance} counts that
user's \texttt{yes} or \texttt{maybe} records. These targets describe
recorded responses rather than verified attendance.

\section{Detailed Results}
\label{app:detailed-results}

This section reports the full backbone comparison, temporal encoding
ablation, and pretraining results, together with record-age and horizon
statistics that supplement \cref{sec:main-results}.
% \cref{sec:results}
Task metrics are reported as the mean $\pm$
standard deviation across four independent runs with seeds
$\{42,43,44,45\}$. Relative gains and
their aggregates are computed from these task means.
\Cref{tab:baseline-results} reports validation performance for backbone
selection; the encoding and pretraining comparisons report test performance.

\subsection{Backbone Selection: \texorpdfstring{\acrshortpl{gnn}}{GNNs} and \texorpdfstring{\glspl{gt}}{GTs}}
\label{app:backbone-selection}

\Cref{tab:baseline-results} compares validation performance for five
backbones under the same RDL pipeline: \heterognn{}, RelGNN, HGT, RelGT,
and \method{}
\citep{robinson2024relbench,chen2025relgnn,hu2020hgt,dwivedi2026relgt,peng2026thgfm}.
For each task, we rank all five backbones by their mean validation metric,
using higher accuracy or ROC--AUC and lower MAE as better performance.
We average these ranks equally across all 11 tasks, then select the
backbone with the lowest mean rank within each family.
Among GNNs, \heterognn{} has a lower mean validation rank than RelGNN
(3.18 versus 3.36), with two and three task wins, respectively.
Among \glspl{gt}, \method{} has the lowest mean validation rank (2.27),
followed by HGT (2.55) and RelGT (3.64), with four, two, and zero task
wins, respectively. We therefore select \heterognn{} and \method{} for
the controlled encoding and pretraining study. Task wins are descriptive;
mean validation rank determines the family representatives.

Selecting one representative per family is a deliberate trade-off to
reduce computational cost: running all encoding ablations, pretraining
objectives, and staged schedules on all five backbones would substantially
increase the training budget. The subsequent comparisons therefore
assess effects on these two selected backbones, and do not establish that
the effects extend to RelGNN, HGT, or RelGT.
Backbone selection uses validation performance only; test performance is
reserved for final evaluation. Downstream checkpoints are also selected
using validation metrics, while pretraining checkpoints follow the
objective-specific criteria in Appendix~\ref{app:experimental-config}.

\begin{table*}[t]
\centering
\caption{Backbone validation performance (mean $\pm$ standard deviation)
used to select one representative per family. Ranks are computed across
all five models for each task and averaged equally over all 11 tasks.
Bold marks the best task mean, the highest task-win count, and the lowest
mean rank within each family; arrows indicate the preferred direction.}
\label{tab:baseline-results}
\scriptsize
\setlength{\tabcolsep}{2.8pt}
\resizebox{\textwidth}{!}{%
\begin{tabular}{llccccc}
\toprule
& & \multicolumn{2}{c}{\acrshort{gnn}} & \multicolumn{3}{c}{\acrshort{gt}} \\
\cmidrule(lr){3-4}
\cmidrule(lr){5-7}
Task & Metric & \textsc{HeteroGNN} & RelGNN & HGT & RelGT & \acrshort{thgfm} \\
\midrule
arxiv/author-category & Acc.$\uparrow$ & $0.1311{\pm}0.0020$ & $0.1298{\pm}0.0010$ & $\mathbf{0.1331{\pm}0.0017}$ & $0.1037{\pm}0.0050$ & $0.1324{\pm}0.0027$ \\
arxiv/paper-citation & \glsdisp{rocauc}{AUC}$\uparrow$ & $0.6495{\pm}0.0069$ & $0.6216{\pm}0.0095$ & $0.6583{\pm}0.0074$ & $0.6550{\pm}0.0027$ & $\mathbf{0.6652{\pm}0.0019}$ \\
avito/user-clicks & \glsdisp{rocauc}{AUC}$\uparrow$ & $0.6435{\pm}0.0021$ & $\mathbf{0.6496{\pm}0.0336}$ & $0.6369{\pm}0.0032$ & $0.6298{\pm}0.0185$ & $0.6405{\pm}0.0008$ \\
avito/user-visits & \glsdisp{rocauc}{AUC}$\uparrow$ & $\mathbf{0.6938{\pm}0.0012}$ & $0.6735{\pm}0.0014$ & $0.6763{\pm}0.0095$ & $0.6899{\pm}0.0186$ & $0.6830{\pm}0.0075$ \\
f1/driver-dnf & \glsdisp{rocauc}{AUC}$\uparrow$ & $0.7442{\pm}0.0042$ & $\mathbf{0.8020{\pm}0.0042}$ & $0.7978{\pm}0.0064$ & $0.7830{\pm}0.0117$ & $0.7999{\pm}0.0068$ \\
hm/user-churn & \glsdisp{rocauc}{AUC}$\uparrow$ & $0.7025{\pm}0.0005$ & $0.7011{\pm}0.0108$ & $0.7042{\pm}0.0006$ & $0.6970{\pm}0.0012$ & $\mathbf{0.7064{\pm}0.0015}$ \\
event/user-ignore & \glsdisp{rocauc}{AUC}$\uparrow$ & $0.8570{\pm}0.0022$ & $0.8618{\pm}0.0275$ & $0.8638{\pm}0.0275$ & $0.8593{\pm}0.0297$ & $\mathbf{0.8661{\pm}0.0094}$ \\
hm/item-sales & \acrshort{mae}$\downarrow$ & $0.0656{\pm}0.0003$ & $\mathbf{0.0640{\pm}0.0007}$ & $0.0692{\pm}0.0021$ & $0.0662{\pm}0.0004$ & $0.0670{\pm}0.0002$ \\
avito/ad-ctr & \acrshort{mae}$\downarrow$ & $\mathbf{0.0355{\pm}0.0004}$ & $0.0398{\pm}0.0000$ & $0.0368{\pm}0.0012$ & $0.0372{\pm}0.0014$ & $0.0380{\pm}0.0000$ \\
f1/driver-position & \acrshort{mae}$\downarrow$ & $3.8399{\pm}0.0078$ & $3.8703{\pm}0.0079$ & $\mathbf{3.0344{\pm}0.2518}$ & $3.1994{\pm}0.0656$ & $3.5500{\pm}0.0034$ \\
event/user-attendance & \acrshort{mae}$\downarrow$ & $0.2570{\pm}0.0006$ & $0.2490{\pm}0.0015$ & $0.2470{\pm}0.0001$ & $0.2530{\pm}0.0015$ & $\mathbf{0.2420{\pm}0.0022}$ \\
\midrule
\multicolumn{2}{l}{No. of task-wise best results $\uparrow$}
& $2$ & $3$ & $2$ & $0$ & $\mathbf{4}$ \\
\multicolumn{2}{l}{Mean rank $\downarrow$}
& $\mathbf{3.18}$ & $3.36$ & $2.55$ & $3.64$ & $\mathbf{2.27}$ \\
\bottomrule
\end{tabular}%
}
\end{table*}

\subsection{Temporal Encoding and Temporal Information Distributions}
\label{app:temporal-representations}

\paragraph{Reference encoding and configurations.}
As introduced in \cref{sec:main-results}, \textsc{DayPE}
(day-based positional encoding) is the fixed day-based single-scale
record-age encoding used by \heterognn{} \citep{robinson2024relbench}.
It serves as the baseline for multi-scale \mix{} on both backbones
and is defined as
\begin{equation}
    \label{eq:daype}
    e_v^{\mathrm{base}}
    =
    W_{\phi(v)}
    \operatorname{PE}\!\left(a_v/s_{\mathrm{day}}\right)
    + b_{\phi(v)},
    \qquad
    s_{\mathrm{day}}=86400\ \mathrm{seconds}.
\end{equation}
Here $a_v$ is measured in seconds,
$\operatorname{PE}$ is sinusoidal positional encoding with
fixed frequencies, and $W_{\phi(v)}$ and $b_{\phi(v)}$ are
type-specific trainable parameters.
The term \emph{single-scale} refers to the fixed day-based normalization;
the sinusoidal positional encoding itself contains multiple frequencies.
Thus, ``without our encodings'' retains \textsc{DayPE} and disables
\rotm{}; it still provides record-age information.
In contrast, \mix{} uses multiple learned time scales and mixes
type-specific projections of log-transformed record ages
(\cref{sec:temporal-encoding}; Appendix~\ref{app:time-mix-details}).

\begin{table*}[t]
\centering
\caption{Temporal encoding ablation (mean $\pm$ standard deviation over four runs).
Gains are relative to the same backbone with \textsc{DayPE}
(\cref{eq:daype}) and without \rotm{} or pretraining in each panel.
\mix{} replaces \textsc{DayPE}; \rotm{} adds rotary interval encoding
to the specified record-age encoder. Mean type gain weights the three
task types equally.
Bold marks the best entry per panel and column.}
\label{tab:temporal-representations}
\label{tab:graphsage-temporal}
\label{tab:drsf-temporal}
\tiny
\setlength{\tabcolsep}{1.2pt}
\renewcommand{\arraystretch}{1.05}
\resizebox{\textwidth}{!}{%
\begin{tabular}{l*{15}{c}}
\toprule
& \multicolumn{2}{c}{Multiclass classification}
& \multicolumn{7}{c}{Binary classification}
& \multicolumn{5}{c}{Regression} & \\
\cmidrule(lr){2-3}\cmidrule(lr){4-10}\cmidrule(lr){11-15}
Encoding & Author & Mean task gain & Citation & Ignore & \acrshort{dnf} & Churn & Clicks & Visits
& Mean task gain & Attend. & Position & Sales & \acrshort{ctr} & Mean task gain & Mean type gain \\
& Acc.$\uparrow$ & $\Delta\%\uparrow$ & \glsdisp{rocauc}{AUC}$\uparrow$ & \glsdisp{rocauc}{AUC}$\uparrow$ & \glsdisp{rocauc}{AUC}$\uparrow$
& \glsdisp{rocauc}{AUC}$\uparrow$ & \glsdisp{rocauc}{AUC}$\uparrow$ & \glsdisp{rocauc}{AUC}$\uparrow$ & $\Delta\%\uparrow$ & \acrshort{mae}$\downarrow$
& \acrshort{mae}$\downarrow$ & \acrshort{mae}$\downarrow$ & \acrshort{mae}$\downarrow$ & $\Delta\%\uparrow$ & $\Delta\%\uparrow$ \\
\midrule
\multicolumn{16}{l}{\textbf{Panel A: \heterognn{} (\acrshort{gnn} backbone)}} \\
\addlinespace[1pt]
\textsc{DayPE} (reference)
& $0.1236{\pm}0.0005$ & $+0.00\%$
& $0.6450{\pm}0.0047$ & $0.8031{\pm}0.0157$ & $0.7101{\pm}0.0105$
& $\mathbf{0.6961{\pm}0.0019}$ & $\mathbf{0.6474{\pm}0.0164}$ & $\mathbf{0.6627{\pm}0.0008}$ & $\mathbf{+0.00\%}$
& $0.2620{\pm}0.0013$ & $4.3596{\pm}0.0168$ & $\mathbf{0.0604{\pm}0.0001}$ & $\mathbf{0.0403{\pm}0.0003}$ & $\mathbf{+0.00\%}$ & $+0.00\%$ \\
\mix{}
& $\mathbf{0.1265{\pm}0.0028}$ & $\mathbf{+2.35\%}$
& $0.6430{\pm}0.0043$ & $0.8083{\pm}0.0095$ & $\mathbf{0.7316{\pm}0.0043}$
& $0.6868{\pm}0.0028$ & $0.6282{\pm}0.0042$ & $0.6355{\pm}0.0090$ & $-0.84\%$
& $0.2577{\pm}0.0083$ & $4.3178{\pm}0.0382$ & $0.0638{\pm}0.0015$ & $0.0408{\pm}0.0006$ & $-0.98\%$ & $+0.18\%$ \\
\textsc{DayPE}+\rotm{}
& $0.1239{\pm}0.0017$ & $+0.24\%$
& $\mathbf{0.6527{\pm}0.0036}$ & $0.8202{\pm}0.0051$ & $0.7227{\pm}0.0099$
& $0.6896{\pm}0.0052$ & $0.6318{\pm}0.0022$ & $0.6454{\pm}0.0050$ & $-0.14\%$
& $0.2616{\pm}0.0021$ & $4.1803{\pm}0.1225$ & $0.0617{\pm}0.0006$ & $0.0421{\pm}0.0001$ & $-0.49\%$ & $-0.13\%$ \\
\mix{}+\rotm{}
& $0.1261{\pm}0.0026$ & $+2.02\%$
& $0.6469{\pm}0.0031$ & $\mathbf{0.8789{\pm}0.0073}$ & $0.6967{\pm}0.0231$
& $0.6920{\pm}0.0011$ & $0.6174{\pm}0.0070$ & $0.6269{\pm}0.0080$ & $-0.46\%$
& $\mathbf{0.2566{\pm}0.0090}$ & $\mathbf{4.0603{\pm}0.0988}$ & $0.0660{\pm}0.0002$ & $0.0410{\pm}0.0005$ & $-0.18\%$ & $\mathbf{+0.46\%}$ \\
\midrule
\multicolumn{16}{l}{\textbf{Panel B: \method{} (\gls{gt} backbone)}} \\
\addlinespace[1pt]
\textsc{DayPE} (reference)
& $0.1235{\pm}0.0020$ & $+0.00\%$
& $\mathbf{0.6603{\pm}0.0012}$ & $0.8194{\pm}0.0326$ & $\mathbf{0.7396{\pm}0.0068}$
& $\mathbf{0.6969{\pm}0.0041}$ & $\mathbf{0.6346{\pm}0.0023}$ & $0.6369{\pm}0.0033$ & $\mathbf{+0.00\%}$
& $0.2439{\pm}0.0009$ & $4.3693{\pm}0.0132$ & $\mathbf{0.0613{\pm}0.0006}$ & $0.0431{\pm}0.0001$ & $+0.00\%$ & $+0.00\%$ \\
\mix{}
& $\mathbf{0.1266{\pm}0.0034}$ & $\mathbf{+2.51\%}$
& $0.6482{\pm}0.0035$ & $0.8308{\pm}0.0195$ & $0.7210{\pm}0.0051$
& $0.6913{\pm}0.0016$ & $0.6195{\pm}0.0021$ & $0.6269{\pm}0.0117$ & $-1.28\%$
& $0.2422{\pm}0.0039$ & $\mathbf{4.2098{\pm}0.1526}$ & $0.0639{\pm}0.0017$ & $0.0412{\pm}0.0006$ & $+1.26\%$ & $+0.83\%$ \\
\textsc{DayPE}+\rotm{}
& $0.1232{\pm}0.0022$ & $-0.24\%$
& $0.6543{\pm}0.0019$ & $0.8196{\pm}0.0250$ & $0.7213{\pm}0.0137$
& $0.6926{\pm}0.0020$ & $0.6311{\pm}0.0019$ & $\mathbf{0.6465{\pm}0.0036}$ & $-0.50\%$
& $\mathbf{0.2415{\pm}0.0016}$ & $4.3080{\pm}0.1579$ & $0.0616{\pm}0.0004$ & $0.0421{\pm}0.0000$ & $+1.08\%$ & $+0.11\%$ \\
\mix{}+\rotm{}
& $0.1265{\pm}0.0045$ & $+2.43\%$
& $0.6515{\pm}0.0053$ & $\mathbf{0.8393{\pm}0.0167}$ & $0.7255{\pm}0.0204$
& $0.6940{\pm}0.0007$ & $0.6321{\pm}0.0077$ & $0.6454{\pm}0.0118$ & $-0.05\%$
& $0.2456{\pm}0.0039$ & $4.2606{\pm}0.2064$ & $0.0621{\pm}0.0025$ & $\mathbf{0.0409{\pm}0.0010}$ & $\mathbf{+1.49\%}$ & $\mathbf{+1.29\%}$ \\
\bottomrule
\end{tabular}}
\end{table*}

\paragraph{Aggregate and task-level effects.}
Relative to \textsc{DayPE}, replacing the age encoder with \mix{}
yields mean type gains of $+0.18\%$ on \heterognn{} and $+0.83\%$
on \method{}, while adding \rotm{} to \textsc{DayPE} yields
$-0.13\%$ and $+0.11\%$, respectively.
Combining \mix{} and \rotm{} gives the highest mean type gain on
both backbones, $+0.46\%$ and $+1.29\%$.
Accordingly, \cref{fig:temporal-encoding-ablation}a shows mean type gain
differences of $-0.59$, $-0.28$, and $-0.46$ percentage points on
\heterognn{}, and $-1.18$, $-0.46$, and $-1.29$ points on \method{},
for \textsc{DayPE}+\rotm{}, \mix{}, and \textsc{DayPE}, respectively.
The combined gains by task type are $+2.02\%$, $-0.46\%$, and
$-0.18\%$ on \heterognn{}, and $+2.43\%$, $-0.05\%$, and
$+1.49\%$ on \method{}, for multiclass classification, binary
classification, and regression, respectively.
Thus, the positive overall gains do not imply improvements in every
task type. In particular, \mix{} alone gives slightly higher multiclass
gains than the combined configuration on both backbones, while the
combination gives higher regression gains than either encoding change alone.

Task-level effects also vary.
With both encodings, \heterognn{} improves author classification by
$2.02\%$, user-ignore AUC by $9.44\%$, and MAE by $7.37\%$ for
driver position and $2.10\%$ for user attendance, but loses performance on
both Avito binary tasks and item sales. On \method{}, the combined
encodings improve ad-CTR MAE by $5.38\%$ and also help author
classification, user visits, user ignore, and driver position; citation
and driver-dnf performance decrease.

\begin{figure}[t]
  \centering
  \includegraphics[width=\textwidth]{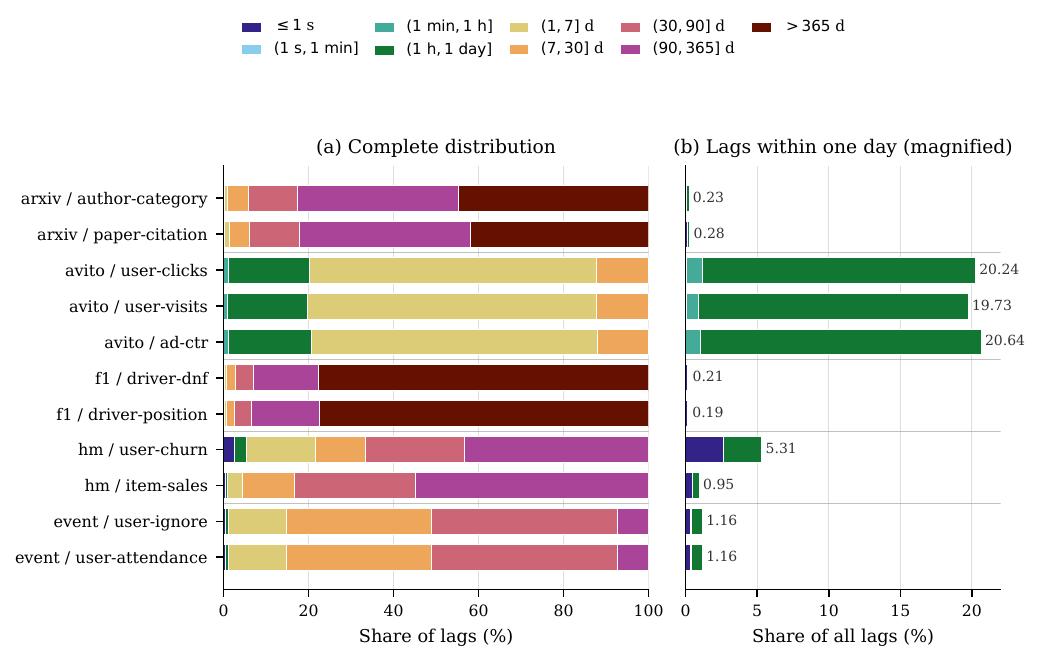}
  \caption{Record-age distributions by task: (a) all nine intervals and (b) a magnified view of ages up to one day, on the original percentage scale.}
  \label{fig:reference-node-lag-distribution}
\end{figure}

\paragraph{Distribution of record ages.}
\Cref{fig:reference-node-lag-distribution} summarizes the ages of records
available to each task. We first deduplicate the training (entity, cutoff)
pairs and sample up to 20,000 pairs per task using seed 42. For each pair,
we collect the seed row, if timestamped, and directly linked timestamped
rows observed by the cutoff. We convert timestamps to integer Unix seconds
and compute record ages as defined in \cref{sec:temporal-encoding}.
These statistics cover the seed and its direct neighbors; the model
samples neighborhoods that may extend to additional hops.
We group ages into nine bins with boundaries at 1 second, 1 minute,
1 hour, 1 day, and 7, 30, 90, and 365 days. The first four boundaries
match the initial \mix{} scales. Panel (b) enlarges the bins up to
one day while retaining their percentages of all observations;
bar-end labels show their combined share.

The observed histories span different time scales across databases.
In Avito, all sampled record ages fall within 30 days, and
$19.7$--$20.6\%$ of the records are at most one day old. Avito also shows variation
at second, minute, and hour scales; minute-scale variation is largely
absent from the other databases under integer-second preprocessing.
Event records are concentrated between 1 and 90 days, while H\&M
records mostly fall between 7 and 365 days. Arxiv and F1 contain much
older records: fewer than $0.3\%$ are at most one day old, and about
$78\%$ of F1 records are more than a year old. These differences
provide descriptive motivation for the multi-scale design of \mix{}.
The distributions alone do not establish its predictive advantage
over a single-scale variant or determine the optimal number or
initialization of scales.
The encoding ablations evaluate \mix{} as a complete replacement
for \textsc{DayPE}; a controlled comparison of
$K_{\mathrm{time}}=1$ and $K_{\mathrm{time}}>1$ within the same
\mix{} formulation remains to be evaluated.

\begin{figure*}[t]
  \centering
  \includegraphics[width=\textwidth]{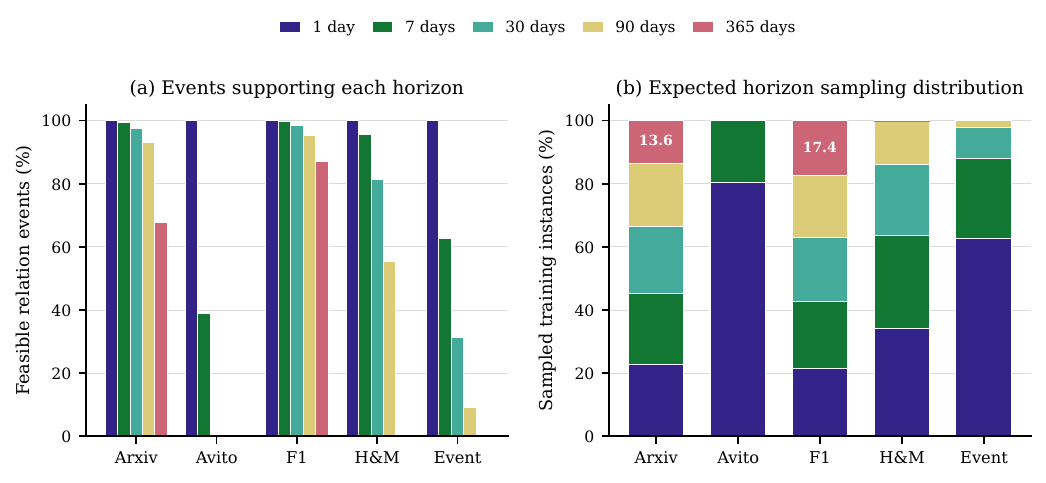}
  \caption{Support for \efuture{} horizons: (a) the relation-balanced percentage of eligible events supporting each horizon and (b) the resulting horizon sampling distribution.}
  \label{fig:rel-future-horizon-distribution}
\end{figure*}

\paragraph{Support for future horizons.}
\Cref{fig:rel-future-horizon-distribution} shows which of the candidate
horizons $\{1,7,30,90,365\}$ days can be used for \efuture{}.
A horizon is eligible when the entity is already observed at the
input cutoff, as specified in \cref{eq:future-horizon-eligibility}.
We compute these statistics from timestamped forward foreign-key
relations up to each database's validation cutoff, excluding static
relations and events with no eligible horizon. Panel (a) reports the
percentage of eligible events supporting each horizon, computed within
each relation and then averaged equally across relations. An event may
support several horizons. Panel (b) shows the expected share of training
examples assigned to each horizon when we uniformly sample a relation,
an eligible event within it, and an eligible horizon for that event.

Support for longer horizons varies substantially across databases.
Avito supports only 1- and 7-day windows. In Event, support falls from
$62.6\%$ at 7 days to $31.3\%$ at 30 days and $9.1\%$ at 90 days,
with no support at 365 days. H\&M supports windows through 90 days,
but only $0.05\%$ of eligible events support a 365-day window.
By comparison, the 365-day horizon is eligible for $67.7\%$ of Arxiv
and $87.1\%$ of F1 events under the same averaging scheme.
A shared set of multiple horizons therefore provides short-window
supervision in databases with limited history and longer-window
supervision where the history permits it.

The sampling distribution reflects these differences. In Avito,
$80.5\%$ of training examples are expected to use the 1-day horizon
and $19.5\%$ the 7-day horizon; unsupported horizons receive no samples.
Thus, using the same candidate set does not imply sampling horizons
equally across databases. These statistics describe the availability
and frequency of supervision, rather than the predictive value of
each horizon.

\subsection{Pretraining Detailed Results}
\label{app:pretraining-transfer}

\Cref{tab:pretraining-heterognn,tab:pretraining-drsf} provide the full
results underlying the pretraining comparisons, including task metrics
and results for \gsubgraph{} alone. Each panel uses a supervised control
with the same backbone and temporal encoding configuration:
A0 uses \textsc{DayPE} without \rotm{}, whereas B0 replaces
\textsc{DayPE} with \mix{} and enables \rotm{}.
Both controls are trained directly on downstream tasks without pretraining.
The reported gains therefore measure the additional effect
of pretraining, matching \cref{fig:pretraining-staging-ablation}b.
The main-results heatmap in \cref{fig:pretraining-gains} instead uses
A0 throughout, measuring the effects of pretraining and, where enabled,
our temporal encodings.
Rows with \gsubgraph{} followed by a second-stage objective can be compared
with either objective alone. Each panel contains 20 pretraining strategies
and one supervised control: nine single-stage baselines, our three
single-stage objectives, and eight two-stage schedules. The latter comprise
TS followed by \nattr{}, \nschema{}, PT-HGNN, CTRL, TGPM, TVE,
\ehist{}, or \efuture{}.
The six baseline pairs and our two proposed pairs are shown in
\cref{fig:pretraining-staging-ablation}b.

Mean type gain weights the multiclass, binary, and regression task types
equally, using the gain definition in
Appendix~\ref{app:temporal-representations}. We aggregate before rounding
and compute differences from the unrounded gains.
The average ranks in \cref{fig:pretraining-gains} rank all 20 pretraining
strategies within each backbone--encoding setting, excluding the supervised
control. Tied task means receive the average of their ranks, and the
11 task ranks are weighted equally.

\begin{table*}[t]
\centering
\caption{Pretraining transfer on \heterognn{} (mean $\pm$ standard deviation over four runs). Panel A uses \textsc{DayPE} without \rotm{}; Panel B uses \mix{} with \rotm{}. Gains are relative to the marked reference within each panel. Bold marks the best value per panel and column.}
\label{tab:pretraining-heterognn}
\tiny
\setlength{\tabcolsep}{1.2pt}
\renewcommand{\arraystretch}{1.05}
\resizebox{\textwidth}{!}{%
\begin{tabular}{lll*{14}{c}c}
\toprule
& \multicolumn{2}{c}{Pretraining strategy} & \multicolumn{2}{c}{Multiclass classification} & \multicolumn{7}{c}{Binary classification} & \multicolumn{5}{c}{Regression} & \\
\cmidrule(lr){2-3}\cmidrule(lr){4-5}\cmidrule(lr){6-12}\cmidrule(lr){13-17}
Temporal & Graph-level & Local stage & Author & Task gain & Citation & Ignore & \acrshort{dnf} & Churn & Clicks & Visits & Mean task gain & Attend. & Position & Sales & \acrshort{ctr} & Mean task gain & Mean type gain \\
& & & Acc.$\uparrow$ & $\Delta\%\uparrow$ & \glsdisp{rocauc}{AUC}$\uparrow$ & \glsdisp{rocauc}{AUC}$\uparrow$ & \glsdisp{rocauc}{AUC}$\uparrow$ & \glsdisp{rocauc}{AUC}$\uparrow$ & \glsdisp{rocauc}{AUC}$\uparrow$ & \glsdisp{rocauc}{AUC}$\uparrow$ & $\Delta\%\uparrow$ & \acrshort{mae}$\downarrow$ & \acrshort{mae}$\downarrow$ & \acrshort{mae}$\downarrow$ & \acrshort{mae}$\downarrow$ & $\Delta\%\uparrow$ & $\Delta\%\uparrow$ \\
\midrule
\multicolumn{18}{l}{\textbf{Panel A: \textsc{DayPE} without \rotm{}.} All $\Delta$ columns use Ref. A0 below.} \\
\addlinespace[1pt]
w/o & \multicolumn{2}{l}{No pretraining (\textbf{Ref. A0})} & $0.1236{\pm}0.0005$ & $+0.00\%$ & $0.6450{\pm}0.0047$ & $0.8031{\pm}0.0157$ & $0.7101{\pm}0.0105$ & $0.6961{\pm}0.0019$ & $0.6474{\pm}0.0164$ & $0.6627{\pm}0.0008$ & $+0.00\%$ & $0.2620{\pm}0.0013$ & $4.3596{\pm}0.0168$ & $0.0604{\pm}0.0001$ & $0.0403{\pm}0.0003$ & $+0.00\%$ & $+0.00\%$ \\
\cmidrule(lr){1-18}
\multicolumn{18}{l}{\textbf{Baseline pretraining strategies}} \\
w/o & -- & Un-SAGE & $0.1209{\pm}0.0039$ & $-2.18\%$ & $0.5023{\pm}0.0053$ & $0.8206{\pm}0.0286$ & $0.7038{\pm}0.0284$ & $0.6902{\pm}0.0011$ & $0.6482{\pm}0.0096$ & $0.6570{\pm}0.0032$ & $-3.74\%$ & $0.2632{\pm}0.0059$ & $4.3955{\pm}0.0202$ & $0.0565{\pm}0.0008$ & $\mathbf{0.0362{\pm}0.0002}$ & $+4.24\%$ & $-0.56\%$ \\
w/o & -- & GPT-GNN & $0.1203{\pm}0.0008$ & $-2.67\%$ & $0.6226{\pm}0.0174$ & $\mathbf{0.8658{\pm}0.0139}$ & $0.7007{\pm}0.0012$ & $0.6895{\pm}0.0008$ & $0.6470{\pm}0.0123$ & $0.6582{\pm}0.0042$ & $+0.22\%$ & $0.2655{\pm}0.0016$ & $4.3884{\pm}0.0273$ & $0.0566{\pm}0.0018$ & $0.0393{\pm}0.0012$ & $+1.82\%$ & $-0.21\%$ \\
w/o & -- & PT-HGNN & $0.1176{\pm}0.0060$ & $-4.85\%$ & $0.6494{\pm}0.0580$ & $0.8182{\pm}0.0549$ & $\mathbf{0.7589{\pm}0.0009}$ & $0.6971{\pm}0.0043$ & $0.6579{\pm}0.0087$ & $0.6625{\pm}0.0096$ & $\mathbf{+1.86\%}$ & $0.2638{\pm}0.0029$ & $4.3516{\pm}0.1148$ & $0.0557{\pm}0.0007$ & $0.0376{\pm}0.0006$ & $+3.78\%$ & $+0.26\%$ \\
w/o & -- & PHE & $0.1225{\pm}0.0010$ & $-0.89\%$ & $0.6412{\pm}0.0014$ & $0.8540{\pm}0.0200$ & $0.6801{\pm}0.0204$ & $0.6948{\pm}0.0010$ & $0.6323{\pm}0.0049$ & $0.6548{\pm}0.0063$ & $-0.36\%$ & $0.2635{\pm}0.0092$ & $5.3604{\pm}0.0053$ & $0.0568{\pm}0.0001$ & $0.0383{\pm}0.0004$ & $-1.92\%$ & $-1.06\%$ \\
w/o & -- & \nattr{} & $0.1214{\pm}0.0019$ & $-1.78\%$ & $0.6382{\pm}0.0207$ & $0.8352{\pm}0.0131$ & $0.7103{\pm}0.0008$ & $\mathbf{0.6998{\pm}0.0017}$ & $0.6424{\pm}0.0077$ & $0.6591{\pm}0.0057$ & $+0.36\%$ & $0.2618{\pm}0.0079$ & $4.2791{\pm}0.1757$ & $0.0558{\pm}0.0011$ & $0.0380{\pm}0.0009$ & $+4.06\%$ & $+0.88\%$ \\
w/o & -- & \nschema{} & $0.1240{\pm}0.0017$ & $+0.32\%$ & $0.5399{\pm}0.0744$ & $0.8274{\pm}0.0219$ & $0.6675{\pm}0.0370$ & $0.6962{\pm}0.0031$ & $0.6386{\pm}0.0117$ & $0.6538{\pm}0.0010$ & $-3.66\%$ & $0.2626{\pm}0.0045$ & $\mathbf{3.9744{\pm}0.0422}$ & $0.0563{\pm}0.0008$ & $0.0455{\pm}0.0001$ & $+1.33\%$ & $-0.67\%$ \\
w/o & -- & CTRL & $0.1238{\pm}0.0025$ & $+0.16\%$ & $0.6441{\pm}0.0161$ & $0.8124{\pm}0.0193$ & $0.6969{\pm}0.0026$ & $0.6961{\pm}0.0007$ & $0.6534{\pm}0.0140$ & $0.6594{\pm}0.0029$ & $-0.07\%$ & $0.2548{\pm}0.0056$ & $4.4945{\pm}0.0180$ & $0.0560{\pm}0.0015$ & $0.0396{\pm}0.0011$ & $+2.36\%$ & $+0.82\%$ \\
w/o & -- & TGPM & $0.1223{\pm}0.0019$ & $-1.05\%$ & $0.6436{\pm}0.0036$ & $0.6870{\pm}0.0220$ & $0.7217{\pm}0.0030$ & $0.6957{\pm}0.0025$ & $0.6536{\pm}0.0071$ & $0.6623{\pm}0.0047$ & $-2.03\%$ & $0.2578{\pm}0.0038$ & $4.2298{\pm}0.0307$ & $0.0563{\pm}0.0020$ & $0.0404{\pm}0.0003$ & $+2.93\%$ & $-0.05\%$ \\
w/o & -- & TVE & $0.1240{\pm}0.0007$ & $+0.32\%$ & $0.6344{\pm}0.0110$ & $0.8189{\pm}0.0083$ & $0.7081{\pm}0.0126$ & $0.6953{\pm}0.0045$ & $0.6460{\pm}0.0095$ & $0.6554{\pm}0.0029$ & $-0.23\%$ & $0.2582{\pm}0.0012$ & $4.3304{\pm}0.0951$ & $0.0561{\pm}0.0007$ & $0.0392{\pm}0.0002$ & $+3.15\%$ & $+1.08\%$ \\
\cmidrule(lr){1-18}
\multicolumn{18}{l}{\textbf{Our pretraining strategies}} \\
w/o & -- & \ehist{} & $\mathbf{0.1260{\pm}0.0019}$ & $\mathbf{+1.94\%}$ & $0.6494{\pm}0.0011$ & $0.7798{\pm}0.0276$ & $0.6951{\pm}0.0052$ & $0.6891{\pm}0.0044$ & $0.6542{\pm}0.0067$ & $0.6632{\pm}0.0026$ & $-0.70\%$ & $0.2634{\pm}0.0099$ & $4.3409{\pm}0.0113$ & $0.0559{\pm}0.0008$ & $0.0373{\pm}0.0017$ & $+4.00\%$ & $+1.75\%$ \\
w/o & -- & \efuture{} & $0.1233{\pm}0.0027$ & $-0.24\%$ & $0.6410{\pm}0.0060$ & $0.8464{\pm}0.0332$ & $0.7027{\pm}0.0451$ & $0.6951{\pm}0.0039$ & $0.6469{\pm}0.0159$ & $0.6613{\pm}0.0009$ & $+0.55\%$ & $0.2543{\pm}0.0118$ & $4.3560{\pm}0.0052$ & $0.0556{\pm}0.0005$ & $0.0391{\pm}0.0009$ & $+3.70\%$ & $+1.34\%$ \\
\midrule
w/o & \gsubgraph{} & -- & $0.1178{\pm}0.0029$ & $-4.69\%$ & $\mathbf{0.6576{\pm}0.0065}$ & $0.7961{\pm}0.0426$ & $0.6861{\pm}0.0146$ & $0.6963{\pm}0.0037$ & $0.6405{\pm}0.0101$ & $0.6597{\pm}0.0026$ & $-0.63\%$ & $0.2660{\pm}0.0059$ & $4.1975{\pm}0.1987$ & $0.0556{\pm}0.0002$ & $0.0404{\pm}0.0009$ & $+2.69\%$ & $-0.88\%$ \\
w/o & \gsubgraph{} & \nattr{} & $0.1232{\pm}0.0037$ & $-0.32\%$ & $0.6461{\pm}0.0060$ & $0.8227{\pm}0.0694$ & $0.6834{\pm}0.0108$ & $0.6982{\pm}0.0037$ & $0.6600{\pm}0.0122$ & $0.6607{\pm}0.0008$ & $+0.13\%$ & $0.2548{\pm}0.0048$ & $4.0122{\pm}0.1722$ & $0.0560{\pm}0.0009$ & $0.0443{\pm}0.0005$ & $+2.58\%$ & $+0.80\%$ \\
w/o & \gsubgraph{} & \nschema{} & $0.1241{\pm}0.0022$ & $+0.40\%$ & $0.6366{\pm}0.0021$ & $0.7895{\pm}0.0456$ & $0.6821{\pm}0.0280$ & $0.6976{\pm}0.0054$ & $0.6511{\pm}0.0163$ & $0.6536{\pm}0.0035$ & $-1.25\%$ & $0.2583{\pm}0.0056$ & $4.1695{\pm}0.2314$ & $0.0560{\pm}0.0001$ & $0.0495{\pm}0.0011$ & $-1.18\%$ & $-0.68\%$ \\
w/o & \gsubgraph{} & PT-HGNN & $0.1208{\pm}0.0037$ & $-2.27\%$ & $0.6535{\pm}0.0133$ & $0.8122{\pm}0.0448$ & $0.7377{\pm}0.0020$ & $0.6985{\pm}0.0021$ & $0.6579{\pm}0.0099$ & $0.6635{\pm}0.0044$ & $+1.40\%$ & $0.2542{\pm}0.0028$ & $4.1614{\pm}0.0564$ & $0.0554{\pm}0.0005$ & $0.0386{\pm}0.0005$ & $+5.32\%$ & $+1.48\%$ \\
w/o & \gsubgraph{} & CTRL & $0.1235{\pm}0.0009$ & $-0.08\%$ & $0.6462{\pm}0.0009$ & $0.8027{\pm}0.0096$ & $0.7037{\pm}0.0091$ & $0.6977{\pm}0.0030$ & $0.6565{\pm}0.0111$ & $0.6594{\pm}0.0021$ & $+0.06\%$ & $0.2516{\pm}0.0028$ & $4.3849{\pm}0.1361$ & $0.0558{\pm}0.0002$ & $0.0374{\pm}0.0002$ & $+4.89\%$ & $+1.62\%$ \\
w/o & \gsubgraph{} & TGPM & $0.1220{\pm}0.0054$ & $-1.29\%$ & $0.6460{\pm}0.0088$ & $0.6776{\pm}0.0353$ & $0.7287{\pm}0.0204$ & $0.6976{\pm}0.0017$ & $0.6571{\pm}0.0073$ & $0.6625{\pm}0.0050$ & $-1.86\%$ & $0.2544{\pm}0.0032$ & $4.1520{\pm}0.0724$ & $0.0557{\pm}0.0004$ & $0.0375{\pm}0.0002$ & $+5.97\%$ & $+0.94\%$ \\
w/o & \gsubgraph{} & TVE & $0.1230{\pm}0.0040$ & $-0.49\%$ & $0.6327{\pm}0.0015$ & $0.8044{\pm}0.0308$ & $0.7106{\pm}0.0096$ & $0.6928{\pm}0.0054$ & $0.6452{\pm}0.0057$ & $0.6514{\pm}0.0018$ & $-0.70\%$ & $0.2563{\pm}0.0045$ & $4.2533{\pm}0.0597$ & $0.0560{\pm}0.0011$ & $0.0369{\pm}0.0007$ & $+5.45\%$ & $+1.42\%$ \\
\hdashline 
w/o & \gsubgraph{} & \ehist{} & $0.1242{\pm}0.0009$ & $+0.49\%$ & $0.6455{\pm}0.0052$ & $0.7633{\pm}0.0244$ & $0.7162{\pm}0.0225$ & $0.6937{\pm}0.0060$ & $\mathbf{0.6610{\pm}0.0070}$ & $0.6596{\pm}0.0009$ & $-0.46\%$ & $0.2527{\pm}0.0092$ & $4.1328{\pm}0.1100$ & $0.0561{\pm}0.0012$ & $0.0376{\pm}0.0004$ & $+6.00\%$ & $+2.01\%$ \\
w/o & \gsubgraph{} & \efuture{} & $0.1234{\pm}0.0035$ & $-0.16\%$ & $0.6569{\pm}0.0032$ & $0.8073{\pm}0.0369$ & $0.7203{\pm}0.0062$ & $0.6997{\pm}0.0017$ & $0.6579{\pm}0.0143$ & $\mathbf{0.6644{\pm}0.0032}$ & $+1.03\%$ & $\mathbf{0.2464{\pm}0.0035}$ & $4.0057{\pm}0.0239$ & $\mathbf{0.0552{\pm}0.0002}$ & $0.0395{\pm}0.0004$ & $\mathbf{+6.65\%}$ & $\mathbf{+2.51\%}$ \\
\midrule
\multicolumn{18}{l}{\textbf{Panel B: \mix{} with \rotm{}.} All $\Delta$ columns use Ref. B0 below.} \\
\addlinespace[1pt]
w/ & \multicolumn{2}{l}{No pretraining (\textbf{Ref. B0})} & $0.1261{\pm}0.0026$ & $+0.00\%$ & $0.6469{\pm}0.0031$ & $\mathbf{0.8789{\pm}0.0073}$ & $0.6967{\pm}0.0231$ & $0.6920{\pm}0.0011$ & $0.6174{\pm}0.0070$ & $0.6269{\pm}0.0080$ & $+0.00\%$ & $0.2566{\pm}0.0090$ & $4.0603{\pm}0.0988$ & $0.0660{\pm}0.0002$ & $0.0410{\pm}0.0005$ & $+0.00\%$ & $+0.00\%$ \\
\cmidrule(lr){1-18}
\multicolumn{18}{l}{\textbf{Baseline pretraining strategies}} \\
w/ & -- & Un-SAGE & $0.1235{\pm}0.0016$ & $-2.06\%$ & $0.5011{\pm}0.0504$ & $0.8058{\pm}0.0046$ & $0.6734{\pm}0.0012$ & $0.6756{\pm}0.0018$ & $0.6538{\pm}0.0088$ & $0.6625{\pm}0.0025$ & $-4.17\%$ & $0.2637{\pm}0.0079$ & $4.3595{\pm}0.0482$ & $0.0565{\pm}0.0004$ & $0.0413{\pm}0.0003$ & $+1.63\%$ & $-1.53\%$ \\
w/ & -- & GPT-GNN & $0.1279{\pm}0.0013$ & $+1.43\%$ & $0.6172{\pm}0.0047$ & $0.8540{\pm}0.0089$ & $0.6867{\pm}0.0013$ & $0.6893{\pm}0.0023$ & $0.6484{\pm}0.0076$ & $\mathbf{0.6638{\pm}0.0075}$ & $+0.28\%$ & $0.2645{\pm}0.0032$ & $4.5646{\pm}0.0259$ & $0.0564{\pm}0.0019$ & $0.0503{\pm}0.0001$ & $-3.88\%$ & $-0.72\%$ \\
w/ & -- & PT-HGNN & $0.1238{\pm}0.0013$ & $-1.82\%$ & $0.6081{\pm}0.0052$ & $0.8047{\pm}0.0524$ & $0.6864{\pm}0.0009$ & $0.6900{\pm}0.0015$ & $0.6537{\pm}0.0130$ & $0.6614{\pm}0.0028$ & $-0.80\%$ & $0.2637{\pm}0.0035$ & $4.3801{\pm}0.0386$ & $\mathbf{0.0556{\pm}0.0006}$ & $0.0379{\pm}0.0002$ & $+4.22\%$ & $+0.53\%$ \\
w/ & -- & PHE & $0.1208{\pm}0.0042$ & $-4.20\%$ & $0.6133{\pm}0.0045$ & $0.8185{\pm}0.0410$ & $\mathbf{0.7335{\pm}0.0109}$ & $0.6857{\pm}0.0022$ & $0.6588{\pm}0.0026$ & $0.6581{\pm}0.0006$ & $+0.66\%$ & $0.2637{\pm}0.0016$ & $5.1706{\pm}0.2662$ & $0.0564{\pm}0.0020$ & $0.0372{\pm}0.0004$ & $+0.77\%$ & $-0.92\%$ \\
w/ & -- & \nattr{} & $0.1289{\pm}0.0019$ & $+2.22\%$ & $0.6579{\pm}0.0197$ & $0.8280{\pm}0.0140$ & $0.6355{\pm}0.0007$ & $0.6936{\pm}0.0012$ & $0.6367{\pm}0.0059$ & $0.6588{\pm}0.0115$ & $-0.74\%$ & $0.2504{\pm}0.0081$ & $\mathbf{3.9469{\pm}0.1745}$ & $0.0565{\pm}0.0011$ & $0.0396{\pm}0.0014$ & $+6.42\%$ & $+2.64\%$ \\
w/ & -- & \nschema{} & $0.1294{\pm}0.0019$ & $+2.62\%$ & $0.6465{\pm}0.0713$ & $0.8219{\pm}0.0223$ & $0.6635{\pm}0.0334$ & $0.6971{\pm}0.0023$ & $0.6471{\pm}0.0082$ & $0.6507{\pm}0.0034$ & $-0.33\%$ & $0.2502{\pm}0.0048$ & $4.0847{\pm}0.0396$ & $0.0565{\pm}0.0009$ & $0.0502{\pm}0.0002$ & $+0.11\%$ & $+0.80\%$ \\
w/ & -- & CTRL & $0.1282{\pm}0.0022$ & $+1.67\%$ & $0.6499{\pm}0.0209$ & $0.8625{\pm}0.0086$ & $0.7162{\pm}0.0078$ & $0.7043{\pm}0.0035$ & $0.6395{\pm}0.0078$ & $0.6470{\pm}0.0131$ & $+1.66\%$ & $0.2487{\pm}0.0044$ & $4.8580{\pm}0.0187$ & $0.0563{\pm}0.0012$ & $0.0437{\pm}0.0004$ & $-0.55\%$ & $+0.93\%$ \\
w/ & -- & TGPM & $0.1274{\pm}0.0022$ & $+1.03\%$ & $0.6290{\pm}0.0099$ & $0.8344{\pm}0.0117$ & $0.6974{\pm}0.0077$ & $0.6854{\pm}0.0022$ & $0.6320{\pm}0.0048$ & $0.6387{\pm}0.0052$ & $-0.74\%$ & $0.2505{\pm}0.0036$ & $4.3612{\pm}0.0073$ & $0.0561{\pm}0.0003$ & $0.0394{\pm}0.0008$ & $+4.31\%$ & $+1.53\%$ \\
w/ & -- & TVE & $0.1277{\pm}0.0016$ & $+1.27\%$ & $0.6299{\pm}0.0070$ & $0.8446{\pm}0.0040$ & $0.6856{\pm}0.0039$ & $\mathbf{0.7049{\pm}0.0035}$ & $0.6457{\pm}0.0048$ & $0.6507{\pm}0.0093$ & $+0.35\%$ & $0.2504{\pm}0.0100$ & $4.3869{\pm}0.0354$ & $0.0560{\pm}0.0003$ & $0.0423{\pm}0.0016$ & $+2.45\%$ & $+1.36\%$ \\
\cmidrule(lr){1-18}
\multicolumn{18}{l}{\textbf{Our pretraining strategies}} \\
w/ & -- & \ehist{} & $0.1228{\pm}0.0022$ & $-2.62\%$ & $0.6434{\pm}0.0012$ & $0.8149{\pm}0.0313$ & $0.7141{\pm}0.0058$ & $0.6925{\pm}0.0058$ & $0.6682{\pm}0.0090$ & $0.6637{\pm}0.0006$ & $+1.47\%$ & $0.2577{\pm}0.0102$ & $4.3594{\pm}0.0104$ & $0.0561{\pm}0.0008$ & $0.0380{\pm}0.0011$ & $+4.56\%$ & $+1.14\%$ \\
w/ & -- & \efuture{} & $0.1275{\pm}0.0026$ & $+1.11\%$ & $0.6267{\pm}0.0065$ & $0.8229{\pm}0.0336$ & $0.6652{\pm}0.0425$ & $0.6967{\pm}0.0029$ & $0.6515{\pm}0.0108$ & $0.6623{\pm}0.0031$ & $-0.36\%$ & $\mathbf{0.2415{\pm}0.0112}$ & $4.3637{\pm}0.0048$ & $0.0557{\pm}0.0005$ & $\mathbf{0.0356{\pm}0.0013}$ & $\mathbf{+8.24\%}$ & $+3.00\%$ \\
\midrule
w/ & \gsubgraph{} & -- & $0.1265{\pm}0.0033$ & $+0.32\%$ & $0.6513{\pm}0.0062$ & $0.7440{\pm}0.0438$ & $0.7074{\pm}0.0159$ & $0.6918{\pm}0.0028$ & $0.6343{\pm}0.0073$ & $0.6615{\pm}0.0062$ & $-0.82\%$ & $0.2518{\pm}0.0064$ & $4.2311{\pm}0.2199$ & $0.0560{\pm}0.0002$ & $0.0468{\pm}0.0014$ & $+0.83\%$ & $+0.11\%$ \\
w/ & \gsubgraph{} & \nattr{} & $0.1277{\pm}0.0042$ & $+1.27\%$ & $\mathbf{0.6596{\pm}0.0064}$ & $0.8144{\pm}0.0681$ & $0.6604{\pm}0.0101$ & $0.6955{\pm}0.0028$ & $0.6508{\pm}0.0086$ & $0.6613{\pm}0.0030$ & $+0.14\%$ & $0.2517{\pm}0.0054$ & $4.1791{\pm}0.1681$ & $0.0563{\pm}0.0009$ & $0.0393{\pm}0.0009$ & $+5.16\%$ & $+2.19\%$ \\
w/ & \gsubgraph{} & \nschema{} & $\mathbf{0.1316{\pm}0.0025}$ & $\mathbf{+4.36\%}$ & $0.6096{\pm}0.0022$ & $0.8515{\pm}0.0449$ & $0.6770{\pm}0.0264$ & $0.6947{\pm}0.0040$ & $0.6511{\pm}0.0110$ & $0.6521{\pm}0.0078$ & $-0.31\%$ & $0.2526{\pm}0.0063$ & $4.0034{\pm}0.2255$ & $0.0563{\pm}0.0002$ & $0.0486{\pm}0.0017$ & $+1.15\%$ & $+1.73\%$ \\
w/ & \gsubgraph{} & PT-HGNN & $0.1260{\pm}0.0009$ & $-0.08\%$ & $0.6261{\pm}0.0057$ & $0.8118{\pm}0.0343$ & $0.7095{\pm}0.0050$ & $0.6927{\pm}0.0029$ & $0.6642{\pm}0.0076$ & $0.6611{\pm}0.0023$ & $+0.69\%$ & $0.2560{\pm}0.0048$ & $4.3282{\pm}0.0624$ & $0.0558{\pm}0.0010$ & $0.0380{\pm}0.0003$ & $+5.05\%$ & $+1.89\%$ \\
w/ & \gsubgraph{} & CTRL & $0.1290{\pm}0.0016$ & $+2.30\%$ & $0.6535{\pm}0.0056$ & $0.8576{\pm}0.0278$ & $0.7222{\pm}0.0041$ & $0.7039{\pm}0.0008$ & $0.6416{\pm}0.0076$ & $0.6461{\pm}0.0034$ & $+1.83\%$ & $0.2486{\pm}0.0037$ & $4.7789{\pm}0.1797$ & $0.0561{\pm}0.0002$ & $0.0372{\pm}0.0001$ & $+4.01\%$ & $+2.71\%$ \\
w/ & \gsubgraph{} & TGPM & $0.1281{\pm}0.0014$ & $+1.59\%$ & $0.6330{\pm}0.0063$ & $0.8303{\pm}0.0075$ & $0.7039{\pm}0.0016$ & $0.6857{\pm}0.0015$ & $0.6346{\pm}0.0072$ & $0.6384{\pm}0.0018$ & $-0.49\%$ & $0.2502{\pm}0.0095$ & $4.2937{\pm}0.0887$ & $0.0560{\pm}0.0003$ & $0.0377{\pm}0.0008$ & $+5.93\%$ & $+2.34\%$ \\
w/ & \gsubgraph{} & TVE & $0.1275{\pm}0.0060$ & $+1.11\%$ & $0.6294{\pm}0.0056$ & $0.8342{\pm}0.0260$ & $0.6872{\pm}0.0214$ & $0.7003{\pm}0.0011$ & $0.6439{\pm}0.0097$ & $0.6458{\pm}0.0069$ & $-0.11\%$ & $0.2519{\pm}0.0047$ & $4.3515{\pm}0.1175$ & $0.0559{\pm}0.0006$ & $0.0375{\pm}0.0003$ & $+5.64\%$ & $+2.22\%$ \\
\hdashline
w/ & \gsubgraph{} & \ehist{} & $0.1278{\pm}0.0010$ & $+1.35\%$ & $0.6409{\pm}0.0055$ & $0.8176{\pm}0.0239$ & $0.7284{\pm}0.0212$ & $0.6950{\pm}0.0045$ & $\mathbf{0.6728{\pm}0.0055}$ & $0.6609{\pm}0.0033$ & $\mathbf{+1.91\%}$ & $0.2497{\pm}0.0103$ & $4.2858{\pm}0.1076$ & $0.0559{\pm}0.0012$ & $0.0381{\pm}0.0006$ & $+5.80\%$ & $\mathbf{+3.02\%}$ \\
w/ & \gsubgraph{} & \efuture{} & $0.1267{\pm}0.0040$ & $+0.48\%$ & $0.6522{\pm}0.0034$ & $0.7962{\pm}0.0363$ & $0.6880{\pm}0.0059$ & $0.6958{\pm}0.0012$ & $0.6616{\pm}0.0098$ & $0.6630{\pm}0.0071$ & $+0.60\%$ & $0.2472{\pm}0.0039$ & $4.0695{\pm}0.0234$ & $0.0560{\pm}0.0003$ & $0.0379{\pm}0.0007$ & $+7.40\%$ & $+2.83\%$ \\
\bottomrule
\end{tabular}}
\end{table*}

\begin{table*}[t]
\centering
\caption{Pretraining transfer on \method{} (mean $\pm$ standard deviation over four runs). Panel A uses \textsc{DayPE} without \rotm{}; Panel B uses \mix{} with \rotm{}. Gains are relative to the marked reference within each panel. Bold marks the best value per panel and column.}
\label{tab:pretraining-drsf}
\tiny
\setlength{\tabcolsep}{1.2pt}
\renewcommand{\arraystretch}{1.05}
\resizebox{\textwidth}{!}{%
\begin{tabular}{lll*{14}{c}c}
\toprule
& \multicolumn{2}{c}{Pretraining strategy} & \multicolumn{2}{c}{Multiclass classification} & \multicolumn{7}{c}{Binary classification} & \multicolumn{5}{c}{Regression} & \\
\cmidrule(lr){2-3}\cmidrule(lr){4-5}\cmidrule(lr){6-12}\cmidrule(lr){13-17}
Temporal & Graph-level & Local stage & Author & Task gain & Citation & Ignore & \acrshort{dnf} & Churn & Clicks & Visits & Mean task gain & Attend. & Position & Sales & \acrshort{ctr} & Mean task gain & Mean type gain \\
& & & Acc.$\uparrow$ & $\Delta\%\uparrow$ & \glsdisp{rocauc}{AUC}$\uparrow$ & \glsdisp{rocauc}{AUC}$\uparrow$ & \glsdisp{rocauc}{AUC}$\uparrow$ & \glsdisp{rocauc}{AUC}$\uparrow$ & \glsdisp{rocauc}{AUC}$\uparrow$ & \glsdisp{rocauc}{AUC}$\uparrow$ & $\Delta\%\uparrow$ & \acrshort{mae}$\downarrow$ & \acrshort{mae}$\downarrow$ & \acrshort{mae}$\downarrow$ & \acrshort{mae}$\downarrow$ & $\Delta\%\uparrow$ & $\Delta\%\uparrow$ \\
\midrule
\multicolumn{18}{l}{\textbf{Panel A: \textsc{DayPE} without \rotm{}.} All $\Delta$ columns use Ref. A0 below.} \\
\addlinespace[1pt]
w/o & \multicolumn{2}{l}{No pretraining (\textbf{Ref. A0})} & $\mathbf{0.1235{\pm}0.0020}$ & $\mathbf{+0.00\%}$ & $0.6603{\pm}0.0012$ & $0.8194{\pm}0.0326$ & $0.7396{\pm}0.0068$ & $0.6969{\pm}0.0041$ & $0.6346{\pm}0.0023$ & $0.6369{\pm}0.0033$ & $+0.00\%$ & $0.2439{\pm}0.0009$ & $4.3693{\pm}0.0132$ & $0.0613{\pm}0.0006$ & $0.0431{\pm}0.0001$ & $+0.00\%$ & $+0.00\%$ \\
\cmidrule(lr){1-18}
\multicolumn{18}{l}{\textbf{Baseline pretraining strategies}} \\
w/o & -- & Un-SAGE & $0.1194{\pm}0.0009$ & $-3.32\%$ & $\mathbf{0.6644{\pm}0.0054}$ & $0.8631{\pm}0.0352$ & $0.7052{\pm}0.0348$ & $0.6944{\pm}0.0014$ & $0.6297{\pm}0.0087$ & $0.6228{\pm}0.0046$ & $-0.34\%$ & $0.2635{\pm}0.0032$ & $4.6955{\pm}0.0240$ & $0.0581{\pm}0.0005$ & $0.0429{\pm}0.0002$ & $-2.10\%$ & $-1.92\%$ \\
w/o & -- & GPT-GNN & $0.1197{\pm}0.0015$ & $-3.08\%$ & $0.6568{\pm}0.0058$ & $0.7367{\pm}0.0405$ & $0.6558{\pm}0.0045$ & $0.6962{\pm}0.0022$ & $0.6513{\pm}0.0040$ & $0.6455{\pm}0.0229$ & $-3.01\%$ & $0.2635{\pm}0.0093$ & $4.3773{\pm}0.2472$ & $0.0640{\pm}0.0002$ & $0.0445{\pm}0.0024$ & $-3.75\%$ & $-3.28\%$ \\
w/o & -- & PT-HGNN & $0.1234{\pm}0.0011$ & $-0.08\%$ & $0.6493{\pm}0.0333$ & $\mathbf{0.8698{\pm}0.0129}$ & $\mathbf{0.7563{\pm}0.0010}$ & $0.6998{\pm}0.0018$ & $\mathbf{0.6737{\pm}0.0024}$ & $0.6370{\pm}0.0009$ & $\mathbf{+2.22\%}$ & $0.2631{\pm}0.0111$ & $4.3530{\pm}0.0658$ & $0.0597{\pm}0.0003$ & $0.0407{\pm}0.0005$ & $+0.41\%$ & $+0.85\%$ \\
w/o & -- & PHE & $0.1177{\pm}0.0009$ & $-4.70\%$ & $0.6464{\pm}0.0305$ & $0.8376{\pm}0.0057$ & $0.7219{\pm}0.0496$ & $0.6965{\pm}0.0013$ & $0.6273{\pm}0.0037$ & $0.6514{\pm}0.0017$ & $-0.20\%$ & $0.2636{\pm}0.0070$ & $5.8698{\pm}0.0560$ & $0.0646{\pm}0.0003$ & $0.0422{\pm}0.0011$ & $-9.00\%$ & $-4.63\%$ \\
w/o & -- & \nattr{} & $0.1203{\pm}0.0045$ & $-2.59\%$ & $0.6502{\pm}0.0094$ & $0.8027{\pm}0.0640$ & $0.6809{\pm}0.0151$ & $0.6939{\pm}0.0023$ & $0.6414{\pm}0.0029$ & $0.6577{\pm}0.0014$ & $-1.27\%$ & $0.2583{\pm}0.0101$ & $4.4562{\pm}0.0466$ & $0.0588{\pm}0.0007$ & $0.0429{\pm}0.0003$ & $-0.70\%$ & $-1.52\%$ \\
w/o & -- & \nschema{} & $0.1169{\pm}0.0014$ & $-5.34\%$ & $0.6506{\pm}0.0086$ & $0.7907{\pm}0.0327$ & $0.6480{\pm}0.0095$ & $0.7007{\pm}0.0043$ & $0.6482{\pm}0.0105$ & $0.6525{\pm}0.0029$ & $-2.04\%$ & $0.2608{\pm}0.0108$ & $4.3993{\pm}0.2041$ & $0.0579{\pm}0.0001$ & $0.0499{\pm}0.0004$ & $-3.73\%$ & $-3.70\%$ \\
w/o & -- & CTRL & $0.1213{\pm}0.0008$ & $-1.78\%$ & $0.6440{\pm}0.0244$ & $0.7958{\pm}0.0104$ & $0.6822{\pm}0.0186$ & $0.6766{\pm}0.0011$ & $0.6446{\pm}0.0043$ & $0.6385{\pm}0.0022$ & $-2.37\%$ & $0.2496{\pm}0.0014$ & $4.2604{\pm}0.0554$ & $0.0605{\pm}0.0009$ & $0.0408{\pm}0.0003$ & $+1.81\%$ & $-0.78\%$ \\
w/o & -- & TGPM & $0.1218{\pm}0.0054$ & $-1.38\%$ & $0.6375{\pm}0.0093$ & $0.8017{\pm}0.0071$ & $0.6760{\pm}0.0101$ & $0.6755{\pm}0.0011$ & $0.6342{\pm}0.0030$ & $0.6299{\pm}0.0085$ & $-3.07\%$ & $0.2455{\pm}0.0093$ & $4.2300{\pm}0.1372$ & $0.0606{\pm}0.0007$ & $0.0424{\pm}0.0002$ & $+1.36\%$ & $-1.03\%$ \\
w/o & -- & TVE & $0.1157{\pm}0.0018$ & $-6.32\%$ & $0.6409{\pm}0.0023$ & $0.7701{\pm}0.0312$ & $0.7451{\pm}0.0014$ & $0.6962{\pm}0.0049$ & $0.6436{\pm}0.0029$ & $0.6437{\pm}0.0070$ & $-0.97\%$ & $0.2545{\pm}0.0069$ & $4.2856{\pm}0.0534$ & $0.0576{\pm}0.0006$ & $0.0415{\pm}0.0004$ & $+2.02\%$ & $-1.76\%$ \\
\cmidrule(lr){1-18}
\multicolumn{18}{l}{\textbf{Our pretraining strategies}} \\
w/o & -- & \ehist{} & $0.1225{\pm}0.0008$ & $-0.81\%$ & $0.6533{\pm}0.0042$ & $0.7932{\pm}0.0032$ & $0.7118{\pm}0.0556$ & $0.6997{\pm}0.0024$ & $0.6263{\pm}0.0070$ & $0.6459{\pm}0.0030$ & $-1.25\%$ & $0.2423{\pm}0.0048$ & $4.4595{\pm}0.0945$ & $0.0605{\pm}0.0006$ & $0.0409{\pm}0.0010$ & $+1.33\%$ & $-0.24\%$ \\
w/o & -- & \efuture{} & $0.1233{\pm}0.0044$ & $-0.16\%$ & $0.6452{\pm}0.0010$ & $0.8203{\pm}0.0141$ & $0.6751{\pm}0.0407$ & $0.7020{\pm}0.0032$ & $0.6619{\pm}0.0062$ & $0.6540{\pm}0.0016$ & $-0.53\%$ & $0.2447{\pm}0.0079$ & $4.3661{\pm}0.0148$ & $0.0612{\pm}0.0007$ & $\mathbf{0.0376{\pm}0.0003}$ & $+3.63\%$ & $+0.98\%$ \\
\midrule
w/o & \gsubgraph{} & -- & $0.1194{\pm}0.0043$ & $-3.32\%$ & $0.6156{\pm}0.0559$ & $0.7996{\pm}0.0211$ & $0.6857{\pm}0.0345$ & $0.6973{\pm}0.0036$ & $0.6203{\pm}0.0171$ & $0.6463{\pm}0.0028$ & $-2.87\%$ & $0.2474{\pm}0.0004$ & $4.3392{\pm}0.1328$ & $0.0569{\pm}0.0001$ & $0.0475{\pm}0.0007$ & $-0.56\%$ & $-2.25\%$ \\
w/o & \gsubgraph{} & \nattr{} & $0.1212{\pm}0.0051$ & $-1.86\%$ & $0.6462{\pm}0.0107$ & $0.8095{\pm}0.0162$ & $0.6872{\pm}0.0202$ & $0.7010{\pm}0.0030$ & $0.6411{\pm}0.0063$ & $0.6523{\pm}0.0032$ & $-1.07\%$ & $0.2503{\pm}0.0063$ & $4.3736{\pm}0.0351$ & $\mathbf{0.0563{\pm}0.0010}$ & $0.0480{\pm}0.0008$ & $-1.00\%$ & $-1.31\%$ \\
w/o & \gsubgraph{} & \nschema{} & $0.1138{\pm}0.0067$ & $-7.85\%$ & $0.5573{\pm}0.0814$ & $0.7770{\pm}0.0256$ & $0.6519{\pm}0.0326$ & $0.6977{\pm}0.0055$ & $0.6652{\pm}0.0164$ & $0.6486{\pm}0.0045$ & $-4.31\%$ & $0.2618{\pm}0.0038$ & $4.2555{\pm}0.3372$ & $0.0567{\pm}0.0009$ & $0.0509{\pm}0.0002$ & $-2.84\%$ & $-5.00\%$ \\
w/o & \gsubgraph{} & PT-HGNN & $0.1230{\pm}0.0013$ & $-0.40\%$ & $0.6481{\pm}0.0122$ & $0.8465{\pm}0.0180$ & $0.7330{\pm}0.0055$ & $0.6991{\pm}0.0010$ & $0.6660{\pm}0.0030$ & $0.6437{\pm}0.0028$ & $+1.15\%$ & $0.2521{\pm}0.0049$ & $4.2452{\pm}0.0648$ & $0.0583{\pm}0.0005$ & $0.0406{\pm}0.0003$ & $+2.74\%$ & $+1.16\%$ \\
w/o & \gsubgraph{} & CTRL & $0.1212{\pm}0.0020$ & $-1.86\%$ & $0.6440{\pm}0.0147$ & $0.8011{\pm}0.0275$ & $0.6841{\pm}0.0188$ & $0.6767{\pm}0.0029$ & $0.6491{\pm}0.0108$ & $0.6404{\pm}0.0077$ & $-2.05\%$ & $0.2492{\pm}0.0015$ & $4.1967{\pm}0.0299$ & $0.0568{\pm}0.0013$ & $0.0406{\pm}0.0024$ & $+4.02\%$ & $+0.04\%$ \\
w/o & \gsubgraph{} & TGPM & $0.1209{\pm}0.0031$ & $-2.11\%$ & $0.6326{\pm}0.0055$ & $0.8008{\pm}0.0367$ & $0.6723{\pm}0.0294$ & $0.6702{\pm}0.0009$ & $0.6339{\pm}0.0063$ & $0.6270{\pm}0.0010$ & $-3.51\%$ & $0.2469{\pm}0.0029$ & $4.2224{\pm}0.0673$ & $0.0607{\pm}0.0016$ & $0.0384{\pm}0.0004$ & $+3.87\%$ & $-0.58\%$ \\
w/o & \gsubgraph{} & TVE & $0.1157{\pm}0.0034$ & $-6.32\%$ & $0.6407{\pm}0.0184$ & $0.7751{\pm}0.0377$ & $0.7467{\pm}0.0043$ & $0.6960{\pm}0.0022$ & $0.6479{\pm}0.0045$ & $0.6454{\pm}0.0130$ & $-0.69\%$ & $0.2541{\pm}0.0039$ & $4.2228{\pm}0.0664$ & $0.0581{\pm}0.0005$ & $0.0389{\pm}0.0001$ & $+3.94\%$ & $-1.02\%$ \\
\hdashline 
w/o & \gsubgraph{} & \ehist{} & $0.1227{\pm}0.0028$ & $-0.65\%$ & $0.6468{\pm}0.0033$ & $0.8231{\pm}0.0233$ & $0.7097{\pm}0.0248$ & $0.6985{\pm}0.0007$ & $0.6582{\pm}0.0050$ & $0.6503{\pm}0.0094$ & $+0.07\%$ & $\mathbf{0.2412{\pm}0.0020}$ & $\mathbf{4.1374{\pm}0.0560}$ & $0.0568{\pm}0.0007$ & $0.0404{\pm}0.0002$ & $\mathbf{+5.33\%}$ & $\mathbf{+1.58\%}$ \\
w/o & \gsubgraph{} & \efuture{} & $0.1228{\pm}0.0016$ & $-0.57\%$ & $0.6506{\pm}0.0021$ & $0.8195{\pm}0.0130$ & $0.6862{\pm}0.0092$ & $\mathbf{0.7022{\pm}0.0054}$ & $0.6546{\pm}0.0071$ & $\mathbf{0.6593{\pm}0.0265}$ & $-0.21\%$ & $0.2440{\pm}0.0029$ & $4.3201{\pm}0.1085$ & $0.0575{\pm}0.0002$ & $0.0383{\pm}0.0053$ & $+5.06\%$ & $+1.43\%$ \\
\midrule
\multicolumn{18}{l}{\textbf{Panel B: \mix{} with \rotm{}.} All $\Delta$ columns use Ref. B0 below.} \\
\addlinespace[1pt]
w/ & \multicolumn{2}{l}{No pretraining (\textbf{Ref. B0})} & $\mathbf{0.1265{\pm}0.0045}$ & $\mathbf{+0.00\%}$ & $0.6515{\pm}0.0053$ & $0.8393{\pm}0.0167$ & $0.7255{\pm}0.0204$ & $0.6940{\pm}0.0007$ & $0.6321{\pm}0.0077$ & $0.6454{\pm}0.0118$ & $+0.00\%$ & $0.2456{\pm}0.0039$ & $4.2606{\pm}0.2064$ & $0.0621{\pm}0.0025$ & $0.0409{\pm}0.0011$ & $+0.00\%$ & $+0.00\%$ \\
\cmidrule(lr){1-18}
\multicolumn{18}{l}{\textbf{Baseline pretraining strategies}} \\
w/ & -- & Un-SAGE & $0.1239{\pm}0.0045$ & $-2.06\%$ & $0.5074{\pm}0.0059$ & $0.8037{\pm}0.0219$ & $0.6043{\pm}0.0038$ & $0.6831{\pm}0.0008$ & $0.6334{\pm}0.0026$ & $0.6223{\pm}0.0066$ & $-8.00\%$ & $0.2635{\pm}0.0102$ & $4.7512{\pm}0.0520$ & $0.0761{\pm}0.0005$ & $0.0431{\pm}0.0003$ & $-10.16\%$ & $-6.74\%$ \\
w/ & -- & GPT-GNN & $0.1209{\pm}0.0032$ & $-4.43\%$ & $0.6382{\pm}0.0027$ & $0.7958{\pm}0.0165$ & $0.6893{\pm}0.0031$ & $0.6936{\pm}0.0038$ & $0.6595{\pm}0.0043$ & $0.6471{\pm}0.0153$ & $-1.28\%$ & $0.2635{\pm}0.0017$ & $4.5845{\pm}0.1405$ & $0.0658{\pm}0.0001$ & $0.0444{\pm}0.0004$ & $-6.84\%$ & $-4.18\%$ \\
w/ & -- & PT-HGNN & $0.1218{\pm}0.0016$ & $-3.72\%$ & $0.6479{\pm}0.0038$ & $0.8189{\pm}0.0086$ & $0.6813{\pm}0.0078$ & $0.6991{\pm}0.0013$ & $0.6559{\pm}0.0023$ & $0.6475{\pm}0.0179$ & $-0.71\%$ & $0.2626{\pm}0.0100$ & $4.4162{\pm}0.0216$ & $0.0575{\pm}0.0021$ & $0.0416{\pm}0.0008$ & $-0.92\%$ & $-1.78\%$ \\
w/ & -- & PHE & $0.1149{\pm}0.0012$ & $-9.17\%$ & $\mathbf{0.6720{\pm}0.0050}$ & $0.8079{\pm}0.0354$ & $0.7147{\pm}0.0022$ & $0.6957{\pm}0.0011$ & $0.6361{\pm}0.0029$ & $0.6431{\pm}0.0252$ & $-0.26\%$ & $0.2634{\pm}0.0110$ & $4.7078{\pm}0.3323$ & $0.0633{\pm}0.0011$ & $0.0409{\pm}0.0004$ & $-4.54\%$ & $-4.66\%$ \\
w/ & -- & \nattr{} & $0.1129{\pm}0.0045$ & $-10.75\%$ & $0.6664{\pm}0.0100$ & $0.7494{\pm}0.0622$ & $0.6539{\pm}0.0134$ & $0.6966{\pm}0.0017$ & $0.6346{\pm}0.0076$ & $0.6402{\pm}0.0064$ & $-3.05\%$ & $0.2442{\pm}0.0090$ & $4.7057{\pm}0.0524$ & $0.0633{\pm}0.0007$ & $0.0432{\pm}0.0006$ & $-4.03\%$ & $-5.94\%$ \\
w/ & -- & \nschema{} & $0.1059{\pm}0.0014$ & $-16.28\%$ & $0.6611{\pm}0.0087$ & $0.7912{\pm}0.0371$ & $0.6764{\pm}0.0098$ & $0.6977{\pm}0.0032$ & $0.6521{\pm}0.0115$ & $0.6530{\pm}0.0056$ & $-1.03\%$ & $0.2617{\pm}0.0096$ & $4.1297{\pm}0.2134$ & $0.0601{\pm}0.0002$ & $0.0500{\pm}0.0006$ & $-4.46\%$ & $-7.26\%$ \\
w/ & -- & CTRL & $0.1209{\pm}0.0036$ & $-4.43\%$ & $0.6613{\pm}0.0018$ & $0.7951{\pm}0.0220$ & $0.7286{\pm}0.0055$ & $0.6884{\pm}0.0040$ & $0.6563{\pm}0.0062$ & $0.6600{\pm}0.0012$ & $+0.32\%$ & $0.2488{\pm}0.0100$ & $4.0945{\pm}0.1010$ & $0.0641{\pm}0.0002$ & $0.0414{\pm}0.0007$ & $-0.39\%$ & $-1.50\%$ \\
w/ & -- & TGPM & $0.1212{\pm}0.0057$ & $-4.19\%$ & $0.6643{\pm}0.0128$ & $\mathbf{0.8518{\pm}0.0151}$ & $0.7378{\pm}0.0237$ & $0.6789{\pm}0.0016$ & $0.6453{\pm}0.0049$ & $0.6557{\pm}0.0039$ & $+1.11\%$ & $0.2423{\pm}0.0062$ & $4.2733{\pm}0.1752$ & $0.0636{\pm}0.0001$ & $0.0421{\pm}0.0011$ & $-1.04\%$ & $-1.37\%$ \\
w/ & -- & TVE & $0.1228{\pm}0.0013$ & $-2.92\%$ & $0.6440{\pm}0.0046$ & $0.8362{\pm}0.0214$ & $0.7233{\pm}0.0038$ & $0.6868{\pm}0.0015$ & $0.6299{\pm}0.0040$ & $0.6464{\pm}0.0072$ & $-0.51\%$ & $0.2467{\pm}0.0045$ & $4.1943{\pm}0.0682$ & $0.0577{\pm}0.0009$ & $0.0415{\pm}0.0008$ & $+1.83\%$ & $-0.54\%$ \\
\cmidrule(lr){1-18}
\multicolumn{18}{l}{\textbf{Our pretraining strategies}} \\
w/ & -- & \ehist{} & $0.1223{\pm}0.0008$ & $-3.32\%$ & $0.6613{\pm}0.0047$ & $0.8495{\pm}0.0036$ & $0.7092{\pm}0.0480$ & $0.6999{\pm}0.0018$ & $\mathbf{0.6637{\pm}0.0050}$ & $0.6504{\pm}0.0041$ & $\mathbf{+1.18\%}$ & $\mathbf{0.2373{\pm}0.0051}$ & $4.3395{\pm}0.1002$ & $0.0657{\pm}0.0007$ & $0.0393{\pm}0.0016$ & $+0.07\%$ & $-0.69\%$ \\
w/ & -- & \efuture{} & $0.1230{\pm}0.0041$ & $-2.77\%$ & $0.6308{\pm}0.0009$ & $0.8137{\pm}0.0145$ & $0.7240{\pm}0.0424$ & $\mathbf{0.7019{\pm}0.0024}$ & $0.6536{\pm}0.0030$ & $\mathbf{0.6607{\pm}0.0066}$ & $+0.08\%$ & $0.2407{\pm}0.0085$ & $4.3989{\pm}0.0157$ & $0.0655{\pm}0.0008$ & $0.0369{\pm}0.0006$ & $+1.14\%$ & $-0.52\%$ \\
\midrule
w/ & \gsubgraph{} & -- & $0.1210{\pm}0.0047$ & $-4.35\%$ & $0.6704{\pm}0.0532$ & $0.8109{\pm}0.0193$ & $0.7107{\pm}0.0366$ & $0.6924{\pm}0.0027$ & $0.6570{\pm}0.0108$ & $0.6535{\pm}0.0071$ & $+0.41\%$ & $0.2434{\pm}0.0004$ & $4.2107{\pm}0.1301$ & $0.0573{\pm}0.0002$ & $0.0455{\pm}0.0011$ & $+0.09\%$ & $-1.28\%$ \\
w/ & \gsubgraph{} & \nattr{} & $0.1109{\pm}0.0053$ & $-12.33\%$ & $0.6168{\pm}0.0102$ & $0.8457{\pm}0.0174$ & $0.7030{\pm}0.0182$ & $0.6963{\pm}0.0022$ & $0.6565{\pm}0.0067$ & $0.6523{\pm}0.0027$ & $-0.40\%$ & $0.2470{\pm}0.0071$ & $4.5481{\pm}0.0305$ & $0.0576{\pm}0.0010$ & $0.0435{\pm}0.0004$ & $-1.26\%$ & $-4.67\%$ \\
w/ & \gsubgraph{} & \nschema{} & $0.1169{\pm}0.0069$ & $-7.59\%$ & $0.6650{\pm}0.0771$ & $0.8162{\pm}0.0274$ & $0.6885{\pm}0.0297$ & $0.6927{\pm}0.0041$ & $0.6389{\pm}0.0044$ & $0.6509{\pm}0.0110$ & $-0.67\%$ & $0.2659{\pm}0.0043$ & $4.2899{\pm}0.2954$ & $\mathbf{0.0562{\pm}0.0009}$ & $0.0557{\pm}0.0008$ & $-6.10\%$ & $-4.79\%$ \\
w/ & \gsubgraph{} & PT-HGNN & $0.1216{\pm}0.0018$ & $-3.87\%$ & $0.6442{\pm}0.0034$ & $0.8122{\pm}0.0136$ & $0.7106{\pm}0.0075$ & $0.7007{\pm}0.0023$ & $0.6552{\pm}0.0041$ & $0.6534{\pm}0.0127$ & $-0.09\%$ & $0.2511{\pm}0.0061$ & $4.1773{\pm}0.0415$ & $0.0572{\pm}0.0006$ & $0.0386{\pm}0.0010$ & $+3.58\%$ & $-0.13\%$ \\
w/ & \gsubgraph{} & CTRL & $0.1206{\pm}0.0052$ & $-4.66\%$ & $0.6624{\pm}0.0542$ & $0.7842{\pm}0.0497$ & $0.7352{\pm}0.0039$ & $0.6884{\pm}0.0027$ & $0.6554{\pm}0.0095$ & $0.6598{\pm}0.0135$ & $+0.26\%$ & $0.2498{\pm}0.0042$ & $4.0577{\pm}0.0812$ & $0.0597{\pm}0.0003$ & $0.0390{\pm}0.0005$ & $+3.05\%$ & $-0.45\%$ \\
w/ & \gsubgraph{} & TGPM & $0.1198{\pm}0.0012$ & $-5.30\%$ & $0.6598{\pm}0.0035$ & $0.8337{\pm}0.0104$ & $0.7381{\pm}0.0163$ & $0.6728{\pm}0.0021$ & $0.6389{\pm}0.0040$ & $0.6498{\pm}0.0086$ & $+0.17\%$ & $0.2454{\pm}0.0028$ & $4.2054{\pm}0.1254$ & $0.0622{\pm}0.0005$ & $0.0398{\pm}0.0007$ & $+1.00\%$ & $-1.37\%$ \\
w/ & \gsubgraph{} & TVE & $0.1226{\pm}0.0020$ & $-3.08\%$ & $0.6456{\pm}0.0029$ & $0.8261{\pm}0.0231$ & $0.7304{\pm}0.0023$ & $0.6872{\pm}0.0015$ & $0.6294{\pm}0.0090$ & $0.6467{\pm}0.0060$ & $-0.50\%$ & $0.2475{\pm}0.0028$ & $4.0914{\pm}0.1265$ & $0.0578{\pm}0.0008$ & $0.0396{\pm}0.0008$ & $+3.52\%$ & $-0.02\%$ \\
\hdashline 
w/ & \gsubgraph{} & \ehist{} & $0.1224{\pm}0.0027$ & $-3.24\%$ & $0.6554{\pm}0.0035$ & $0.7925{\pm}0.0217$ & $\mathbf{0.7397{\pm}0.0274}$ & $0.6988{\pm}0.0010$ & $0.6562{\pm}0.0160$ & $0.6511{\pm}0.0023$ & $+0.39\%$ & $0.2405{\pm}0.0018$ & $4.1372{\pm}0.0644$ & $0.0563{\pm}0.0006$ & $0.0393{\pm}0.0001$ & $+4.87\%$ & $+0.67\%$ \\
w/ & \gsubgraph{} & \efuture{} & $0.1214{\pm}0.0017$ & $-4.03\%$ & $0.6418{\pm}0.0020$ & $0.8077{\pm}0.0140$ & $0.7301{\pm}0.0083$ & $0.7017{\pm}0.0040$ & $0.6548{\pm}0.0111$ & $0.6574{\pm}0.0108$ & $+0.32\%$ & $0.2435{\pm}0.0033$ & $\mathbf{4.0181{\pm}0.0943}$ & $0.0570{\pm}0.0002$ & $\mathbf{0.0366{\pm}0.0012}$ & $\mathbf{+6.90\%}$ & $\mathbf{+1.06\%}$ \\
\bottomrule
\end{tabular}}
\end{table*}

\paragraph{Comparison with constituent objectives.}
Using the matched reference within each panel of
\cref{tab:pretraining-heterognn,tab:pretraining-drsf}, the best staged
schedule in each backbone--encoding setting achieves a higher mean type
gain than either of its constituent objectives alone.
With both encodings, TS$\rightarrow$\ehist{} reaches $+3.02\%$ on
\heterognn{}, compared with $+0.11\%$ for \gsubgraph{} and $+1.14\%$
for \ehist{} alone. On \method{}, TS$\rightarrow$\efuture{} reaches
$+1.06\%$, compared with $-1.28\%$ for \gsubgraph{} and $-0.52\%$
for \efuture{} alone.
Without our encodings, TS$\rightarrow$\efuture{} leads on \heterognn{}
at $+2.51\%$, compared with $-0.88\%$ for \gsubgraph{} and $+1.34\%$
for \efuture{} alone. TS$\rightarrow$\ehist{} leads on \method{}
at $+1.58\%$, compared with $-2.25\%$ for \gsubgraph{} and $-0.24\%$
for \ehist{} alone. The gains over the stronger constituent are thus
$1.88$ and $1.58$ percentage points with our encodings, and $1.17$ and
$1.83$ percentage points without our encodings, for \heterognn{} and
\method{}, respectively.

\paragraph{Comparison with baseline pretraining.}
Under the matched A0/B0 references, TS$\rightarrow$CTRL is the strongest
baseline by mean type gain on \heterognn{} in both panels, reaching
$+1.62\%$ without our encodings and $+2.71\%$ with both encodings.
On \method{}, TS$\rightarrow$PT-HGNN leads the baselines without our
encodings at $+1.16\%$, while TS$\rightarrow$TVE leads with both
encodings at $-0.02\%$. Our best schedules exceed these baseline gains
by $0.88$ and $0.31$ percentage points on \heterognn{}, and $0.42$ and
$1.08$ points on \method{}, respectively. All six TS-initialized
baselines remain below the matched supervised control on \method{}
with both encodings, whereas both of our proposed schedules exceed it.

\paragraph{Dependence on the second-stage objective.}
Staging does not improve every objective. On \heterognn{} with both
encodings, the mean type gain of \efuture{} decreases from $+3.00\%$
alone to $+2.83\%$ after \gsubgraph{}, and that of \nattr{} decreases
from $+2.64\%$ to $+2.19\%$. In the same panel, \ehist{} improves
from $+1.14\%$ to $+3.02\%$, while PT-HGNN improves from $+0.53\%$
to $+1.89\%$. The effect also varies across backbones: with both
encodings on \method{}, \efuture{} improves from $-0.52\%$ alone
to $+1.06\%$ after \gsubgraph{}, unlike the decrease on \heterognn{}.
The temporal baselines show the same dependence on the backbone and
encoding configuration. With both encodings on \heterognn{}, CTRL,
TGPM, and TVE improve from $+0.93\%$, $+1.53\%$, and $+1.36\%$ alone
to $+2.71\%$, $+2.34\%$, and $+2.22\%$ after \gsubgraph{}, respectively.
On \method{} with both encodings, staging improves CTRL from $-1.50\%$
to $-0.45\%$ and TVE from $-0.54\%$ to $-0.02\%$, while TGPM is
nearly unchanged at $-1.37\%$ in both cases at the reported precision.
These comparisons show that the benefit of \gsubgraph{} initialization
depends on both the subsequent objective and the backbone, consistent
with the variation observed in the main ablation study.

\section{Architecture and Training Configuration}
\label{app:experimental-config}

All five backbones use three graph layers with hidden dimension 128;
the \glspl{gt} use eight attention heads. \Cref{tab:architecture-config,tab:training-config}
list the detailed settings for \heterognn{} and \method{} in the controlled
study. Both use the same materialized table features
and temporally sampled neighborhoods. We sample three hops with fanouts
of 128, 64, and 32 at the first, second, and third hops, respectively.

\paragraph{Attribute processing.}
\label{app:attribute-encoding}
Primary and foreign keys define graph structure and are excluded from
attribute features. The PyTorch Frame encoders in
\cref{tab:attribute-encoders} map each remaining column to 128 dimensions.
Parameters are specific to each table, with separate embeddings or maps
for its columns. Tables without attributes receive a constant numerical
feature. Text vectors are precomputed with
\texttt{average\_word\_embeddings\_glove.6B.300d} and remain fixed;
the column-specific maps from 300 to 128 dimensions are trainable.
Timestamp attributes encode year, month, day, day of week, hour, minute,
and second, with median-timestamp imputation for missing values. These
calendar features are separate from the record-age encoding in \mix{}.

\begin{table}[htbp]
\centering
\small
\caption{Column encoders used by both backbones. Only types present in a
table are instantiated; all outputs have dimension 128.}
\label{tab:attribute-encoders}
\begin{tabular}{@{}>{\raggedright\arraybackslash}p{0.16\textwidth}>{\raggedright\arraybackslash}p{0.37\textwidth}>{\raggedright\arraybackslash}p{0.39\textwidth}@{}}
\toprule
Attribute type & PyTorch Frame encoder & Operation \\
\midrule
Categorical & \texttt{EmbeddingEncoder} & Learned category lookup. \\
Numerical & \texttt{LinearEncoder} & Standardize using column statistics, then apply a learned affine map. Missing values are imputed with the column mean. \\
Multicategorical & \texttt{MultiCategorical\allowbreak EmbeddingEncoder} & Average learned embeddings of the categories in a cell. \\
Vector / text embedding & \texttt{LinearEmbeddingEncoder} & Apply a learned affine map to the input vector. \\
Timestamp & \texttt{TimestampEncoder} & Encode the year positionally and other calendar fields cyclically, then apply a learned linear map with bias. \\
\bottomrule
\end{tabular}
\end{table}

\label{app:column-fusion}
For a row $v$ in table $a=\phi(v)$ with $m_a$ attribute columns, let
$b_{v,j}\in\mathbb R^{128}$ be the embedding of column $j$.
We concatenate these embeddings in a fixed column order and apply the
table's ResNet:
\begin{equation}
    b_v=[b_{v,1}\Vert\cdots\Vert b_{v,m_a}]
    \in\mathbb R^{128m_a},\qquad
    x_v=\operatorname{ResNet}_a(b_v)\in\mathbb R^{128},
\end{equation}
where $\Vert$ denotes concatenation. Each table uses a four-layer ResNet
with hidden dimension 128 to produce its row representations.
The column encoders, ResNets, graph layers, and enabled temporal modules
are trained jointly with the current objective's head.

\begin{table}[htbp]
\centering
\small
\caption{Architecture settings for the selected GNN and \gls{gt}. Temporal
settings apply when the corresponding proposed encoding is enabled.}
\label{tab:architecture-config}
\begin{tabular}{>{\raggedright\arraybackslash}p{0.33\textwidth}>{\raggedright\arraybackslash}p{0.28\textwidth}>{\raggedright\arraybackslash}p{0.28\textwidth}}
\toprule
Setting & GNN (\heterognn{}) & \acrshort{gt} (\method{}) \\
\midrule
Graph layers / hidden width & 3 / 128 & 3 / 128 \\
Table feature encoder & 4-layer ResNet, width 128 & 4-layer ResNet, width 128 \\
Table-encoder dropout & 0.2 & 0.2 \\
Text embedding / learned mapping & Fixed GloVe / $300\rightarrow128$ & Fixed GloVe / $300\rightarrow128$ \\
Heads per attention branch & Not applicable & 8 (16 dimensions each) \\
Graph-operator dropout & None & 0.05 \\
Across-relation aggregation & Sum & Sum \\
Post-layer normalization / activation & Node-wise LayerNorm / ReLU & Node-wise LayerNorm / ReLU \\
Downstream head & Linear, $128\rightarrow C$ & Linear, $128\rightarrow C$ \\
Temporal sampling / hop fanouts & Uniform / $(128,64,32)$ & Uniform / $(128,64,32)$ \\
\mix{} dimension / fusion & 128 / addition & 128 / addition \\
\rotm{} mode / time unit / phase cap & Symmetric / 1 hour / 16 & Symmetric / 1 hour / 16 \\
\bottomrule
\end{tabular}
\par\smallskip
\parbox{\textwidth}{\small $C$ is the number of classes for multiclass
classification and 1 for binary classification or regression. Graph-operator
dropout refers to the graph module, separately from the table encoder.
The temporal operators are detailed in Appendix~\ref{app:encoding-details}.}
\end{table}

\begin{table}[htbp]
\centering
\small
\caption{Training settings shared by the selected GNN and \gls{gt}.
Single- and two-stage pretraining each receive 100,000 steps in total.}
\label{tab:training-config}
\begin{tabular}{>{\raggedright\arraybackslash}p{0.43\textwidth}>{\raggedright\arraybackslash}p{0.46\textwidth}}
\toprule
Setting & GNN and \acrshort{gt} \\
\midrule
Optimizer / learning rate / weight decay & AdamW / $2\times10^{-3}$ / $10^{-5}$ \\
Learning-rate schedule & Constant; no warmup or decay \\
Single-stage pretraining epochs / steps per epoch & 10 / 10,000 \\
Two-stage pretraining epochs / steps per epoch (each stage) & 10 / 5,000 \\
Total pretraining steps (either schedule) & 100,000 \\
Pretraining batch size & 256; 512 for the H\&M curricula \\
Downstream epochs / steps per epoch & 10 / 10,000 \\
Downstream batch size & 1,024 for Arxiv and F1; 512 for Avito, H\&M, and Event \\
Fine-tuning scope & Full encoder and newly initialized task head \\
Downstream losses & Cross-entropy (multiclass), binary cross-entropy with logits (binary), $L_1$ (regression) \\
Validation frequency & Every downstream epoch \\
Downstream checkpoint criterion & Maximum validation accuracy or ROC--AUC; minimum validation MAE \\
Random seeds & 42, 43, 44, 45 \\
Reported task statistics & Mean and standard deviation across four runs \\
\bottomrule
\end{tabular}
\end{table}

\paragraph{Training and evaluation.}
We optimize all models using AdamW at the constant learning rate and weight
decay in \cref{tab:training-config}. Single-stage pretraining uses 10 epochs
with 10,000 sampled steps per epoch, for a total budget of 100,000 steps.
Two-stage pretraining allocates 10 epochs with 5,000 sampled steps per epoch
to each stage, giving 50,000 steps per stage and the same total budget of
100,000 steps. These budgets apply to both our objectives and the pretraining
baselines, including their staged variants. Each downstream fine-tuning run
uses 10 epochs with 10,000 steps per epoch, totaling 100,000 steps; models
trained directly on downstream tasks without pretraining use the same budget.
There is no patience-based early stopping. For all pretraining strategies, we select the checkpoint with the lowest training loss.
The next stage loads the selected encoder and initializes its
objective-specific modules and AdamW optimizer afresh. For each seed in
$\{42,43,44,45\}$, we run pretraining where applicable,
fine-tuning, validation-based checkpoint selection, and test evaluation
independently, using the same chronological splits. The mean validation ranks used to choose one study backbone per family, as a trade-off to reduce computational cost, are described in
Appendix~\ref{app:backbone-selection}.
Reported means and standard deviations summarize these runs;
relative gains are computed from the task means.

\paragraph{Objective settings.}
\label{app:objective-settings}
\ehist{} uses $K_{\mathrm{neg}}=4$ negative destinations per positive link.
\efuture{} uses $K_{\mathrm{fut}}=2$ negatives, candidate horizons
$\mathcal H=\{1,7,30,90,365\}$ days, count-loss weight
$\lambda_{\mathrm{count}}=0.25$, and Smooth L1 transition threshold
$\beta_{\mathrm{SL}}=1$.
\gsubgraph{} uses edge-drop probability $p_{\mathrm{edge}}=0.2$,
feature-mask probability $p_{\mathrm{mask}}=0.2$, and contrastive
temperature $T=0.2$. Its graph and seed contrastive losses both have unit weight.
The seed regularizer uses variance weight $\lambda_{\mathrm{var}}=1$,
covariance weight $\lambda_{\mathrm{cov}}=0.01$, target standard deviation
$\gamma=1$, and variance stabilizer $\epsilon_{\mathrm{var}}=10^{-4}$;
its full definition is in Appendix~\ref{app:subgraph-details}.
These settings are shared across backbones and stages. We performed no
hyperparameter search; the values here and in
Appendix~\ref{app:encoding-details} are fixed experimental choices,
and their optimality has not been evaluated.

\paragraph{Horizon eligibility and window boundaries.}
\label{app:future-example}
A horizon $\delta$ is eligible only if the seed is available at
$\tau=t_{\mathrm{evt}}-\delta$. Events at $\tau$ belong to the history
and are excluded from the target count; events at $\tau+\delta$ are
included. A negative seed must be available at $\tau$ and have no events
of the selected relation in this window. An event at the seed's first
observation cannot define a positive example for any positive horizon.

\section{Temporal Encoding Details}
\label{app:encoding-details}

\paragraph{\mix{} parameterization.}
\label{app:time-mix-details}
\mix{} uses $K_{\mathrm{time}}=4$ trainable log-scales $\eta_k$,
initialized so that the scales $\exp(\eta_k)$ equal 1 second, 1 minute, 1 hour,
and 1 day, respectively. Durations are represented in seconds, with positive scales
$s_k=\operatorname{clip}(\exp(\eta_k),s_{\min},s_{\max})$,
where $s_{\min}=1$ second and $s_{\max}=2592000$ seconds (30 days).
The bounds constrain the scales, not the record ages. The mixture weights are
\begin{equation}
    \pi_k=\frac{\exp(\gamma_k)}
    {\sum_{j=1}^{K_{\mathrm{time}}}\exp(\gamma_j)},
\end{equation}
where $\gamma_k$ is a trainable logit, initialized equally across scales
to give uniform weights. Parameter sharing and missing-timestamp handling
follow \cref{sec:temporal-encoding}.

\paragraph{\rotm{} implementation.}
\label{app:time-rope-details}
\rotm{} computes the bounded, signed log-interval $\rho_{uv}$ defined in
\cref{sec:temporal-encoding}, with time scale $s_{\mathrm R}=3600$ seconds
(1 hour) and phase cap $\rho_{\max}=16$.
For a feature dimension $d$, let $d_{\mathrm{even}}$ be the largest even
integer no greater than $d$. Consecutive dimension pairs are rotated with
frequencies $\omega_j=10000^{-2j/d_{\mathrm{even}}}$,
$j=0,\ldots,d_{\mathrm{even}}/2-1$. For pair $(x_{2j},x_{2j+1})$,
\begin{equation}
    R_j(\rho)
    \begin{bmatrix}x_{2j}\\x_{2j+1}\end{bmatrix}
    =\begin{bmatrix}
    \cos(\omega_j\rho)&-\sin(\omega_j\rho)\\
    \sin(\omega_j\rho)&\cos(\omega_j\rho)
    \end{bmatrix}
    \begin{bmatrix}x_{2j}\\x_{2j+1}\end{bmatrix}.
\end{equation}
An unmatched final dimension is unchanged. GNN messages use the feature
width; \gls{gt} rotations use the attention-head width.

\paragraph{GNN message construction.}
\label{app:rotation-implementation}
This message construction extends temporal rotation
to GNN aggregation without query--key attention.

\paragraph{GT attention weighting.}
\label{app:gt-rotary-attention}
We retain the query--key rotation mechanism of THGFM's Rotary Temporal
Attention \citep{peng2026thgfm}, using our bounded, signed log-interval
$\rho_{uv}$ as its phase input. For a fixed relation $r$ and attention
head, let $q_v,k_u,p_u\in\mathbb R^{d_h}$ be the query, key, and value
projected from the current layer inputs $y_v$ and $y_u$, with any
branch-specific endpoint adapters applied before projection.
Relation, branch, and head indices are omitted for clarity.
The symmetric \texttt{qk} mode computes
\begin{equation}
    \begin{aligned}
    \widehat q_{uv}&=R(+\rho_{uv}/2)q_v,\qquad
    \widehat k_{uv}=R(-\rho_{uv}/2)k_u,\\
    \alpha_{uv}&=\operatorname{softmax}_{u\in\mathcal N_r(v)}
    \left(\frac{\widehat q_{uv}^{\top}\widehat k_{uv}}{\sqrt{d_h}}\right).
    \end{aligned}
    \label{eq:app-gt-rotary-attention}
\end{equation}
Here $\mathcal N_r(v)$ contains incoming neighbors under relation $r$.
Values remain unrotated, and the head output is
$\sum_{u\in\mathcal N_r(v)}\alpha_{uv}p_u$.
For \method{}, rotation is applied separately in the relation-specific
attention branch and the shared-space branch with endpoint adapters. Their
outputs are then combined by the backbone's type-conditioned gated sum.

\section{Pretraining Objective Details}
\label{app:objective-details}

This section specifies the decoders, losses, and sampling details for
\cref{sec:ehist,sec:efuture,sec:gsubgraph}. We retain the main text's notation,
including historical node representations $h_v\in\mathbb R^d$.
Hyperparameter values are in Appendix~\ref{app:objective-settings}.

\paragraph{Historical link recovery.}
\label{app:hist-details}
A link's observation time is the timestamp of the row containing the
foreign key, or the other endpoint's timestamp if the first is missing.
Links with no timestamped endpoint use the pretraining cutoff. For each
positive link, we draw $K_{\mathrm{neg}}$ negative destinations under the
constraints in \cref{sec:ehist}. The bilinear score is
\begin{equation}
    s_{\mathrm{hist}}(u,v,r)=h_u^\top W_r h_v,
\end{equation}
where $W_r\in\mathbb R^{d\times d}$ is learned for relation $r$.
The ranking loss is
\begin{equation}
    \mathcal L_{\mathrm{Rel\text{-}Hist}}
    =\frac{1}{K_{\mathrm{neg}}}\sum_{k=1}^{K_{\mathrm{neg}}}
    \operatorname{softplus}\!\left[
    s_{\mathrm{hist}}(u,v_k^-,r)-s_{\mathrm{hist}}(u,v,r)\right],
\end{equation}
where $v_k^-$ is the $k$-th negative destination and
$\operatorname{softplus}(x)=\log(1+e^x)$.

\paragraph{Future relation activity.}
\label{app:future-details}
Each forward foreign-key link $e=(u\rightarrow s)$ contributes one event
of relation $r$ to $\mathcal E_r(s)$; reverse edges are not counted again.
The timestamp assignment uses $t_e=t_u$ when the source table has a
timestamp field, and otherwise uses $t_s$ when the destination table has
one. Relations between two tables without timestamps are excluded.
We retain only events at or before the pretraining cutoff whose timestamped
endpoints are available by $t_e$.

For a timestamped seed, $t_{\mathrm{first}}(s)=t_s$. Otherwise, availability
is estimated from its earliest incident relation time within the cutoff.
Seeds without an assigned availability time are excluded.

We use candidate horizons $\mathcal H=\{1,7,30,90,365\}$ days,
converted to the same unit as the timestamps. For an event involving
$s$ at $t_{\mathrm{evt}}$, the eligible horizons are
\begin{equation}
    \label{eq:future-horizon-eligibility}
    \mathcal D(s,t_{\mathrm{evt}})=
    \{\delta\in\mathcal H:
    t_{\mathrm{first}}(s)\leq t_{\mathrm{evt}}-\delta\}.
\end{equation}
If time assignment falls back to $t_s$, then
$t_e=t_{\mathrm{first}}(s)$, so no positive horizon is eligible.
Thus, relations retained for this objective use source timestamps,
as described in \cref{sec:efuture}.
The eligible event and relation sets are
\begin{equation}
    \mathcal E_r^+=\{(s,e):e\in\mathcal E_r(s),\,
    \mathcal D(s,t_e)\neq\varnothing\},\qquad
    \mathcal R_{\mathrm{fut}}=\{r:\mathcal E_r^+\neq\varnothing\}.
\end{equation}
We sample a relation, an event, and a horizon uniformly from these sets:
\begin{equation}
    \begin{aligned}
    r&\sim\operatorname{Unif}(\mathcal R_{\mathrm{fut}}),&
    (s^+,e^+)&\sim\operatorname{Unif}(\mathcal E_r^+),\\
    \delta&\sim\operatorname{Unif}(\mathcal D(s^+,t_{e^+})),&
    \tau&=t_{e^+}-\delta.
    \end{aligned}
\end{equation}
Here $\operatorname{Unif}$ denotes uniform sampling and
$t_{e^+}=t_{\mathrm{evt}}$ in the main text. Event sampling uses replacement.
For the same relation and window, negative seeds are drawn from
\begin{equation}
    \mathcal V^-(s^+,r,\tau,\delta)=
    \{s:\phi(s)=\phi(s^+),\;
    t_{\mathrm{first}}(s)\leq\tau,\;
    c_r(s;\tau,\delta)=0\}.
\end{equation}
We draw $K_{\mathrm{fut}}$ negatives, using distinct seeds when possible
and reusing valid draws if necessary. Attempts without valid negatives
are skipped. Targets follow \cref{eq:future-targets}; window boundaries
are specified in Appendix~\ref{app:future-example}.

The occurrence and count decoders share a horizon encoder and layer
normalization, but use separate relation embeddings and biases.
The occurrence logit is
\begin{equation}
    w_{r,\delta}=w_r+g_{\mathrm{hor}}(\log(1+\delta_{\mathrm{day}})),\qquad
    s_{\mathrm{act}}(s,r,\delta)=
    \frac{\operatorname{LN}(h_s)^\top
    \bigl(w_{r,\delta}/\|w_{r,\delta}\|_2\bigr)}{\sqrt d}+b_r,
\end{equation}
where $w_r\in\mathbb R^d$ and $b_r$ are learned relation parameters,
$g_{\mathrm{hor}}:\mathbb R\rightarrow\mathbb R^d$ is the horizon encoder,
$\delta_{\mathrm{day}}$ is the horizon in days,
$\operatorname{LN}$ is layer normalization, and $\|\cdot\|_2$
is the Euclidean norm. The count score $s_{\mathrm{count}}(s,r,\delta)$
uses the same form with its own relation embedding and bias. Its
nonnegative prediction is
\begin{equation}
    \widehat y_{\mathrm{count}}=
    \operatorname{softplus}(s_{\mathrm{count}}(s,r,\delta)).
\end{equation}
Using the targets in \cref{eq:future-targets}, the losses for a positive
$s^+$ and negatives $s_k^-$ at fixed $(r,\tau,\delta)$ are
\begin{equation}
    \begin{aligned}
    \mathcal L_{\mathrm{act}}&=\tfrac12\left[
    \operatorname{softplus}(-s_{\mathrm{act}}(s^+,r,\delta))+
    \frac{1}{K_{\mathrm{fut}}}\sum_{k=1}^{K_{\mathrm{fut}}}
    \operatorname{softplus}(s_{\mathrm{act}}(s_k^-,r,\delta))\right],\\
    \mathcal L_{\mathrm{count}}&=\tfrac12\left[
    \operatorname{SmoothL1}(\widehat y_{\mathrm{count}}^+,y_{\mathrm{count}}^+)+
    \frac{1}{K_{\mathrm{fut}}}\sum_{k=1}^{K_{\mathrm{fut}}}
    \operatorname{SmoothL1}(\widehat y_{\mathrm{count},k}^-,0)\right],\\
    \mathcal L_{\mathrm{Rel\text{-}Future}}
    &=\mathcal L_{\mathrm{act}}+\lambda_{\mathrm{count}}\mathcal L_{\mathrm{count}}.
    \end{aligned}
\end{equation}
Superscripts $+$ and $-$ identify positive and negative examples;
negative count targets are zero. $\operatorname{SmoothL1}$ uses threshold
$\beta_{\mathrm{SL}}$. Each loss equally weights the positive example and
the mean over negatives, and $\lambda_{\mathrm{count}}$ weights the count loss.

\paragraph{Symmetric subgraph contrast.}
\label{app:subgraph-details}
For each minibatch, we sample seeds of one node type without replacement.
For its $B$ historical neighborhoods, let $z_i^{(1)}$ and $z_i^{(2)}$ be
neighborhood $i$'s two pooled views after the shared projection head.
The symmetric InfoNCE loss is
\begin{equation}
    S_{ij}=\frac{\cos(z_i^{(1)},z_j^{(2)})}{T},\qquad
    \mathcal L_{\mathrm{graph}}=\frac{1}{2B}\sum_{i=1}^{B}
    [\operatorname{CE}(S_{i,:},i)+\operatorname{CE}(S_{:,i},i)],
\end{equation}
where $S_{ij}$ is cosine similarity scaled by temperature $T$,
$\operatorname{CE}(\cdot,i)$ is cross-entropy with positive index $i$,
and $S_{i,:}$ and $S_{:,i}$ are row $i$ and column $i$ of the score matrix.
The same loss applied to seed representations through a separate
projection head gives $\mathcal L_{\mathrm{seed}}$.
Both projection heads have the architecture
$\operatorname{Linear}(d,d)\!\to\!\operatorname{LayerNorm}(d)
\!\to\!\operatorname{GELU}\!\to\!\operatorname{Linear}(d,d)$.
Each head is shared across the two views; the graph and seed heads
have separate parameters.

\paragraph{Seed representation regularization.}
Let $X^{(v)}\in\mathbb R^{B\times d}$ contain the encoder's seed
representations in view $v\in\{1,2\}$ before the seed projection head.
We apply the variance and covariance components of
VICReg~\citep{bardes2022vicreg} to these representations, with
cross-view agreement provided by $\mathcal L_{\mathrm{seed}}$.
For a single view $X$, define
\begin{equation}
    \begin{aligned}
    \bar X_j&=\frac{1}{B}\sum_{i=1}^{B}X_{ij},\qquad
    \sigma_j(X)=\sqrt{\frac{1}{B}\sum_{i=1}^{B}
    (X_{ij}-\bar X_j)^2+\epsilon_{\mathrm{var}}},\\
    \mathcal V(X)&=\frac{1}{d}\sum_{j=1}^{d}
    \max\!\left(0,\gamma-\sigma_j(X)\right).
    \end{aligned}
\end{equation}
The variance uses divisor $B$ and is computed on the unnormalized
encoder outputs. This hinge penalizes feature dimensions whose
standard deviation across seeds falls below $\gamma$.

For the covariance term, we first apply LayerNorm independently to
each row of $X$, without learnable scale or bias and with numerical
stabilizer $\epsilon_{\mathrm{LN}}=10^{-5}$. Let
$Y=\operatorname{LayerNorm}(X)$ and
$\bar Y_j=B^{-1}\sum_{i=1}^{B}Y_{ij}$. We then compute
\begin{equation}
    C_{jk}(X)=\frac{1}{B-1}\sum_{i=1}^{B}
    (Y_{ij}-\bar Y_j)(Y_{ik}-\bar Y_k),\qquad
    \mathcal C(X)=\frac{1}{d}\sum_{j\ne k}C_{jk}(X)^2.
\end{equation}
Thus, covariance is computed across seeds after normalization within
each seed; its off-diagonal penalty reduces redundancy between feature
dimensions. The sum includes all ordered pairs $j\ne k$ and is
normalized by $d$. Both regularizers are set to zero when $B<2$;
the covariance term is also zero when $d<2$.

We average each term over the two views and form
\begin{equation}
    \mathcal R_{\mathrm{reg}}=\frac{1}{2}\sum_{v=1}^{2}
    \left[\lambda_{\mathrm{var}}\mathcal V(X^{(v)})
    +\lambda_{\mathrm{cov}}\mathcal C(X^{(v)})\right],\qquad
    \mathcal L_{\mathrm{Temp\text{-}Sub}}
    =\mathcal L_{\mathrm{graph}}+\mathcal L_{\mathrm{seed}}
    +\mathcal R_{\mathrm{reg}}.
\end{equation}
The weights and variance parameters are given in
Appendix~\ref{app:objective-settings}.

\section{Pretraining and Fine-Tuning Curves}
\label{app:training-dynamics}

The curves cover both backbones under two settings: \textsc{DayPE} without \rotm{}, and \mix{} with \rotm{}. These are the encoding configurations of the A0 and B0 controls, respectively.
Pretraining panels are grouped by database; solid lines show single-objective
training and dashed lines show stage 2 after \gsubgraph{} (TS). Epoch
counts restart at each stage. Each epoch comprises 10,000 steps for
single-stage pretraining and 5,000 steps for each stage of two-stage
pretraining, so equal epoch counts do not indicate equal numbers of steps
in these panels. Metrics should be compared within a row because objectives
use different scales. Fine-tuning panels use 10,000 steps per epoch and are grouped by
task and show validation accuracy or ROC--AUC (higher is better) and MAE
(lower is better). Test results are in
\cref{tab:pretraining-heterognn,tab:pretraining-drsf}.

% Compact float pages: two related plots per page, with readable captions.
\begingroup
\setlength{\floatsep}{10pt plus 2pt minus 2pt}
\captionsetup[figure]{skip=4pt}
\makeatletter
\setlength{\@fptop}{0pt}
\setlength{\@fpsep}{10pt plus 2pt minus 2pt}
\setlength{\@fpbot}{0pt plus 1fil}
\makeatother

\begin{figure}[p]
\centering
\includegraphics[width=0.87\textwidth]{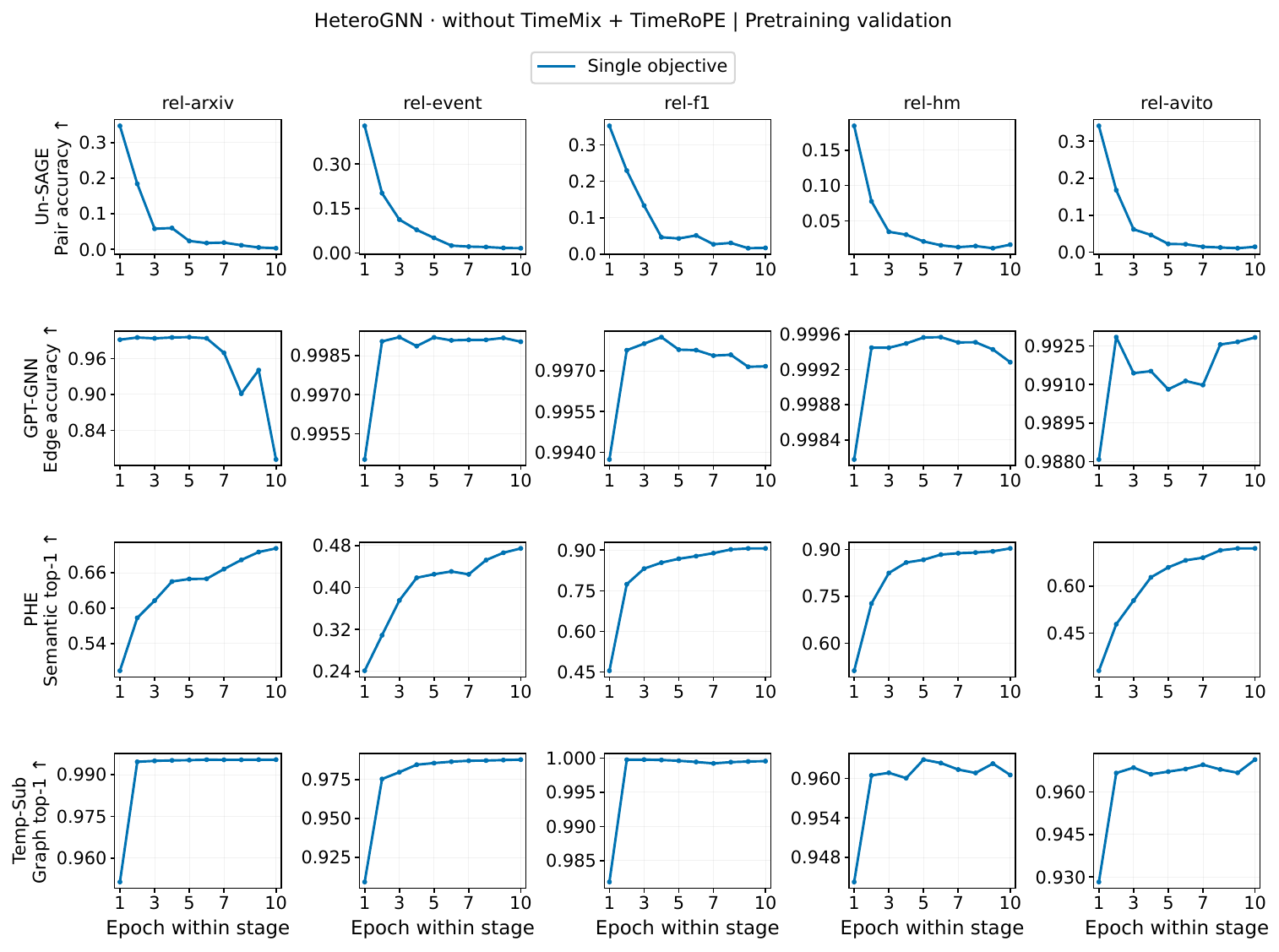}
\caption{Pretraining metrics for \heterognn{} with \textsc{DayPE} and without \rotm{}: single objectives.}
\label{fig:display-pretrain-heterognn-without-1}
\label{fig:curve-pretrain-heterognn-without-1}
\end{figure}
\begin{figure}[p]
\centering
\includegraphics[width=0.87\textwidth]{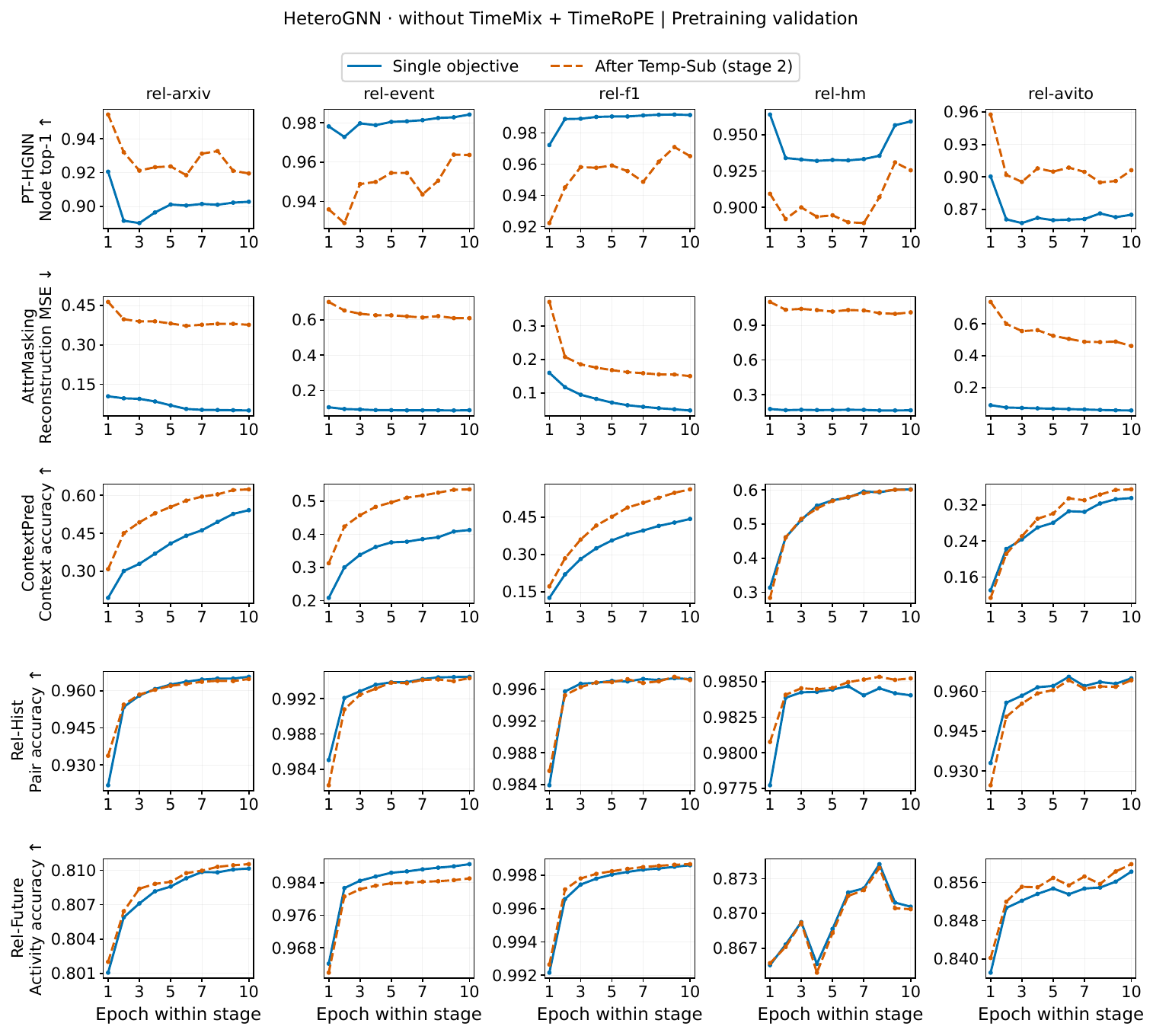}
\caption{Pretraining metrics for \heterognn{} with \textsc{DayPE} and without \rotm{}: single objectives and stage 2 after \gsubgraph{}.}
\label{fig:display-pretrain-heterognn-without-2}
\label{fig:curve-pretrain-heterognn-without-2}
\end{figure}
\begin{figure}[p]
\centering
\includegraphics[width=\textwidth]{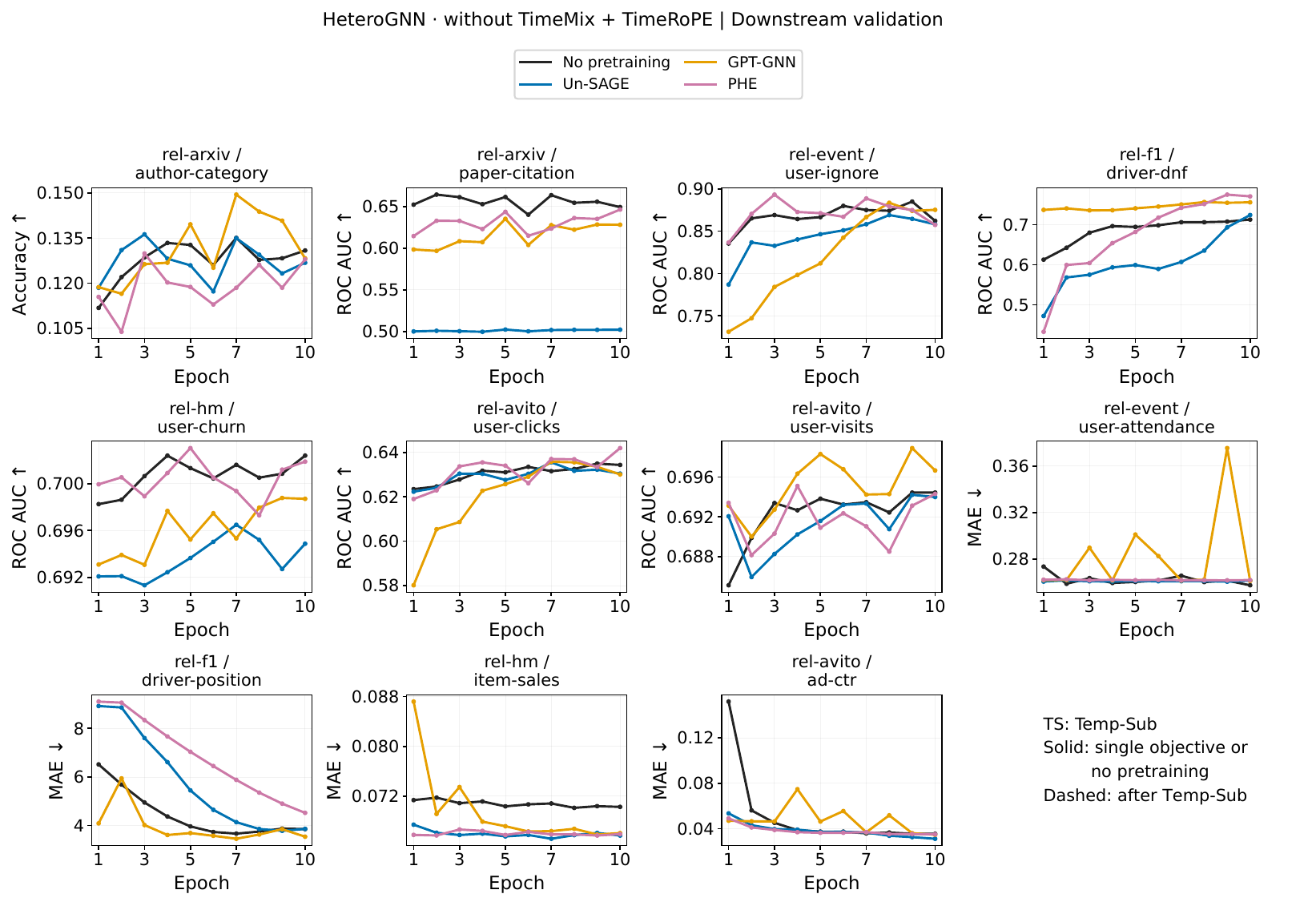}
\caption{Fine-tuning validation metrics for \heterognn{} with \textsc{DayPE} and without \rotm{}: single-stage baselines.}
\label{fig:display-finetune-heterognn-without-1}
\label{fig:curve-finetune-heterognn-without-1}
\end{figure}
\begin{figure}[p]
\centering
\includegraphics[width=\textwidth]{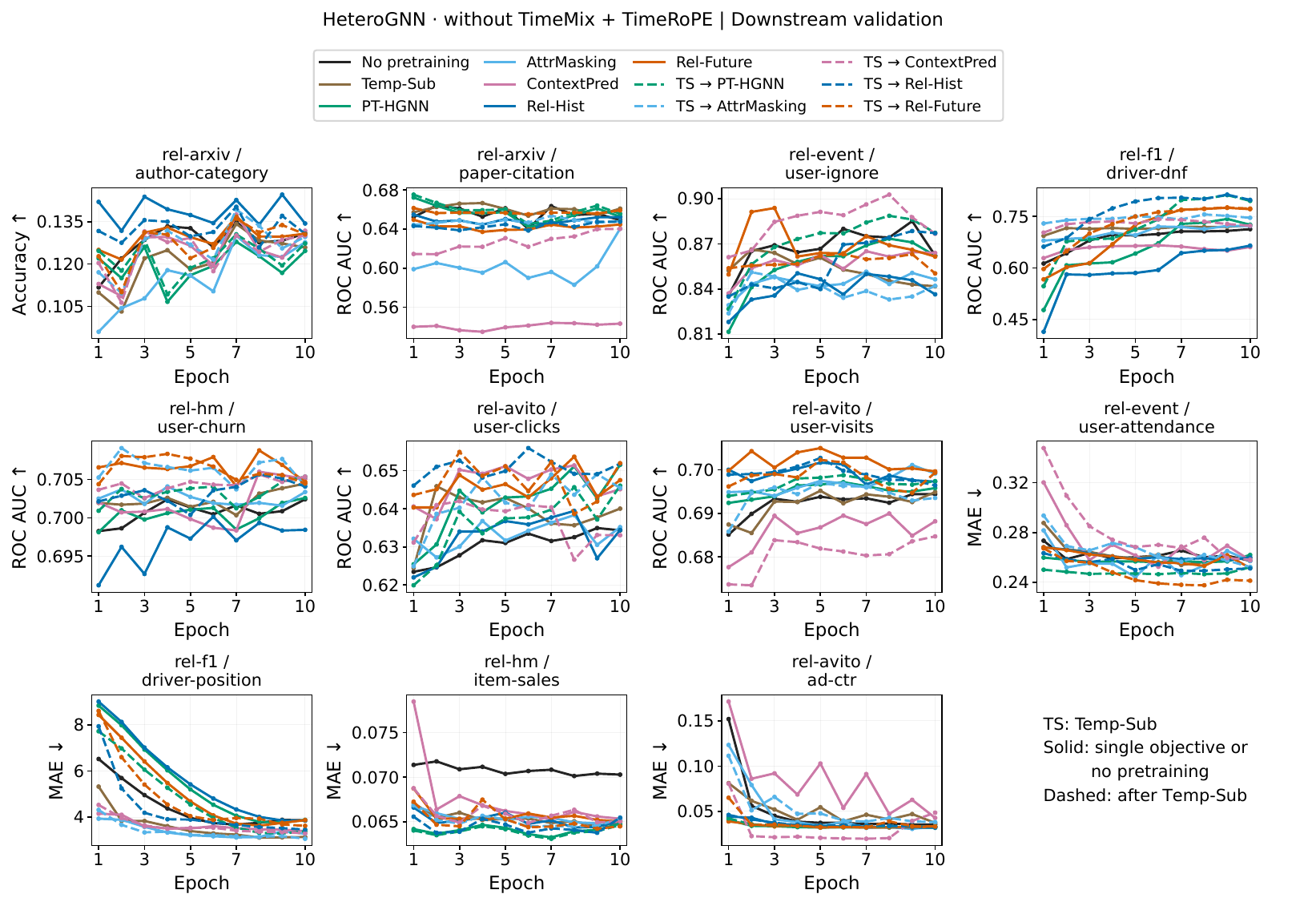}
\caption{Fine-tuning validation metrics for \heterognn{} with \textsc{DayPE} and without \rotm{}: single and staged objectives.}
\label{fig:display-finetune-heterognn-without-2}
\label{fig:curve-finetune-heterognn-without-2}
\end{figure}

\begin{figure}[p]
\centering
\includegraphics[width=0.87\textwidth]{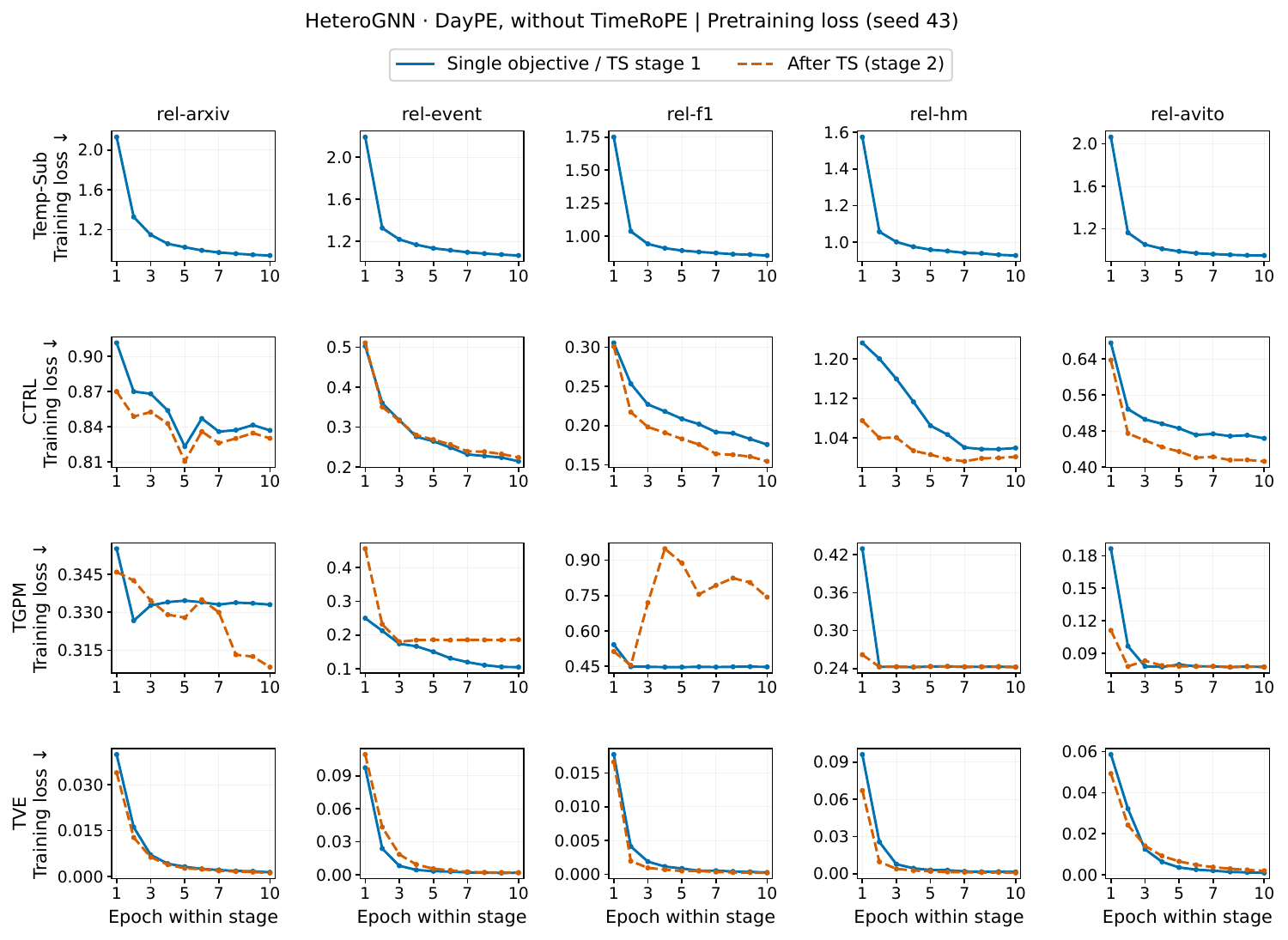}
\caption{Pretraining losses for \heterognn{} with \textsc{DayPE} and without \rotm{}: CTRL, TGPM, TVE, and the shared \gsubgraph{} first stage.}
\label{fig:curve-pretrain-heterognn-without-3}
\end{figure}

\begin{figure}[p]
\centering
\includegraphics[width=\textwidth]{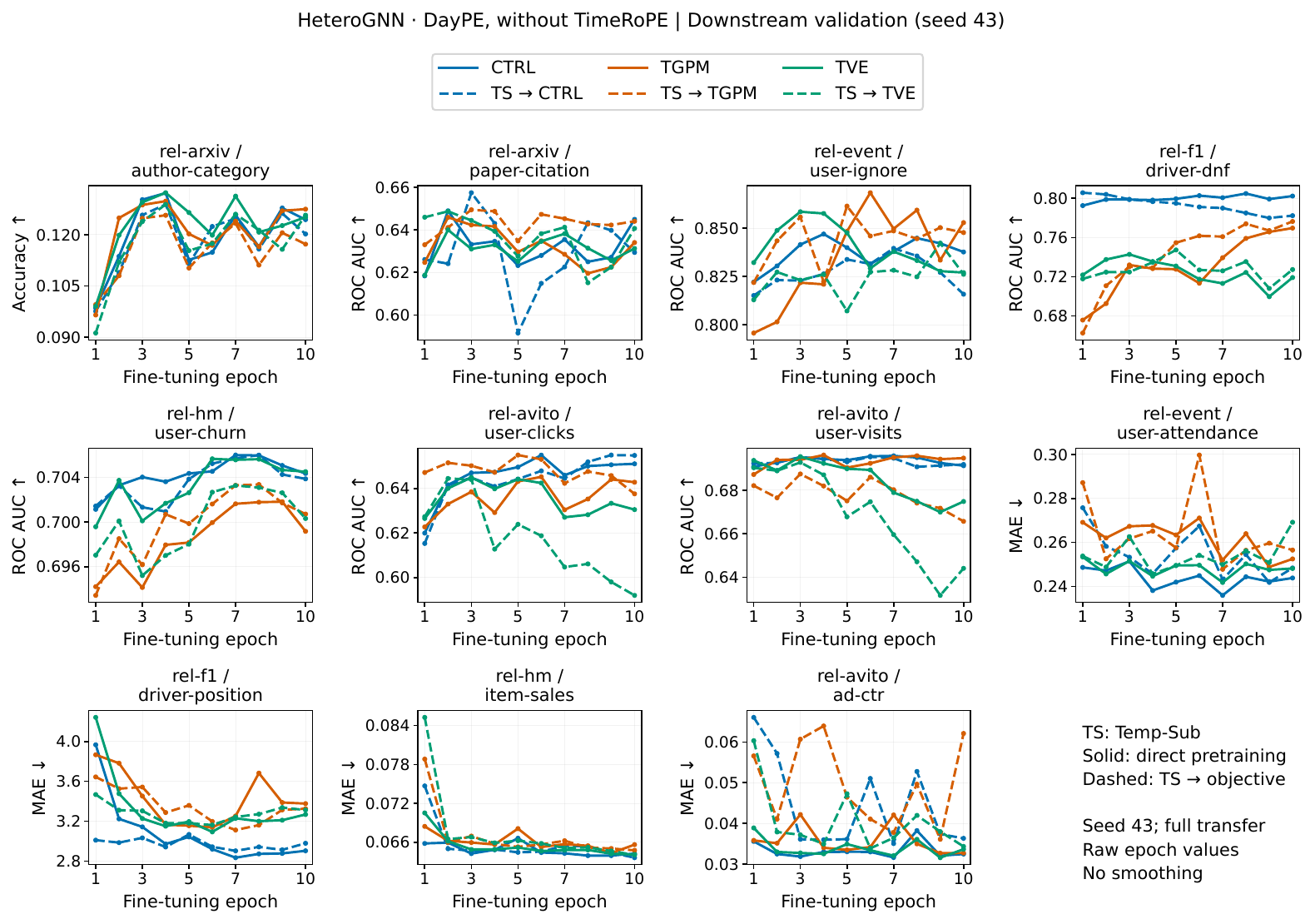}
\caption{Fine-tuning validation metrics for \heterognn{} with \textsc{DayPE} and without \rotm{}: CTRL, TGPM, TVE, and their TS-initialized variants.}
\label{fig:curve-finetune-heterognn-without-3}
\end{figure}

\begin{figure}[p]
\centering
\includegraphics[width=0.87\textwidth]{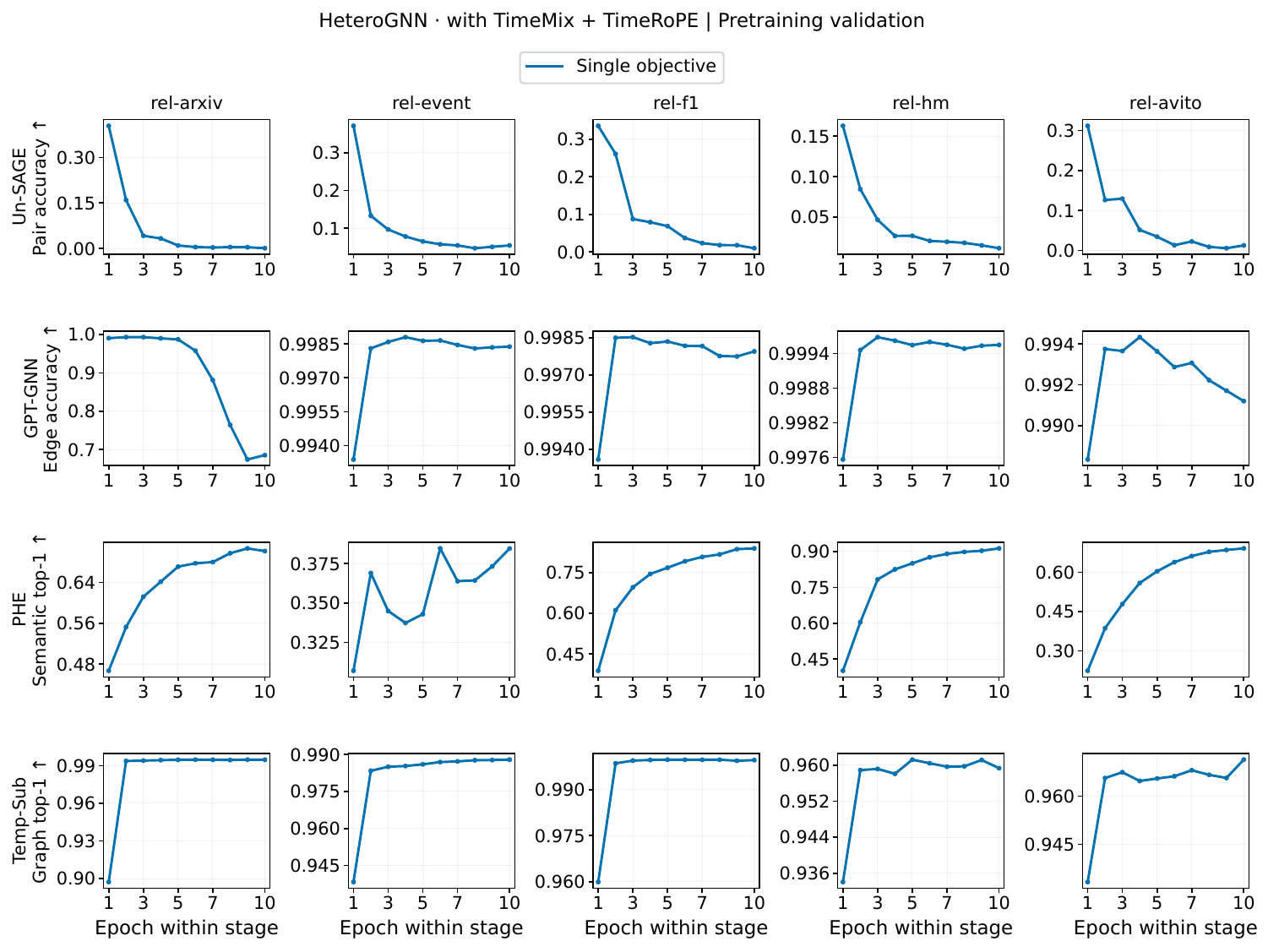}
\caption{Pretraining metrics for \heterognn{} with \mix{} and \rotm{}: single objectives.}
\label{fig:display-pretrain-heterognn-with-1}
\label{fig:curve-pretrain-heterognn-with-1}
\end{figure}
\begin{figure}[p]
\centering
\includegraphics[width=0.87\textwidth]{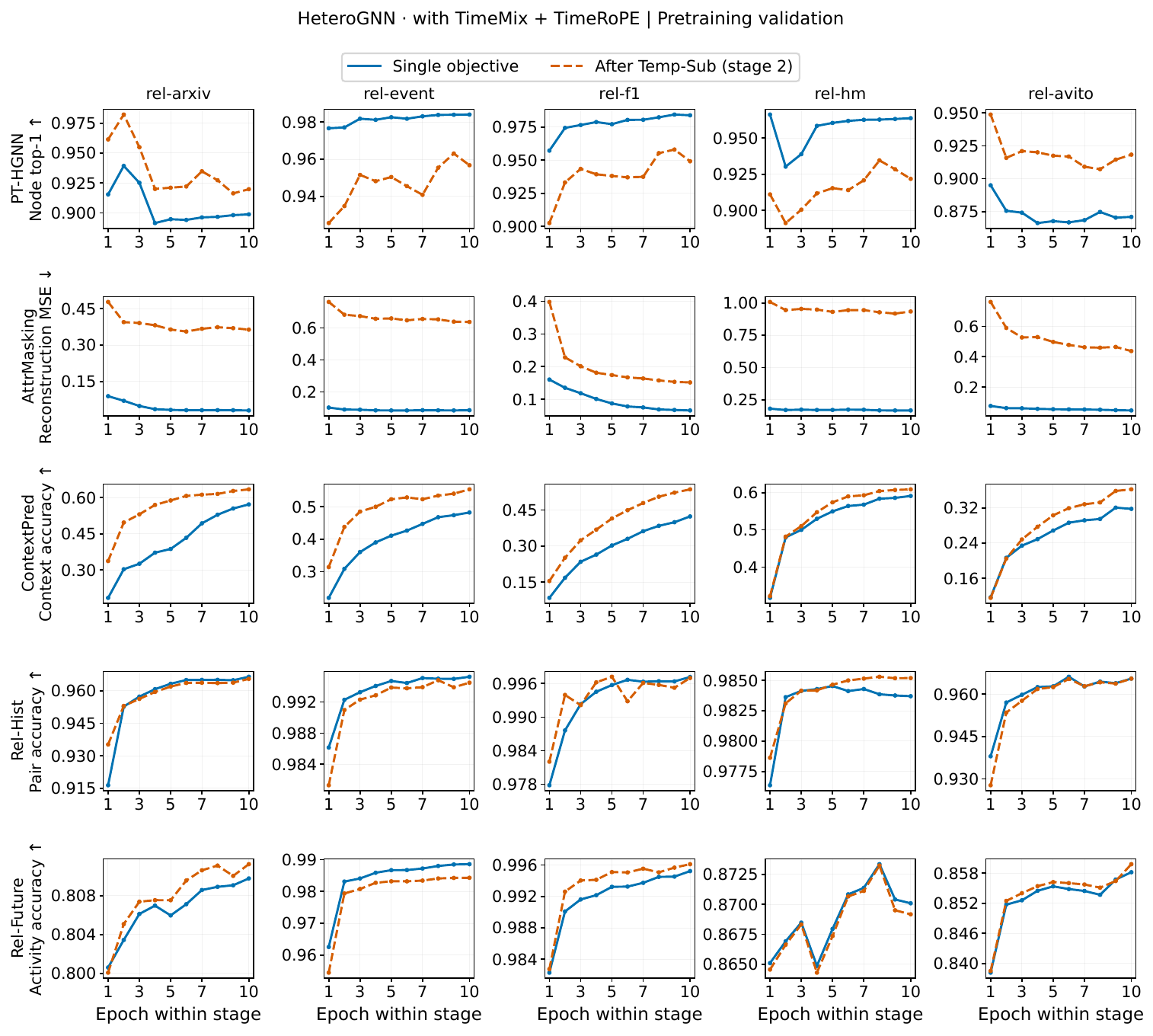}
\caption{Pretraining metrics for \heterognn{} with \mix{} and \rotm{}: single objectives and stage 2 after \gsubgraph{}.}
\label{fig:display-pretrain-heterognn-with-2}
\label{fig:curve-pretrain-heterognn-with-2}
\end{figure}
\begin{figure}[p]
\centering
\includegraphics[width=\textwidth]{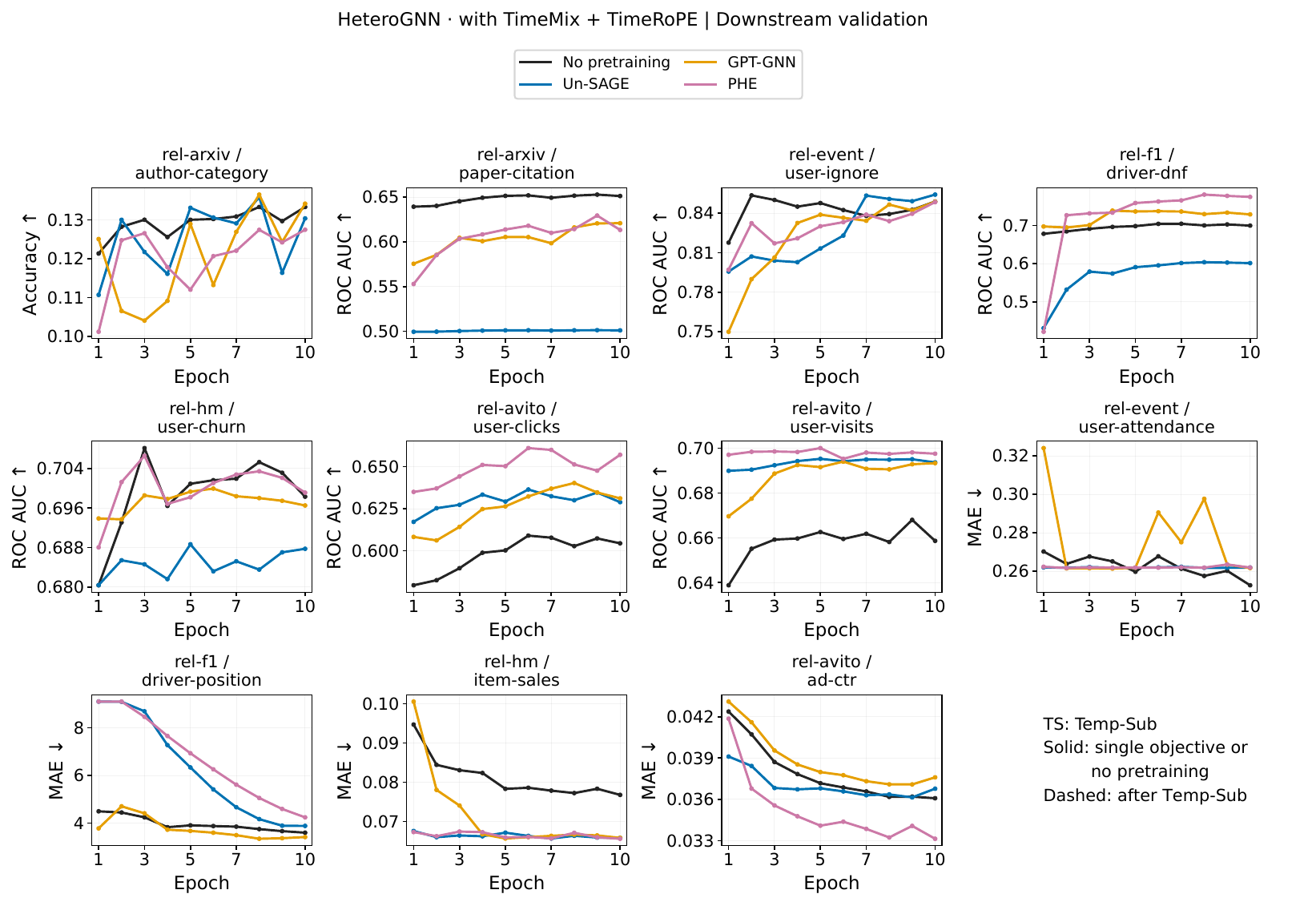}
\caption{Fine-tuning validation metrics for \heterognn{} with \mix{} and \rotm{}: single-stage baselines.}
\label{fig:display-finetune-heterognn-with-1}
\label{fig:curve-finetune-heterognn-with-1}
\end{figure}
\begin{figure}[p]
\centering
\includegraphics[width=\textwidth]{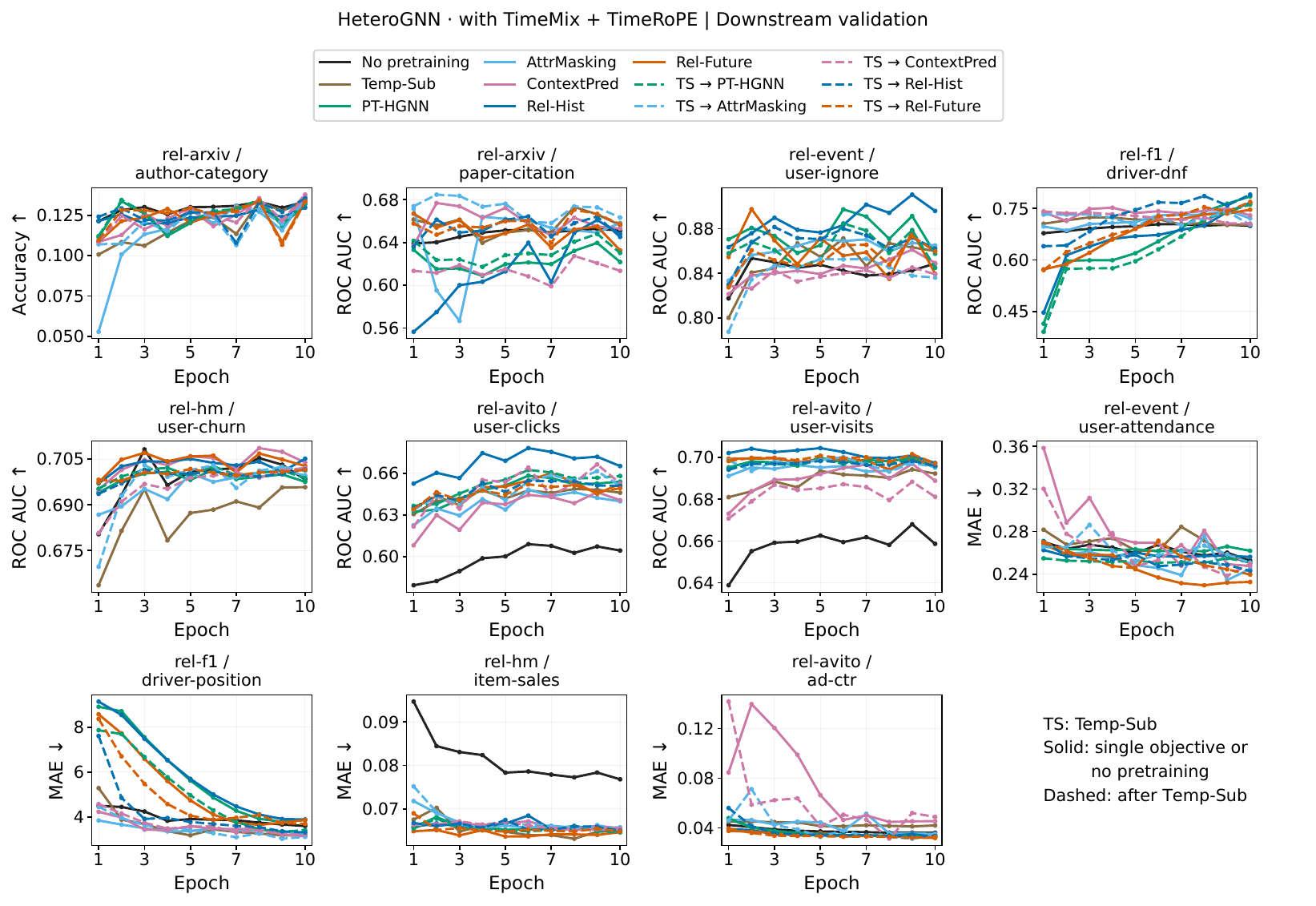}
\caption{Fine-tuning validation metrics for \heterognn{} with \mix{} and \rotm{}: single and staged objectives.}
\label{fig:display-finetune-heterognn-with-2}
\label{fig:curve-finetune-heterognn-with-2}
\end{figure}

\begin{figure}[p]
\centering
\includegraphics[width=0.87\textwidth]{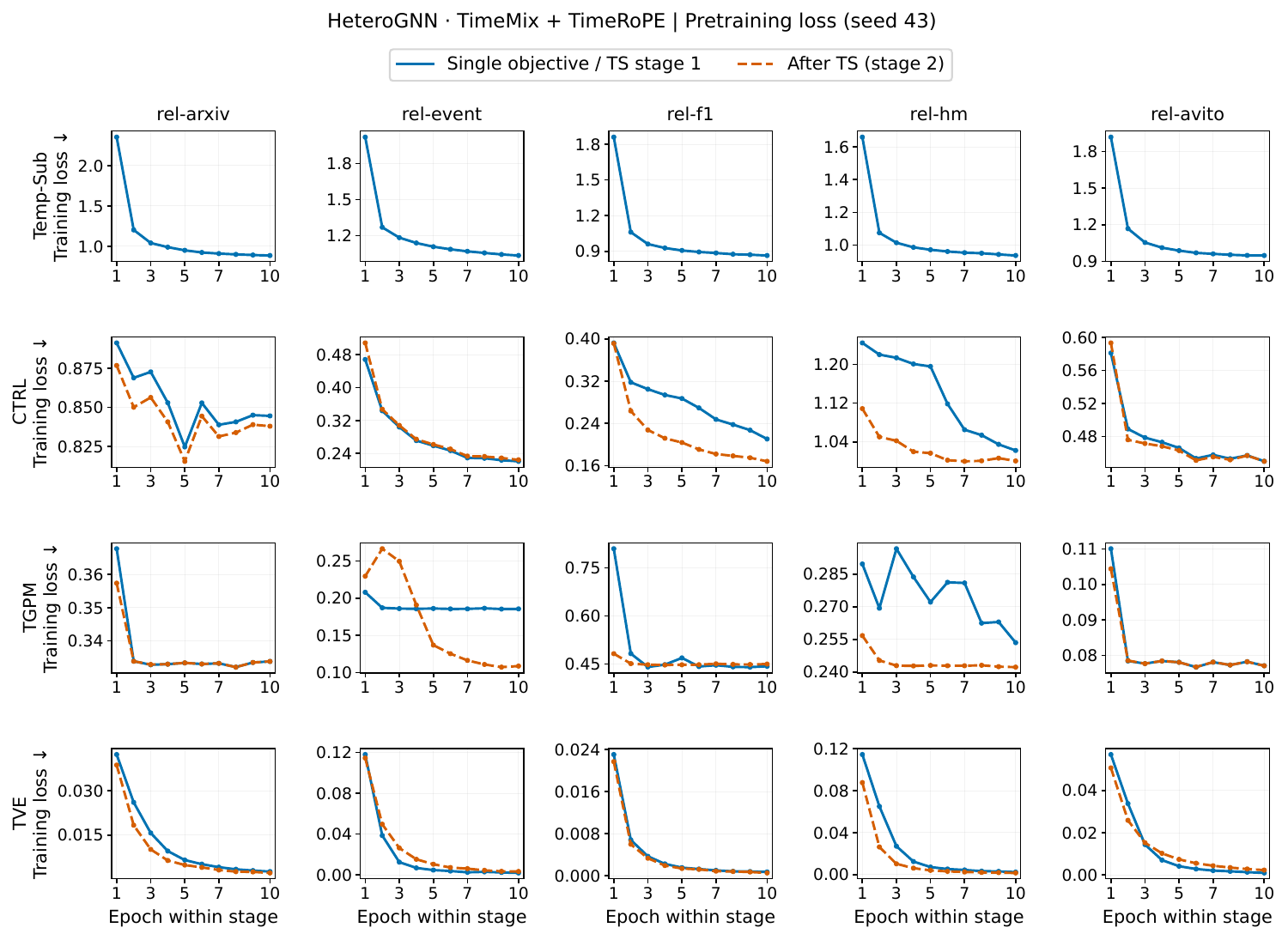}
\caption{Pretraining losses for \heterognn{} with \mix{} and \rotm{}: CTRL, TGPM, TVE, and the shared \gsubgraph{} first stage.}
\label{fig:curve-pretrain-heterognn-with-3}
\end{figure}

\begin{figure}[p]
\centering
\includegraphics[width=\textwidth]{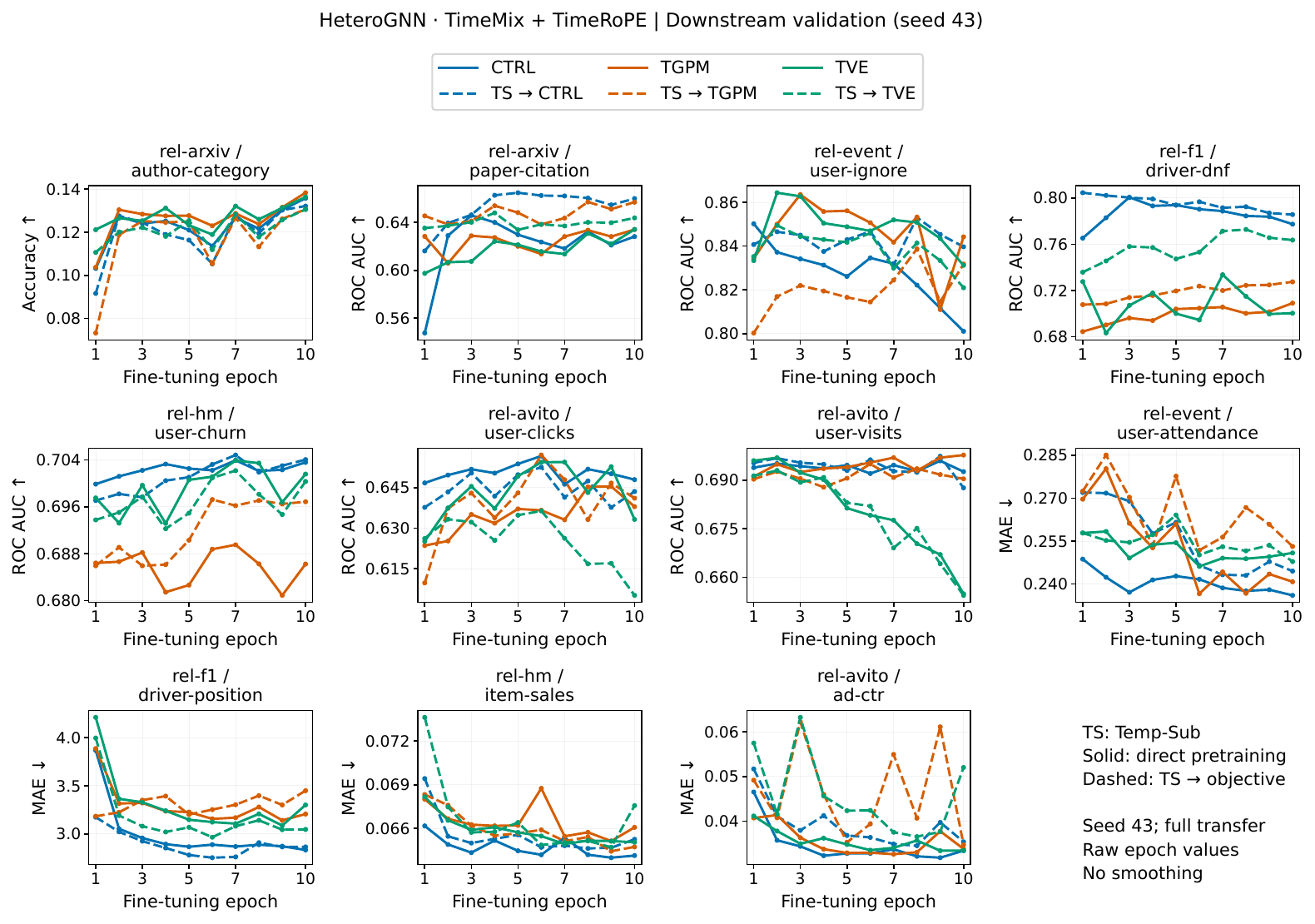}
\caption{Fine-tuning validation metrics for \heterognn{} with \mix{} and \rotm{}: CTRL, TGPM, TVE, and their TS-initialized variants.}
\label{fig:curve-finetune-heterognn-with-3}
\end{figure}

\begin{figure}[p]
\centering
\includegraphics[width=0.87\textwidth]{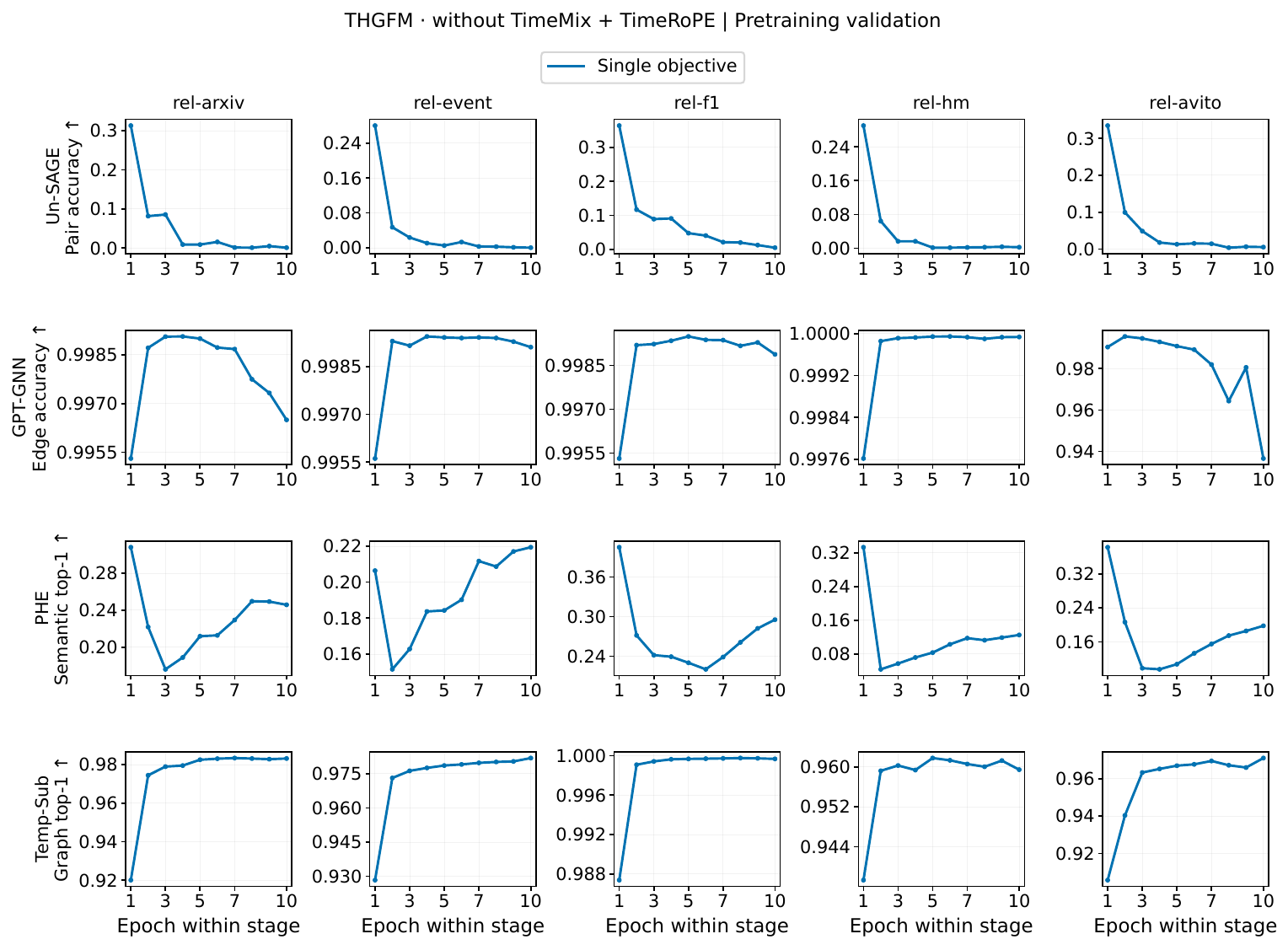}
\caption{Pretraining metrics for \method{} with \textsc{DayPE} and without \rotm{}: single objectives.}
\label{fig:display-pretrain-thgfm-without-1}
\label{fig:curve-pretrain-thgfm-without-1}
\end{figure}
\begin{figure}[p]
\centering
\includegraphics[width=0.87\textwidth]{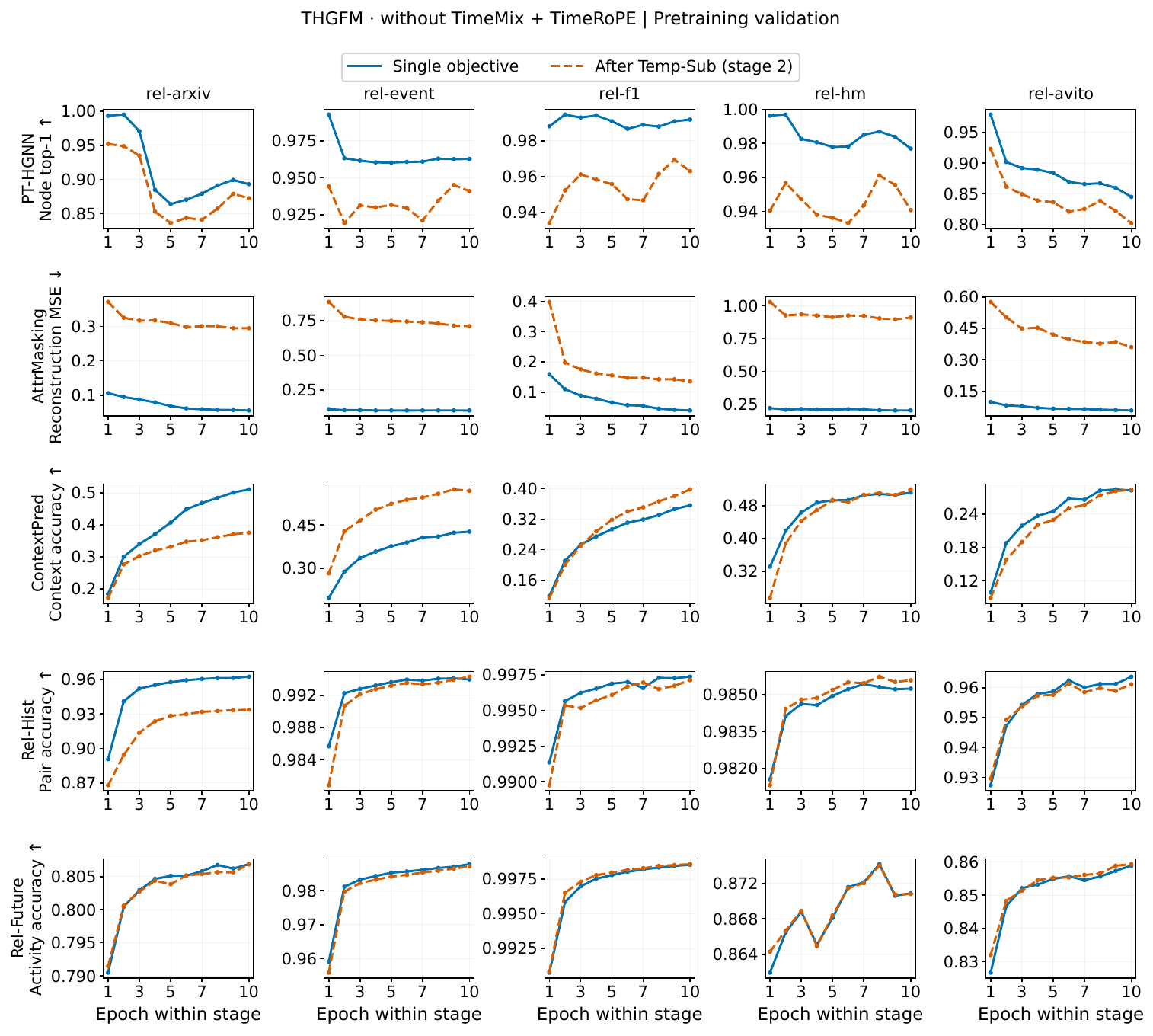}
\caption{Pretraining metrics for \method{} with \textsc{DayPE} and without \rotm{}: single objectives and stage 2 after \gsubgraph{}.}
\label{fig:display-pretrain-thgfm-without-2}
\label{fig:curve-pretrain-thgfm-without-2}
\end{figure}
\begin{figure}[p]
\centering
\includegraphics[width=\textwidth]{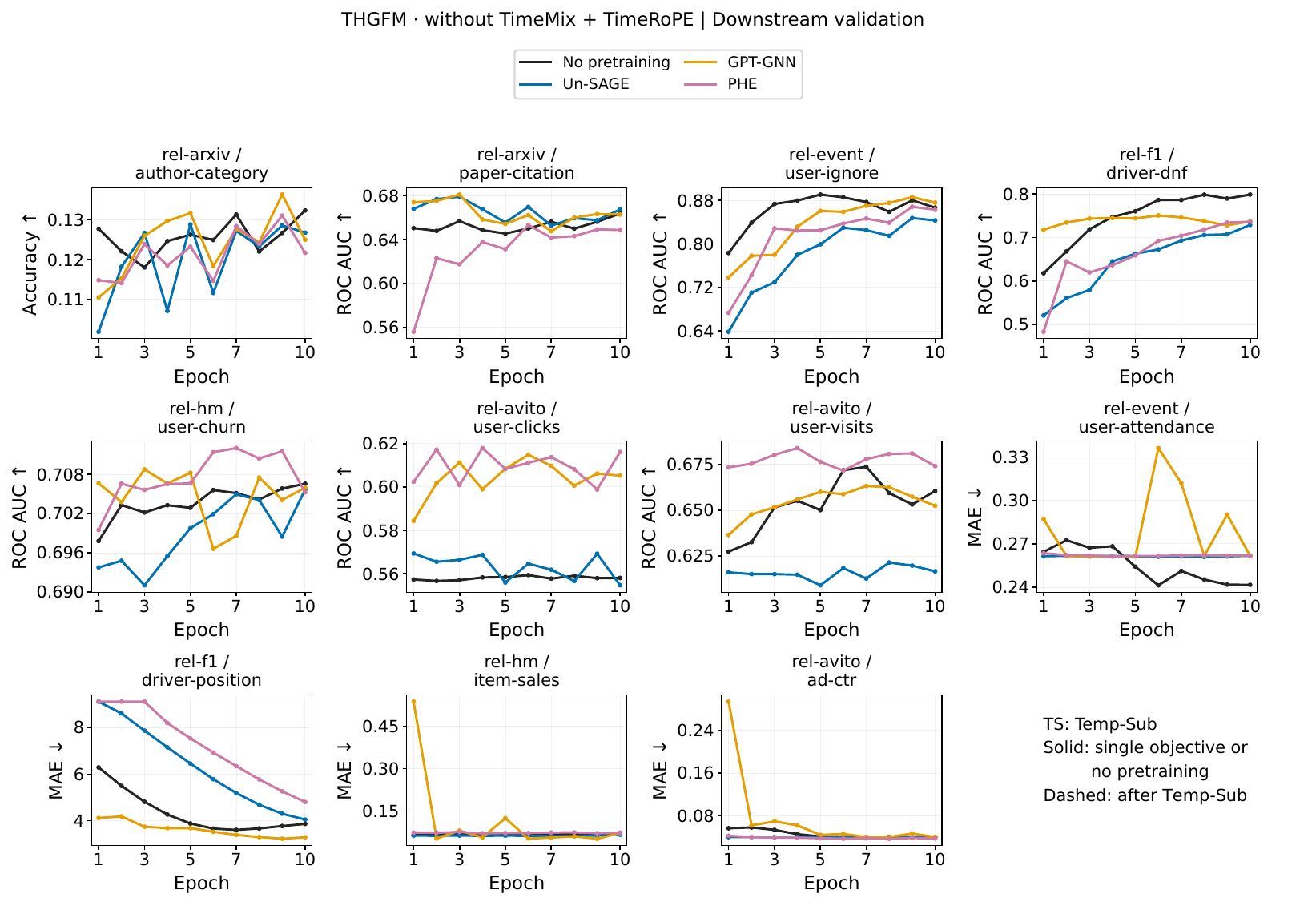}
\caption{Fine-tuning validation metrics for \method{} with \textsc{DayPE} and without \rotm{}: single-stage baselines.}
\label{fig:display-finetune-thgfm-without-1}
\label{fig:curve-finetune-thgfm-without-1}
\end{figure}
\begin{figure}[p]
\centering
\includegraphics[width=\textwidth]{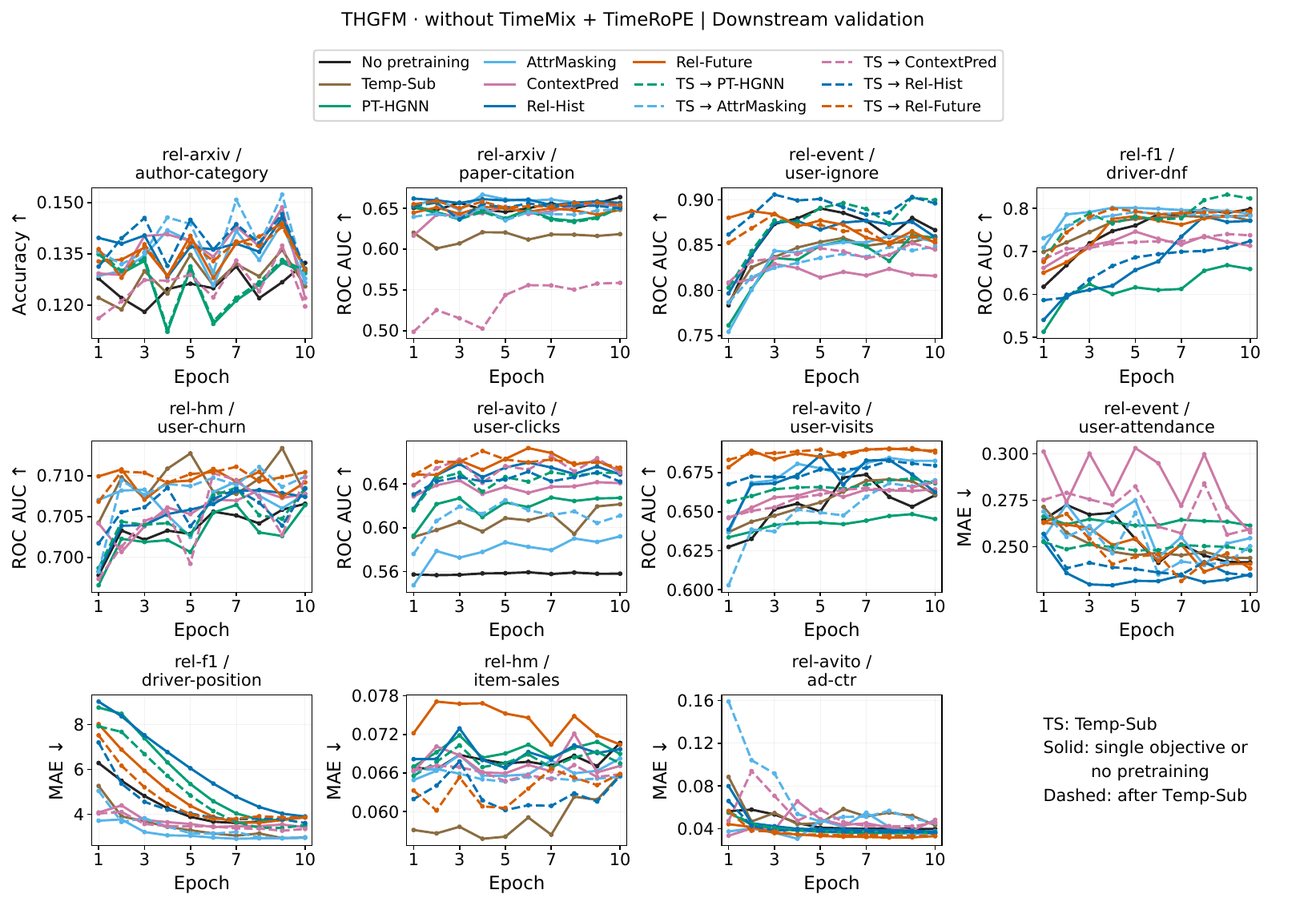}
\caption{Fine-tuning validation metrics for \method{} with \textsc{DayPE} and without \rotm{}: single and staged objectives.}
\label{fig:display-finetune-thgfm-without-2}
\label{fig:curve-finetune-thgfm-without-2}
\end{figure}

\begin{figure}[p]
\centering
\includegraphics[width=0.87\textwidth]{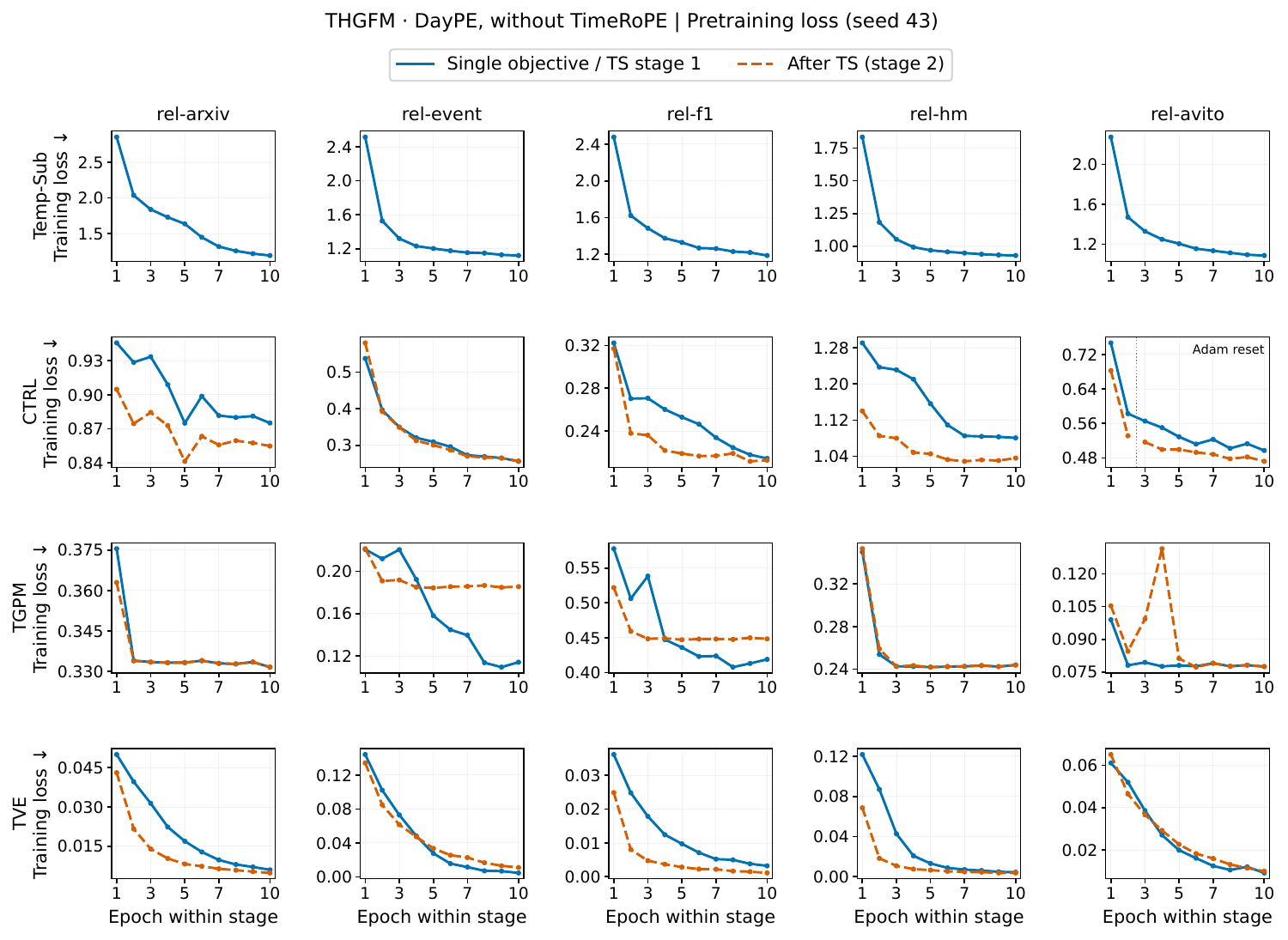}
\caption{Pretraining losses for \method{} with \textsc{DayPE} and without \rotm{}: CTRL, TGPM, TVE, and the shared \gsubgraph{} first stage.}
\label{fig:curve-pretrain-thgfm-without-3}
\end{figure}

\begin{figure}[p]
\centering
\includegraphics[width=\textwidth]{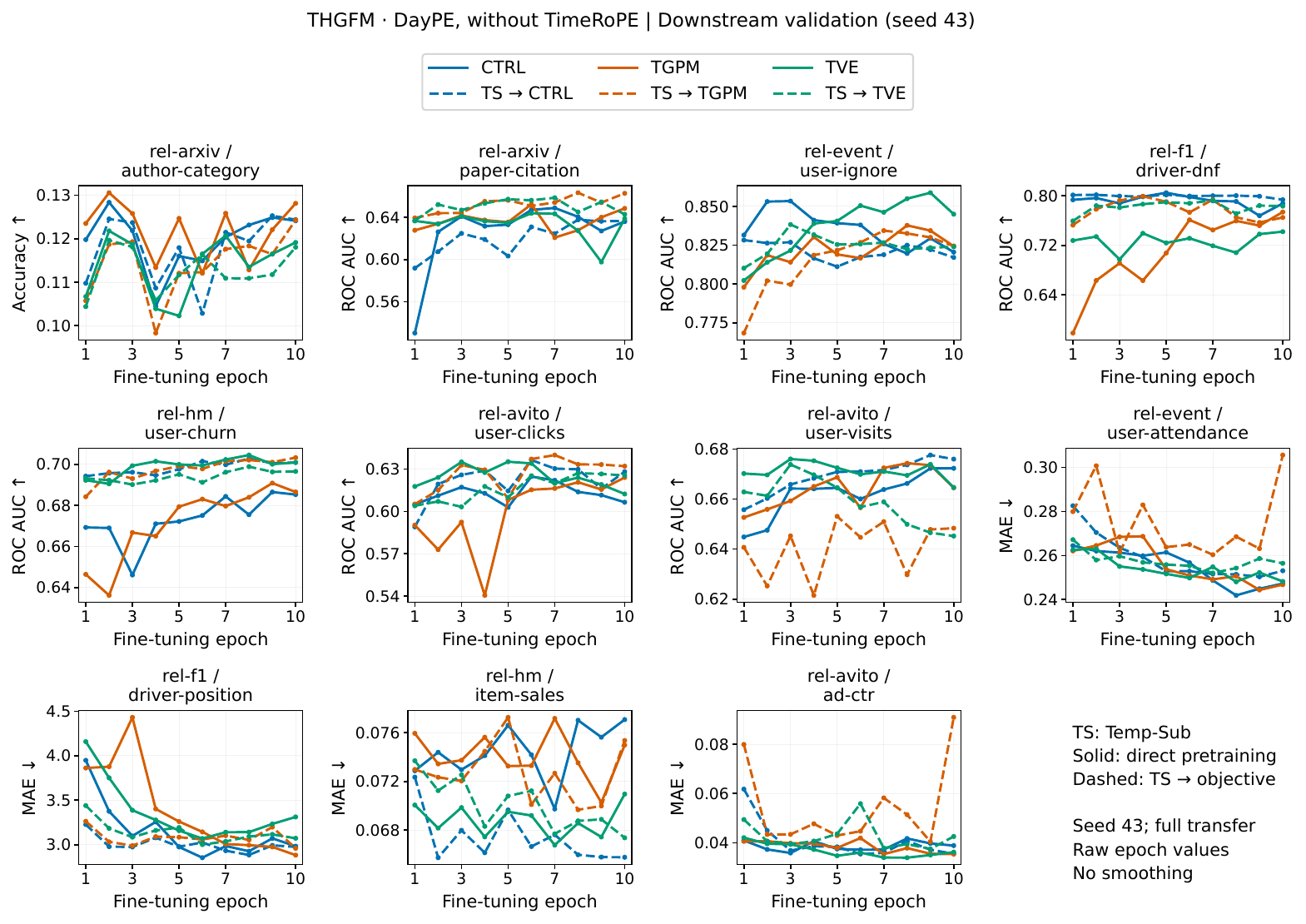}
\caption{Fine-tuning validation metrics for \method{} with \textsc{DayPE} and without \rotm{}: CTRL, TGPM, TVE, and their TS-initialized variants.}
\label{fig:curve-finetune-thgfm-without-3}
\end{figure}

\begin{figure}[p]
\centering
\includegraphics[width=0.87\textwidth]{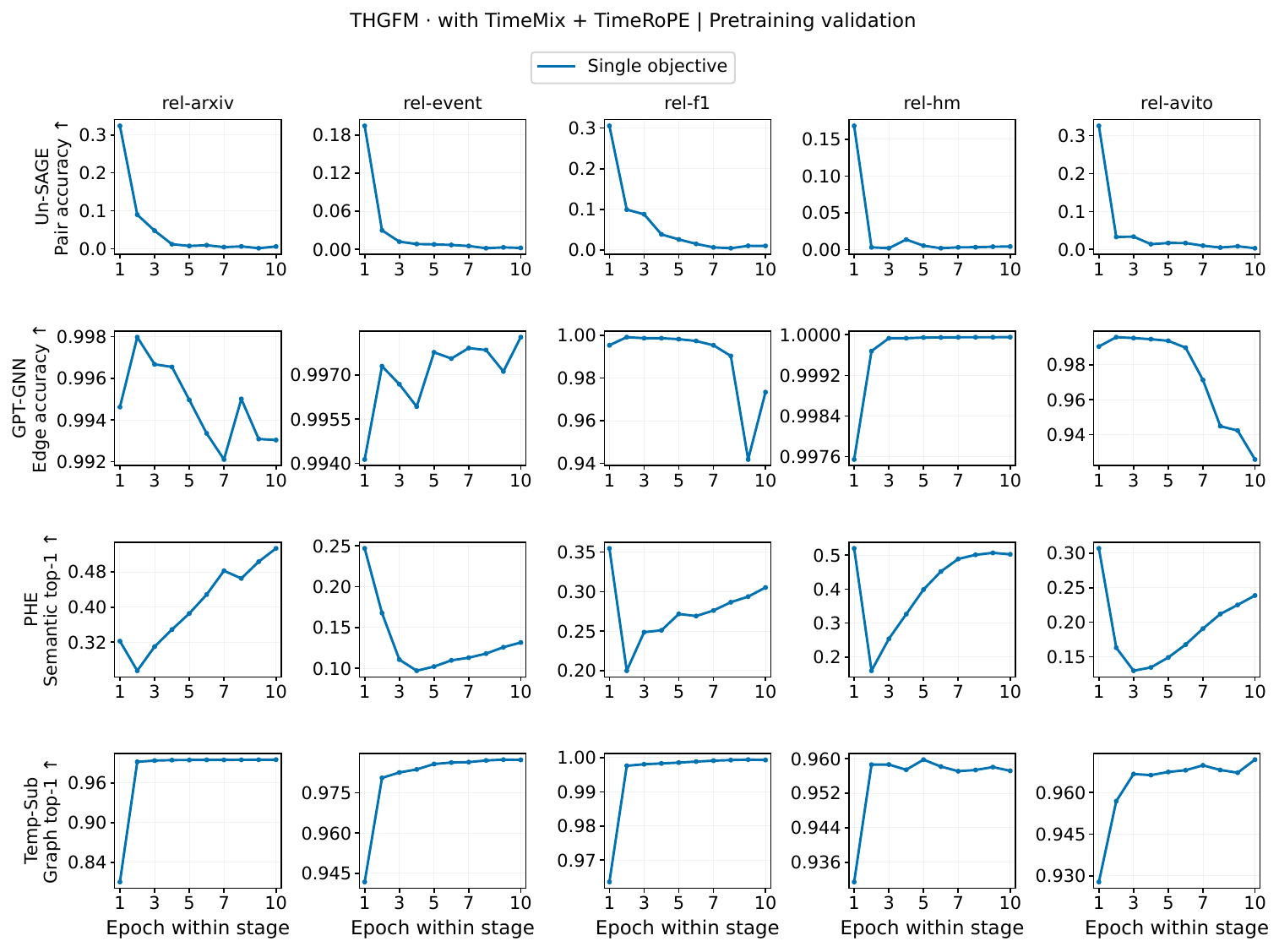}
\caption{Pretraining metrics for \method{} with \mix{} and \rotm{}: single objectives.}
\label{fig:display-pretrain-thgfm-with-1}
\label{fig:curve-pretrain-thgfm-with-1}
\end{figure}
\begin{figure}[p]
\centering
\includegraphics[width=0.87\textwidth]{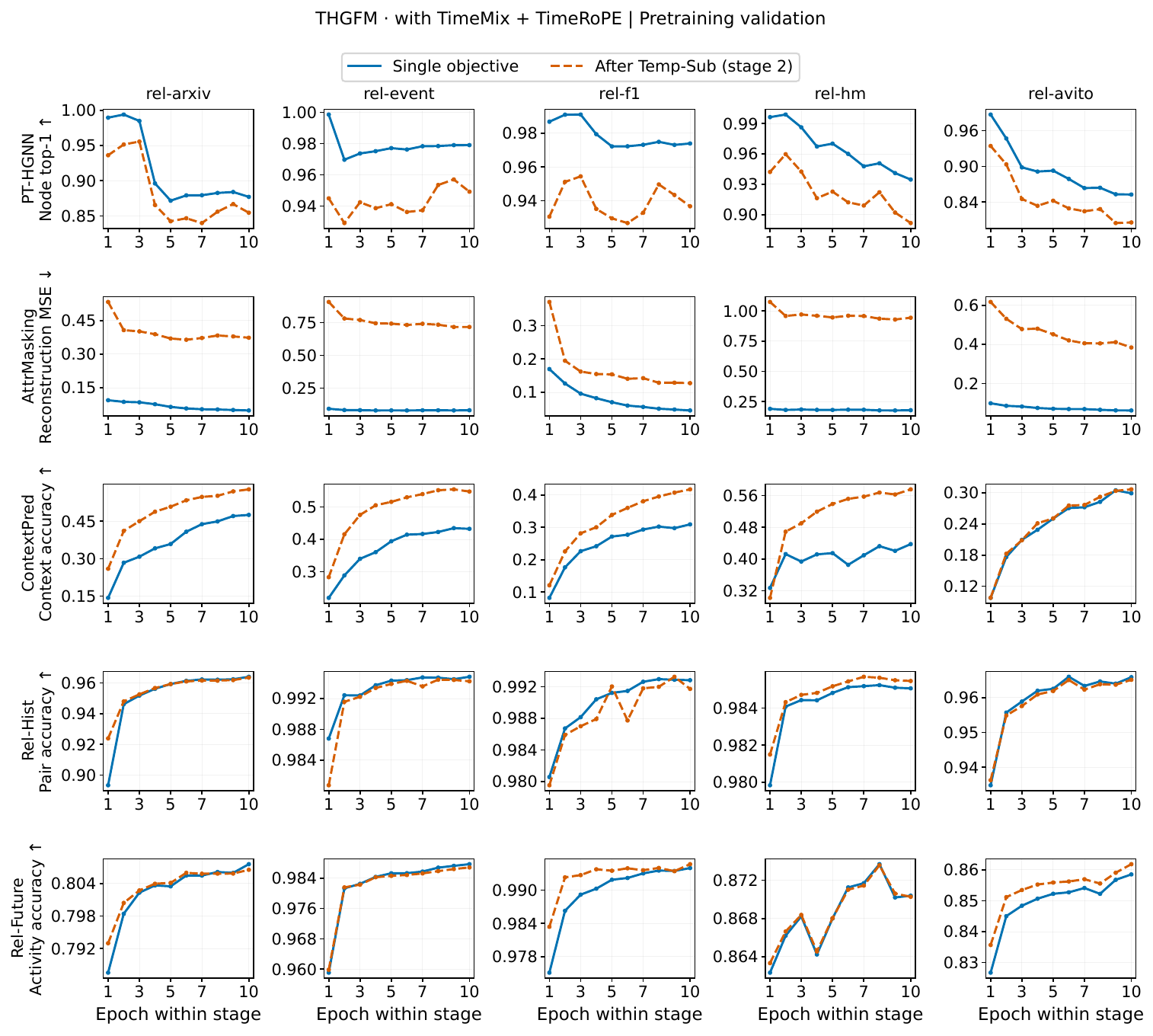}
\caption{Pretraining metrics for \method{} with \mix{} and \rotm{}: single objectives and stage 2 after \gsubgraph{}.}
\label{fig:display-pretrain-thgfm-with-2}
\label{fig:curve-pretrain-thgfm-with-2}
\end{figure}
\begin{figure}[p]
\centering
\includegraphics[width=\textwidth]{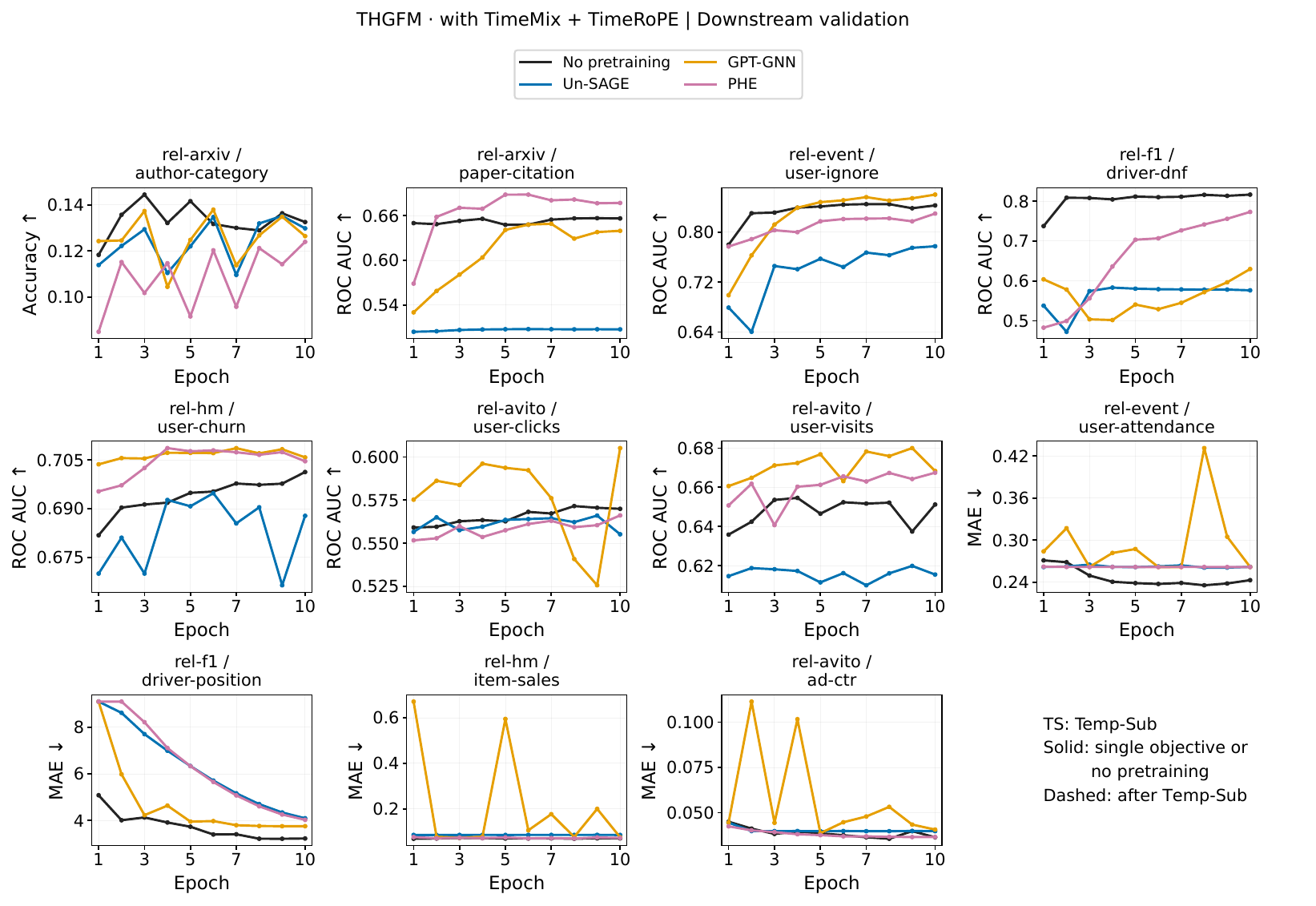}
\caption{Fine-tuning validation metrics for \method{} with \mix{} and \rotm{}: single-stage baselines.}
\label{fig:display-finetune-thgfm-with-1}
\label{fig:curve-finetune-thgfm-with-1}
\end{figure}
\begin{figure}[p]
\centering
\includegraphics[width=\textwidth]{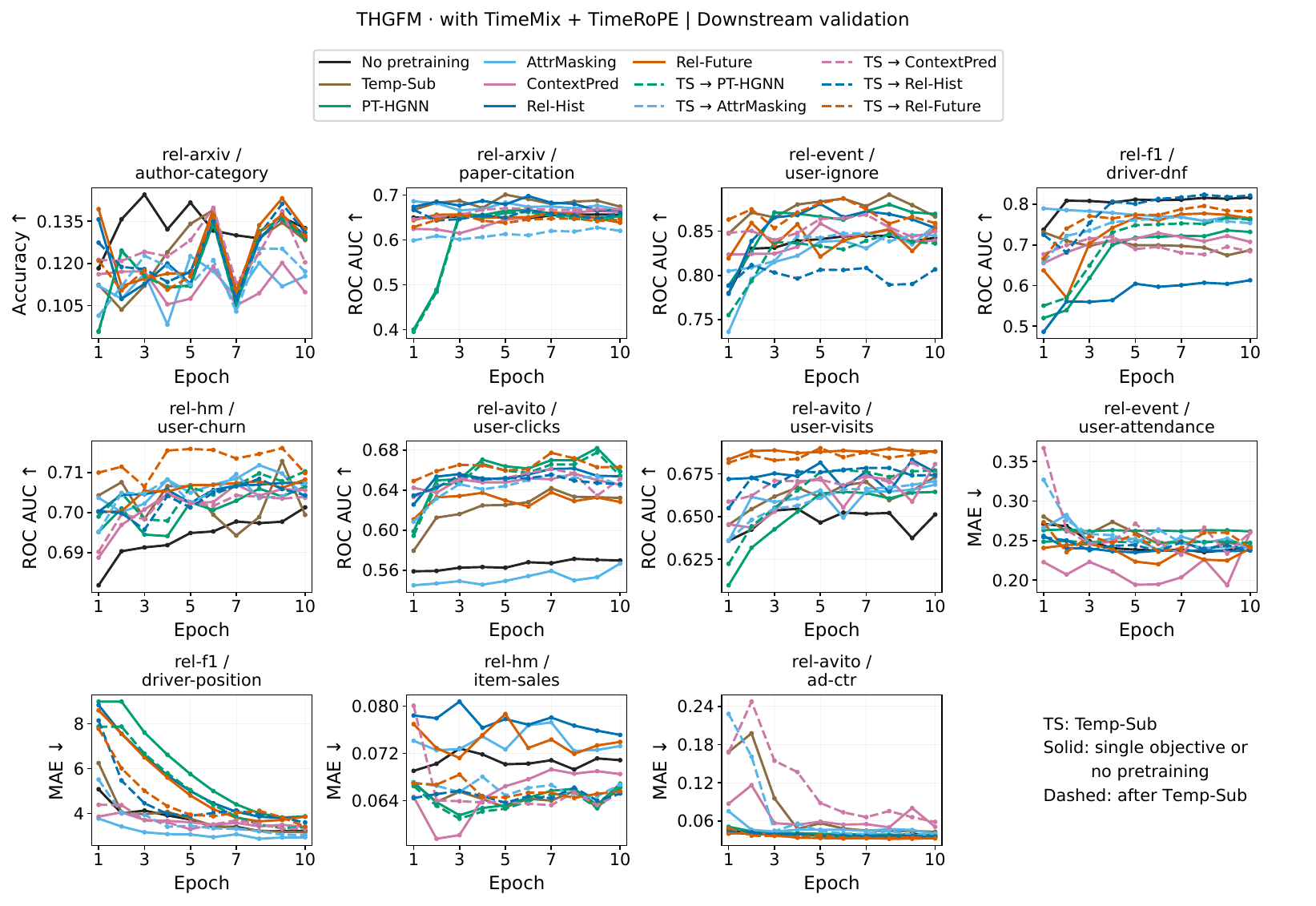}
\caption{Fine-tuning validation metrics for \method{} with \mix{} and \rotm{}: single and staged objectives.}
\label{fig:display-finetune-thgfm-with-2}
\label{fig:curve-finetune-thgfm-with-2}
\end{figure}

\begin{figure}[p]
\centering
\includegraphics[width=0.87\textwidth]{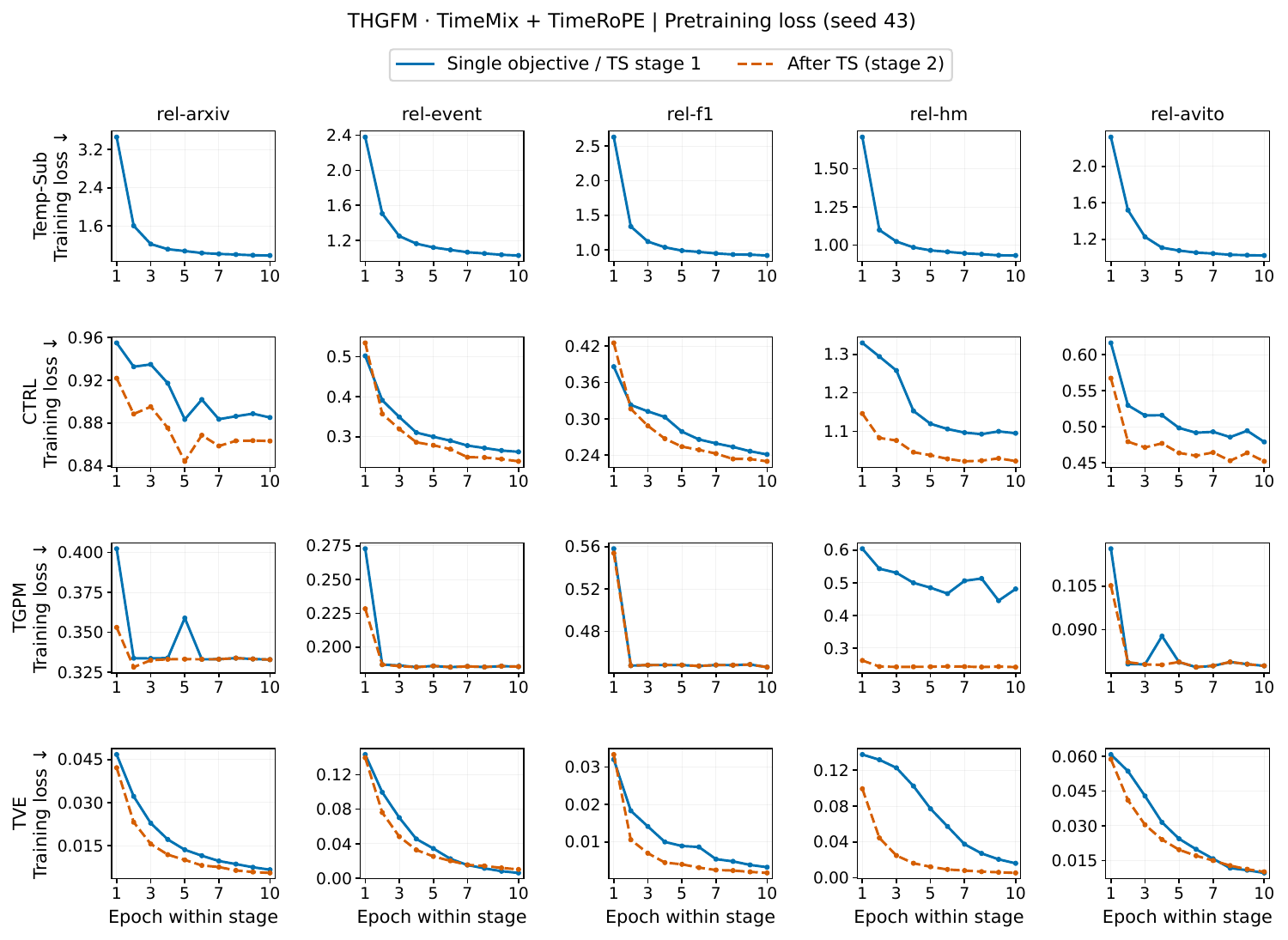}
\caption{Pretraining losses for \method{} with \mix{} and \rotm{}: CTRL, TGPM, TVE, and the shared \gsubgraph{} first stage.}
\label{fig:curve-pretrain-thgfm-with-3}
\end{figure}

\begin{figure}[p]
\centering
\includegraphics[width=\textwidth]{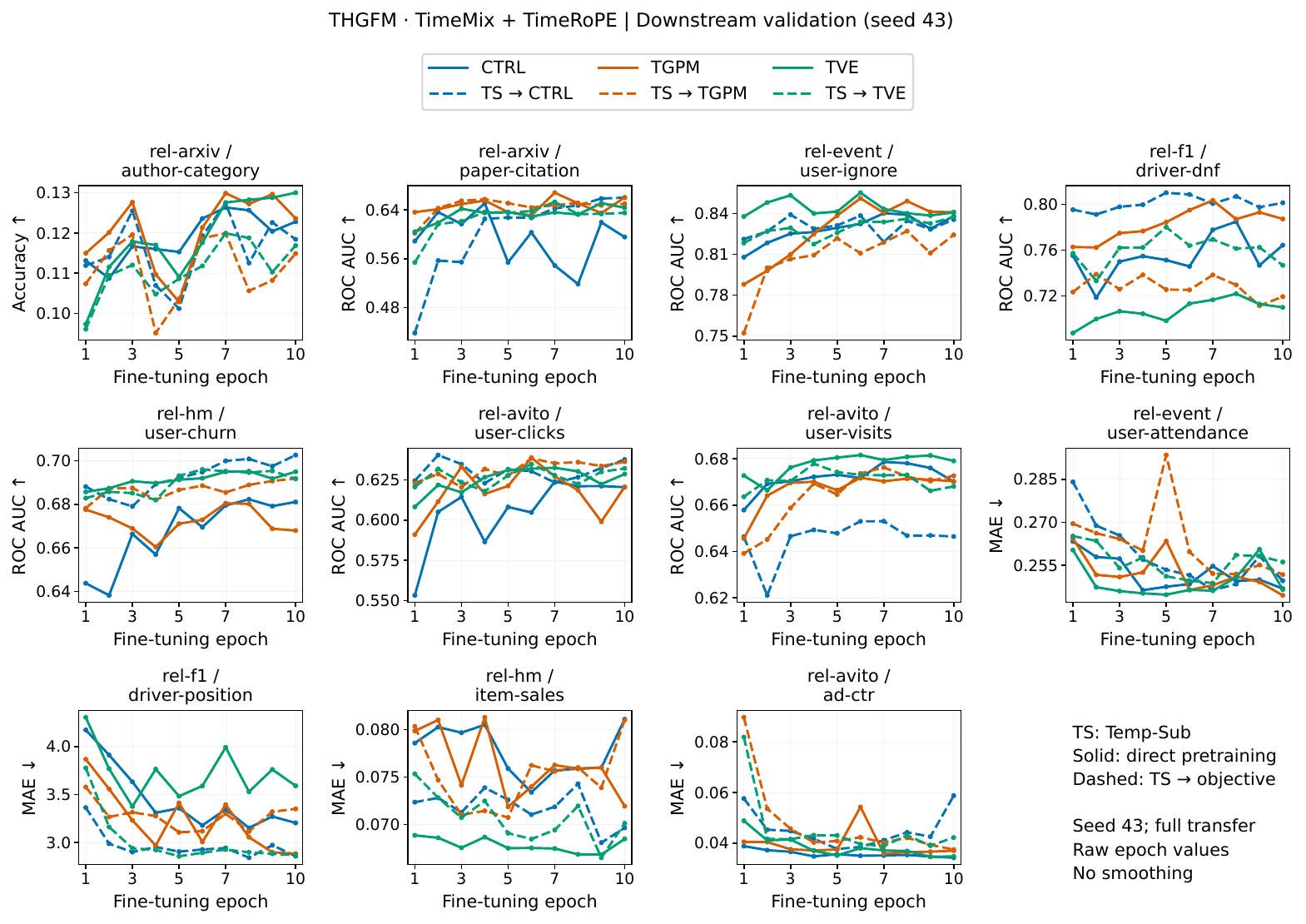}
\caption{Fine-tuning validation metrics for \method{} with \mix{} and \rotm{}: CTRL, TGPM, TVE, and their TS-initialized variants.}
\label{fig:curve-finetune-thgfm-with-3}
\end{figure}

\clearpage
\endgroup

\end{document}